\documentclass{article}

\PassOptionsToPackage{numbers, compress, sort}{natbib}

\usepackage[main, final]{neurips_2026}

\usepackage[utf8]{inputenc} %
\usepackage[T1]{fontenc}    %
\usepackage{hyperref}       %
\usepackage{url}            %
\usepackage{booktabs}       %
\usepackage{amsfonts}       %
\usepackage{nicefrac}       %
\usepackage{microtype}      %
\usepackage{xcolor}         %

\usepackage{amsmath,amsthm,amssymb}
\usepackage{tabulary}
\usepackage{colortbl}
\usepackage{algorithmic}
\usepackage{graphicx}       %
\usepackage{subcaption}
\usepackage[bottom]{footmisc}
\usepackage{indentfirst} 
\usepackage{stmaryrd}
\usepackage{endnotes}
\usepackage{verbatim}
\usepackage{tikz}
\usepackage{threeparttable}
\usepackage{comment}
\usepackage{algorithm}
\usepackage{titlesec}
\usepackage{multirow}
\usepackage{makecell}

\usepackage{chngcntr}
\usepackage{dsfont}
\usepackage{hyperref}
\usepackage{enumitem}
\usepackage{array}

\newtheorem{lemma}{Lemma}
\newtheorem{theorem}{Theorem}
\newtheorem{proposition}{Proposition}

\newtheorem{definition}{Definition}
\newtheorem{rmk}{Remark}

\newtheorem{assumption}{Assumption}

\newcommand{\utwi}[1]{\mbox{\boldmath $ #1$}}

\newcommand{\bx}{\mathrm{\bf x}}
\newcommand{\by}{\mathrm{\bf y}}

\newcommand{\bd}{\mathrm{\bf d}}

\newcommand{\bB}{\mathrm{\bf B}}

\newcommand{\bM}{\mathrm{\bf M}}

\newcommand{\bF}{\mathrm{\bf F}}

\newcommand{\bz}{\mathrm{\bf z}}
\newcommand{\bH}{\mathrm{\bf H}}

\newcommand{\bR}{\mathrm{\bf R}}

\newcommand{\bU}{\mathrm{\bf U}}
\newcommand{\bX}{\mathrm{\bf X}}
\newcommand{\bY}{\mathrm{\bf Y}}
\newcommand{\bLambda}{{\utwi{\mathnormal\Lambda}}}

\newcommand{\bphi}{{\utwi{\mathnormal\phi}}}
\newcommand{\balpha}{{\utwi{\mathnormal\alpha}}}

\newcommand{\cA}{{\cal A}}
\newcommand{\cB}{{\cal B}}
\newcommand{\cC}{{\cal C}}

\newcommand{\cG}{{\cal G}}

\def\calA{\mathcal{A}} \def\calB{\mathcal{B}} \def\calC{\mathcal{C}}
\def\calD{\mathcal{D}}  
  \def\calI{\mathcal{I}}

\newcommand{\bOmega}{\boldsymbol{\Omega}}

\def\E{\mathbb{E}}
\def\P{\mathbb{P}}

\def\R{\mathbb{R}}

\title{Covariate-Adjusted Deep Causal Learning for Heterogeneous Panel Data Models}

\author{
Guanhao Zhou, Yuefeng Han, and Xiufan Yu\\
  Department of Applied and Computational Mathematics and Statistics\\
  University of Notre Dame \\
  \texttt{\{gzhou4, yuefeng.han, xiufan.yu\}@nd.edu}
}

\begin{document}

\maketitle

\begin{abstract}
  This paper studies the task of estimating heterogeneous treatment effects in causal panel data models with covariate effects. We propose a novel \textbf{Co}variate-Adjusted \textbf{DE}ep C\textbf{A}usal \textbf{L}earning (\textbf{CoDEAL}) framework that cohesively deals with the underlying heterogeneity and nonlinearity of both panel units and covariate effects. CoDEAL integrates nonlinear covariate effect components (parameterized by a feed-forward neural network) with nonlinear factor structures (modeled by a multi-output autoencoder) to form a heterogeneous causal panel model. 
  The nonlinear covariate component flexibly captures complex covariate influences on outcomes, and the nonlinear factor decomposition enables CoDEAL to effectively capture both cross-sectional and temporal dependencies inherent in the data panel. 
  This latent structural information is subsequently integrated into a customized matrix completion algorithm, thereby facilitating more accurate counterfactual imputation.
  Moreover, the use of a multi-output autoencoder explicitly accounts for heterogeneity across units and enhances the model interpretability of the latent factors.
  We establish theoretical guarantees on the convergence of the estimated counterfactuals and demonstrate the compelling performance of the proposed method using extensive simulation studies and real-world data applications. 
\end{abstract}

\section{Introduction}
\vspace{-2ex}
Causal inference in panel data settings has attracted growing interest in recent years, with broad applications across various fields such as economics \citep{clarke2024synthetic}, life science \citep{helske2024estimating}, political science \citep{imai2021use}, and social science \citep{imbens2024causal}. 
Causal panel data analysis is inherently a missing data problem due to the \emph{fundamental problem of causal inference}: for each unit and time period, we can only observe either the treated or the untreated outcome, but never both. The panel data structure brings unique challenges to causal analysis. Different units may get exposed to the treatment at various time periods, forming a block-wise non-random missing pattern \citep{athey2022design}. The presence of both temporal and cross-sectional dependencies could bias the estimation if not properly addressed \citep{sun2021estimating}, and unit-specific heterogeneity can confound causal interpretation \citep{millimet2023fixed}, especially if additional covariates are present. 

\vspace{-0.5ex}
To estimate causal effects, one natural approach is to first impute the missing counterfactual, and then examine the difference between the imputed counterfactuals and the actual observed outcomes. \citet{athey2021matrix} formulated the imputation of missing counterfactuals as a matrix completion task with nuclear norm minimization. 
It is important to note that standard matrix completion approaches are not directly suitable for causal imputation \citep{choi2024matrix}. Most conventional matrix completion methods operate under the assumption that missing entries are \emph{missing at random} (MAR), which enables estimation of missing values without explicitly modeling the missingness mechanism. However, in causal panel data settings where different units receive treatment at different times, missing entries correspond to counterfactual outcomes that are systematically unobserved due to treatment assignments, invalidating the MAR assumption. 
As a result, applying standard matrix completion blindly without properly accounting for this structured, non-random missingness can lead to biased estimation \citep{agarwal2023causal}. 

\vspace{-0.5ex}
In recent years, a growing body of work has focused on developing causal matrix completion methods that handle missing-not-at-random panel designs. 
Besides the optimization-based causal matrix completion \citep{athey2021matrix,choi2024matrix}, another line of research leverages factor models \citep{bai2003,bai2009panel} to capture the latent low-rank structure of observed entries. 
Most factor-model-based causal matrix completion methods in the literature \citep{xu2017generalized,bai2021matrix,agarwal2023causal,xiong2023large,yan2024entrywise} posit a linear factor structure. Linear factor analysis helps capture cross-sectional and temporal dependencies to some extent \citep{han2024tensor,han2024cp,yu2024power,han2022rank}, but is limited in capturing complex nonlinear relationships, motivating the generalization to nonlinear factor models \citep{yalcin2001nonlinear,xiu2024deep}.  

\vspace{-0.5ex}
Another key aspect in panel analysis is to account for unit-specific heterogeneity \citep{millimet2023fixed,semenova2023inference}. Most relevant works in the literature still focus on the overall average treatment effects across units \citep{athey2021matrix,athey2025identification}, as opposed to unit-specific heterogeneous treatment effects. Pooled estimates may mask important variations across units. Units in a panel often differ in characteristics and respond to treatment differently \citep{wager2018estimation,nandy2023generalized}, which may bias the estimates of treatment effects, motivating a critical need to estimate unit-specific heterogeneous treatment effects and adjust for covariate effects if possible.  

\vspace{-0.5ex}
In this paper, we develop \textbf{Co}variate-Adjusted \textbf{DE}ep C\textbf{A}usal \textbf{L}earning  (\textbf{CoDEAL}) for panel data models, a deep-learning-based causal learning method for estimating heterogeneous treatment effects. CoDEAL captures covariate effects with a deep neural network regression, and models the covariate-adjusted outcomes via a nonlinear factor structure. 
Our contributions can be summarized as follows.

\vspace{-0.5ex}
\vspace{-1ex}
\begin{itemize}[leftmargin = *]
    \item %
    CoDEAL provides \textbf{a causal matrix completion method} that addresses the non-random missingness mechanism in causal panel data analysis. 
    By leveraging nonlinear factor analysis, CoDEAL \textbf{exploits both cross-sectional and temporal dependencies}.
    This latent structural information is subsequently integrated into the matrix completion procedure, thereby facilitating more accurate imputation of missing counterfactual outcomes. Theoretically, we rigorously prove that CoDEAL provides consistent estimates of unit-specific heterogeneous treatment effects. 
    \vspace{-0.5ex}
    \item %
    CoDEAL employs flexible model structures and powerful neural network architectures to cohesively deal with the underlying \textbf{heterogeneity} and \textbf{nonlinearity} of \textbf{both panel units and covariate effects}.
    Algorithmically, we begin with a feed-forward deep neural network (DNN) to remove nonlinear covariate effects, followed by a multi-output autoencoder (AE) to recover nonlinear latent factors and explicitly account for unit-specific heterogeneity.
    \vspace{-0.5ex}
    \item %
    CoDEAL offers \textbf{a unifying framework} for heterogeneous causal panel models in the presence of covariates. By combining nonlinear covariate adjustment with deep latent factor models, CoDEAL generalizes beyond traditional linear factor models and unifies a class of factor-model-based causal matrix completion approaches \citep[e.g.,][]{%
    bai2021matrix,agarwal2023causal,xiong2023large,yan2024entrywise}. Moreover, with different choices of factor models and neural network architectures, the proposed framework can be further generalized to 
    accommodate various data types such as spatial and tensor data. 
\end{itemize}

\vspace{-0.5ex}
\vspace{-1ex}
\noindent \textbf{Related Work.} %
\textbf{(a) An overview of causal learning methods in panel data models.} One traditional approach to estimate treatment effects in causal panel models is \emph{the difference-in-differences} (DiD) \citep{imai2021use,athey2022design,wing2024designing} that compares the average changes in outcomes over time between the treated and untreated units under a \emph{parallel trend} assumption. Alternatively, one can estimate the causal effects by first imputing the missing counterfactual outcomes, and then examining the difference between the imputed counterfactuals and the actual observed outcomes. 
Following this strand, two approaches have emerged in recent causal panel literature: \emph{the unconfoundedness-based approach} (also known as horizontal regression) \citep{rosenbaum1983central,imbens2015causal} and \emph{the synthetic control method} (also known as vertical regression) \citep{abadie2021using,abadie2003economic,abadie2010synthetic,abadie2015comparative,doudchenko2016balancing}. %
The close connections between the DiD, unconfoundedness-based methods, synthetic control approaches, and causal matrix completion estimators were revealed by \citet{athey2021matrix}. Despite their distinct appearances, \citet{athey2021matrix} showed that the linear versions of all four estimators can be characterized as solutions to the same optimization problem with the exact same objective function, but subject to different restrictions on parameters of that optimization. 
In parallel, a separate semiparametric literature develops doubly robust and machine-learning-based semiparametric estimators \citep{chernozhukov2018, sant2020doubly, kennedy2024semiparametric, lan2025meta, testa2025efficient} under DiD-type identification assumptions. Our framework aligns conceptually with this push for flexibility; however, it departs from the semiparametric approach by adopting nonlinear, time-varying covariate functions and a nonlinear latent factor structure.
We refer readers to \cite{arkhangelsky2024causal} for a comprehensive review of past and recent progress on causal panel models.

\vspace{-0.5ex}
\textbf{(b) Causal matrix completion and deep matrix completion.} 
Conceptually, framing causal estimation and inference as the recovery of systematically missing counterfactual information is widely adopted across heterogeneous data structures and missingness patterns \citep{holland1986statistics,ding2018causal,de2025causal,de2025recovery}. Here, we focus specifically on matrix completion approaches that are most relevant to our study on causal panel data models.
A surge of interest in the recent past has led to new research on causal matrix completion methods capable of addressing missing-not-at-random panel data, e.g., optimization-based methods \citep{athey2021matrix,choi2024matrix} and factor-model-based methods \citep{xu2017generalized,bai2021matrix,agarwal2023causal,xiong2023large,yan2024entrywise}. %
There has been a handful of deep-learning-based methods for matrix completion, e.g., \citep{fan2018matrix,radhakrishnan2022simple,xiu2024deep,fan2024neuron}. However, most of them assume MAR, and therefore, are not directly suitable for causal imputation in panel data models.

\section{Methodology}
\vspace{-1ex}
\subsection{Problem Setup}

\vspace{-1ex}
\textbf{Heterogeneous Treatment Effects %
in Causal Panel Data Models. }
We begin by setting up the causal learning problem in panel data. Consider a panel data setting with $N$ units over $T$ periods. 
Let $[n]$ denote the set $\{1, 2, \ldots, n\}$.
For each unit $i \in [N]$ and time point $t \in [T]$, there are two potential outcomes, $Y_{it}(1)$ and $Y_{it}(0)$, representing the outcomes the $i$-th unit would experience at time $t$ under treatment and control, respectively.
The treatment assignment is captured by a binary indicator variable $W_{it} \in \{0, 1\}$, where $W_{it} = 1$ indicates that the unit $i$ receives the treatment at time $t$ and $W_{it} = 0$ otherwise. Suppose we also have access to some unit-specific covariates $\bX_i \in \mathbb{R}^P$ that could potentially affect the observations $\{Y_{it}, t \in [T]\}$. We denote the matrix forms of the observed outcomes, treatment indicators, and covariates by $\bY = (Y_{it}) \in \mathbb{R}^{N \times T}$, $\bOmega = (W_{it}) \in \mathbb{R}^{N \times T}$, and $\bX = (\bX_1, \ldots, \bX_N)^\top \in \mathbb{R}^{N \times P}$, respectively. Let $\bY(1) = (Y_{it}(1))\in \mathbb{R}^{N \times T}$ and $\bY(0) = (Y_{it}(0))\in \mathbb{R}^{N \times T}$ be the matrices of potential outcomes of the treated and untreated, respectively. Our aim %
in this work is to estimate the unit-specific (potentially heterogeneous) average treatment effects on the treated (ATT), $\tau_i^\star = \mathbb{E}_{t} [Y_{it}(1) - Y_{it}(0) \mid W_{it} = 1]$ for each treated unit $i$.%

\textbf{Staggered Adoption Design (SAD)}. 
This work focuses on SAD \citep{athey2022design}, where treatments can be rolled out at different times across various units and are irreversible once initiated. 
We assume there is at least one never-treated panel unit. %
Details of SAD are in Appendix \ref{append:staggered-design}. 

\textbf{Causal Assumptions. } To ensure valid interpretations of the estimators, we adopt the following assumptions throughout the analysis. (i) (Stable Unit Treatment Value Assumption (SUTVA).) The potential outcomes for a given unit at a particular time depend solely on the treatment administered to that unit at that time, and are unaffected by treatments assigned to other units or at other time points. (ii) (Static Treatment Effects.)\footnote{The static treatment assumption is not intrinsic to the proposed algorithm, but can be extended to accommodate time-varying contemporaneous treatment effects. We discuss this extension in Remark \ref{rmk:time-varying-treatment}. For clarity of exposition, we maintain the static treatment assumption when presenting the methodology and theoretical results preceding Remark \ref{rmk:time-varying-treatment}.} %
The treatment effects are non-dynamic, i.e., do not change over time.

\textbf{Challenges and Limitations of Existing Methods. } The \emph{fundamental problem of causal inference} implies that $Y_{it}(0)$ and $Y_{it}(1)$ cannot be observed simultaneously. The panel data forms two partially observed data matrices $\bY(0)$ and $\bY(1)$, where the missing pattern is determined according to the treatment assignments. To estimate the treatment effects, one natural idea is to estimate the unobserved counterfactuals (impute the missing potential outcomes). Since our focus is the unit-specific ATT and $\{Y_{it}(1): W_{it}=1\}$ is observed, it suffices to estimate $\{Y_{it}(0): W_{it}=1\}$ to impute missing elements in $\bY(0)$. Let $\{\widehat{Y}_{it}(0): W_{it}=1\}$ denote imputed values.
The unit-specific ATT can then be estimated by $\widehat\tau_i = \sum_{t: W_{it} = 1} [Y_{it}(1)-\widehat{Y}_{it}(0)]/\sum_{t} W_{it}$ for each treated unit $i$. 

There are three main limitations of existing methods that this work aims to overcome in the task of modeling unit-specific ATT and imputing missing potential outcomes. 
First,  the vast majority of matrix completion methods rely on the MAR assumption, which is typically violated in SADs, %
resulting in biased estimates. 
Second, existing matrix-completion-based causal panel models mostly focus on the overall (non-unit-specific) ATT, without adequately accounting for unit-level heterogeneity.
Third, many existing methods often lack the capacity to disentangle the causal impact of treatment from confounding covariate effects and to model the underlying complex nonlinear dependencies, limiting both the precision of counterfactual predictions and the interpretability of causal analysis. 

\vspace{-1ex}
\subsection{Covariate-Adjusted Deep Causal Learning (\textbf{CoDEAL}) for Panel Data Models}
\label{subsec:method-CoDEAL}

\vspace{-1ex}
To tackle the above challenges, we propose the CoDEAL method that (i) leverages the staggered adoption structure to fully exploit both cross-sectional and temporal dependencies for more accurate imputation of missing potential outcomes, (ii) models the unit-specific heterogeneity explicitly to accommodate variation in treatment effects across units, and (iii) involves flexible model structures and employs powerful neural network architectures to efficiently capture complex nonlinear relationships. 

\vspace{-1ex}
For each unit $i \in [N]$ at time $t \in [T]$, we model the observed outcome by
\begin{equation}
   Y_{it}= \phi_i^\star(\bF^\star_{t}) + g^\star_{t}(\bX_i) + \tau_i^\star \cdot \mathds{1}\{W_{it}=1\} + \varepsilon_{it},
\label{eq:CoDEAL-Model}
\end{equation}
\vspace{-4ex}

where $\bF^\star_t \in \mathbb{R}^K$ is a $K$-dimensional vector of latent factors, $\phi^\star_{i}(\cdot): \mathbb{R}^K \rightarrow \mathbb{R}$ is a potentially nonlinear factor loading function,  $g^\star_t(\cdot): \mathbb{R}^P \rightarrow \mathbb{R}$ is a potentially nonlinear function that captures covariate effects
and $ \mathds{1}\{\cdot\}$ is an indicator function, and $\varepsilon_{it}$ is the idiosyncratic noise. An illustrative DAG of Model \eqref{eq:CoDEAL-Model} is in Appendix \ref{append:DAG}.
CoDEAL employs a nonlinear factor model \citep{yalcin2001nonlinear} to characterize the underlying low-rank structure of the covariate-adjusted outcomes. By letting $\phi^\star_{i}(\cdot)$ take different forms, CoDEAL unifies a class of factor-model-based causal matrix completion approaches, e.g., \citep{xu2017generalized,bai2021matrix,agarwal2023causal,xiong2023large,yan2024entrywise}. 
As shown by \eqref{eq:CoDEAL-Model},
CoDEAL accommodates heterogeneous but non-dynamic treatment effects. The covariate effects can be heterogeneous across units and time-varying over periods. 
Here, we allow $N$ and $T$ to be large and diverging, while $P$ and $K$ are fixed and small.

\begin{figure*}[!t]
\centering
\includegraphics[width=0.8\textwidth]{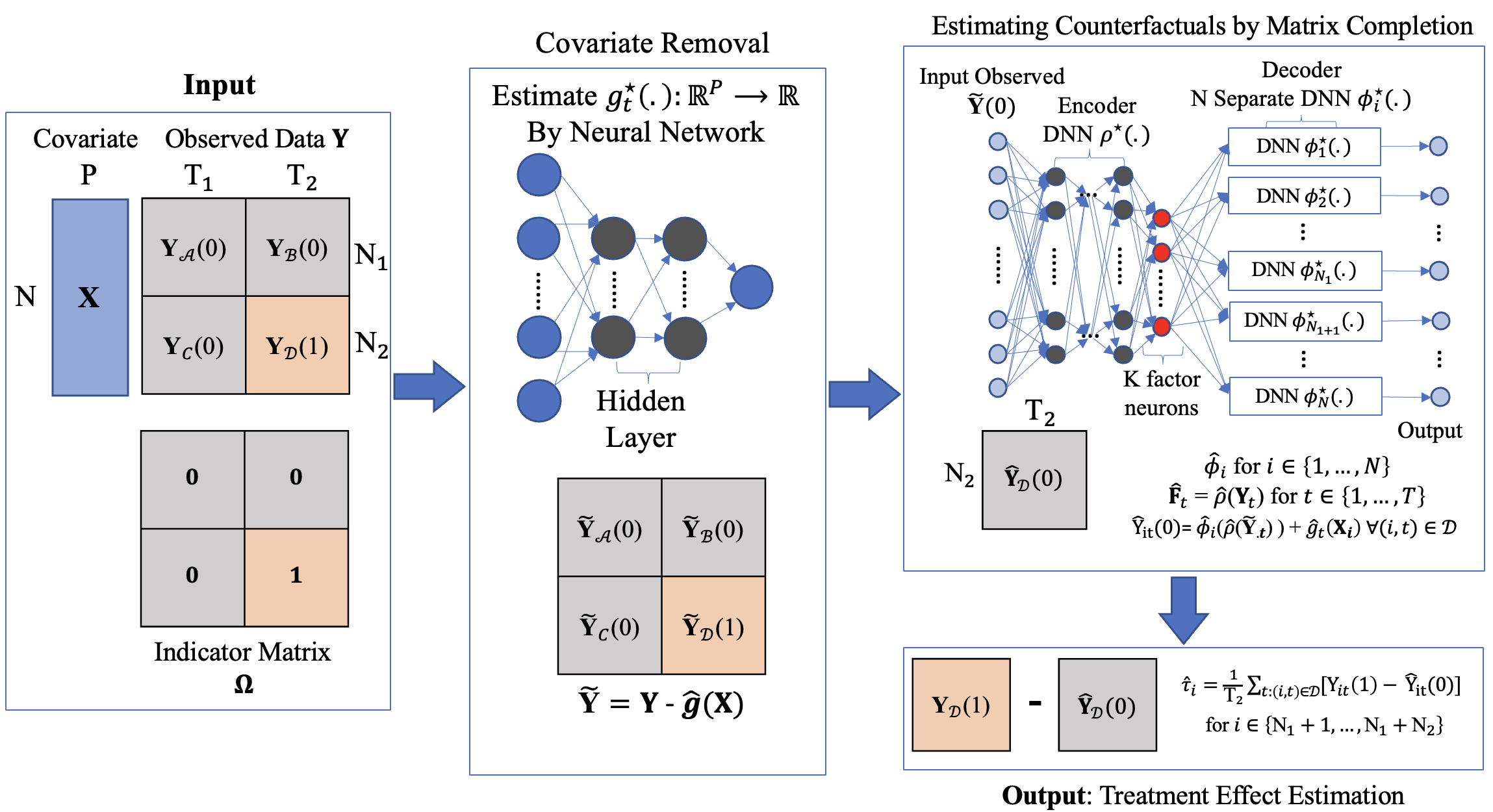}
\vspace{-1ex}
    \caption{\small A graphical illustration of the proposed CoDEAL in a four-block design. Gray and orange refer to the untreated and treated blocks, respectively. Inputs are observed outcomes \(\mathbf{Y}\in\mathbb{R}^{N\times T}\), covariates \(\bX\in\mathbb{R}^{N\times P}\), indicator matrix \(\boldsymbol{\Omega}\in\mathbb{R}^{N\times T}\). Output is the estimated unit-specific ATT $\widehat{\tau}_i$.}
    \label{fig:method}
\end{figure*}

For clarity, we first present the learning procedures using the simplest form of a $2\times 2$ staggered adoption (also known as the four-block design) in Section \ref{subsec:CoDeal-Four-Block}, and generalize the algorithm to the more general SAD in Section \ref{subsec:CoDeal-Staggered-Design}. More details about the four-block and SAD as well as their graphical illustrations are in  Appendix \ref{append:staggered-design}. A graphical illustration of CoDEAL in the four-block design is provided in Figure \ref{fig:method}. 
CoDEAL in general SAD is summarized in Algorithm \ref{alg:staggered}. 

\vspace{-1ex}
\subsubsection{CoDEAL in Four-Block Design} \label{subsec:CoDeal-Four-Block}
\vspace{-1ex}

We begin with the four-block design, the simplest form of staggered adoption and often regarded as a foundational building block for more general SADs. The four-block design is a $2 \times 2$ SAD, where a subset of $N_1$ units never receive the treatment while the remaining $N_2 = N-N_1$ units are exposed to the irreversible treatment starting at time $t=T_1+1$. Without loss of generality, we rearrange the panel such that the first $N_1$ units ($i=1,\ldots, N_1$) are untreated. We refer to the time periods $t\in [T_1]$ as pre-treatment periods and $t\in [T]\backslash [T_1]$ as post-treatment periods. Let $T_2 = T-T_1$ denote the duration of post-treatment periods. The design naturally decomposes the panel into four blocks:
\vspace{-0.5ex}
\begin{equation}
\begin{aligned}
\bY = 
\begin{bmatrix}
\bY_\calA & \bY_\calB \\
\bY_\calC & \bY_\calD
\end{bmatrix}
=
\begin{bmatrix}
\bY_{\calA}(0) & \bY_{\calB}(0) \\
\bY_{\calC}(0) & \bY_{\calD}(1)
\end{bmatrix},
\text{ and }  & \bY(0)  =
\begin{bmatrix}
\bY_{\calA}(0) & \bY_{\calB}(0) \\
\bY_{\calC}(0) & ?
\end{bmatrix}
\end{aligned}
\label{eq:four-block}
\end{equation}
where $\bY_\calA \in \mathbb{R}^{N_1\times T_1}$, $\bY_\calB \in \mathbb{R}^{N_1\times T_2}$, $\bY_\calC \in \mathbb{R}^{N_2\times T_1}$, and $\bY_\calD \in \mathbb{R}^{N_2\times T_2}$. By construction, $\bY_\calA$, $\bY_\calB$, $\bY_\calC$ are observed outcomes of the untreated with $Y_{it}  = Y_{it}(0)$ for $(i,t)\in \calA \cup \calB \cup \calC$, while $\bY_\calD$ are observed outcomes of the treated with $Y_{it}  = Y_{it}(1)$ for $(i,t)\in \calD$, leaving a missing block in $\bY(0)$. To estimate $\tau_i^\star$, it suffices to impute the missing values of $\bY_{\calD}(0)$.  Denote the imputed block as $\widehat{\bY}_{\calD}(0)$. The unit-specific ATT can be then estimated by 
$\widehat{\tau}_i = \frac{1}{T_2} \sum_{t: (i,t)\in\calD} [Y_{it}(1) -\widehat{Y}_{it}(0)] \text{ for } i \in [N]\backslash [N_1]$. %

\noindent \textbf{Modeling Covariate Effects. } To account for the potentially nonlinear, heterogeneous, and time-varying effects brought by covariates, we consider a fully connected DNN for $g_t^\star(\cdot)$. 
We construct our model using a fully connected DNN with ReLU activation, %
due to its strong empirical performance. We refer to this architecture as a \textit{deep ReLU network}. Let $L\in\mathbb N$ denote the network depth and let $\bd = (d_1, \ldots, d_{L+1}) \in \mathbb{N}^{L+1}$ specify the layer widths. A deep ReLU network is a function mapping $\mathbb{R}^{d_0}$ to $\mathbb{R}^{d_{L+1}}$ and takes the form
\vspace{-1ex}
\begin{equation}\label{eq:relu}
h(x) = \mathcal{L}_{L+1} \circ \bar{\sigma}_{L} \circ \mathcal{L}_L \circ \bar{\sigma}_{L-1} \circ \cdots \circ \mathcal{L}_2 \circ \bar{\sigma}_{1} \circ \mathcal{L}_1(x),
\end{equation}
\vspace{-4ex}

where each $\mathcal{L}_\ell(z) = M_\ell z + b_\ell$ is an affine transformation with weight matrix $M_\ell \in \mathbb{R}^{d_\ell \times d_{\ell-1}}$ and bias vector $b_\ell \in \mathbb{R}^{d_\ell}$, and $\bar{\sigma}_\ell : \mathbb{R}^{d_\ell} \to \mathbb{R}^{d_\ell}$ applies the ReLU activation function entrywise. For simplicity, we refer to both $M_\ell$ and $b_\ell$ as the weights of the deep ReLU network.

\begin{definition}[Deep ReLU network class]
For any $L \in \mathbb{N}$, $\bd \in \mathbb{N}^{L+1}$, and constants $B, C>0$, we define the class of deep ReLU networks with depth $L$, width $\bd$, and weights $\bM=\{M_1,...,M_{L+1},b_1,...,b_{L+1}\}$ bounded by $C$ as
\vspace{-0.5ex}
{%
\begin{align*}
\mathcal{G}_{d_0}^{d_{L+1}}(L,\bd, B, C) = \{ \tilde{h}(x) =  \text{sgn}(h(x))(|h(x)| \wedge B) : h \text{ of form } \eqref{eq:relu},  \, \|M_\ell\|_{\infty} \leq C, \, \|b_\ell\|_{\infty} \leq C \}.
\end{align*}
}
\end{definition}

\vspace{-1ex}
The covariate effects can be estimated via 
\begin{equation}
    \widehat{g}_t(\cdot) = \underset{g_t \in \mathcal{G}_{P}^{1}(L,\bd, B, C) }{\arg\min} \sum_{i: W_{it}=0} (Y_{it} - g_t(\bX_i) )^2 ,
    \label{eq:g-hat}
\end{equation}
for a certain functional class $\mathcal{G}_{P}^{1}(L,\bd, B, C)$ specified in Appendix~\ref{section:proofs}. 
We define the covariate-adjusted observed outcomes for unit $i \in [N]$ at time $t \in [T]$ as $\widetilde{Y}_{it} = Y_{it} - \widehat{g}_t(\bX_i).$

\noindent \textbf{Estimating Counterfactuals by Matrix Completion via a Multi-Output Autoencoder.} With covariate-adjusted outcomes, it remains to model the nonlinear factor structure $\phi^\star_i(\bF_t^\star)$ in \eqref{eq:CoDEAL-Model}. Inspired by \cite{xiu2024deep}, we consider tackling $\phi^\star_i(\bF_t^\star)$ using a multi-output AE.

The architecture of the multi-output AE consists of a single DNN encoder and multiple independent DNN decoders. The encoder $\rho(\cdot) \in \mathcal{G}^{K_1}_{N}(L_1, \bd_1, B,C)$ maps the input to a $K_1$-dimensional %
bottleneck representation. For each of the $N$ output components, a distinct DNN decoder is used; specifically, the decoder for the $i$-th output is given by $\phi_i(\cdot) \in \mathcal{G}^{1}_{K_1}(L_2, \bd_2, B, C)$ for $i \in [N]$. Consequently, the $i$-th output of the multi-output AE for an input $x$ is given by $\phi_i \circ \rho(x)$. We formally define the function class of the multi-output AE as:

\vspace{-3ex}
{
\small
\begin{equation}\label{eq:ae-class}
\begin{aligned}
\mathcal{G}_{\text{AE}}^{K_1} 
:= \left\{ (\phi_1, \ldots, \phi_N) \circ \rho : \rho \in \mathcal{G}^{K_1}_{N}(L_1, \bd_1, B, C), \, \right.\left.\phi_i \in \mathcal{G}^{1}_{K_1}(L_2, \bd_2, B, C), \, i \in [N] \right\}  
\end{aligned}
\end{equation}
}
\vspace{-3ex}

where $K_1\ge K$. More details on this function class are provided in Appendix~\ref{section:proofs}.

With the multi-output AE, the encoder $\rho^\star(\cdot)$ effectively approximates the latent factors, facilitating the factor-model-based matrix completion. The unit-specific decoders $\phi_i^\star(\cdot)$, $i \in [N]$, are better equipped to capture the heterogeneity across units, in contrast to a traditional single-output AE which limits $\phi_1^\star = \ldots =\phi_N^\star$. 
The DNNs $(\widehat{\phi}_1,\dots,\widehat{\phi}_N)\circ\widehat{\rho}(\cdot)$ in the multi-output AE are estimated by 

\vspace{-3ex}
\begin{equation}
\begin{aligned}
\underset{(\phi_1,\dots,\phi_N)\circ\rho(\cdot)\in\mathcal{G}_{\text{AE}}^{K_1}}{\arg\min}
\sum_{(i,t):W_{it}=0}\bigl(\phi_i(\rho(\widetilde{\bY}_{\cdot t}))-\widetilde{Y}_{it}\bigr)^2 .
\end{aligned}
\label{eq:multi-output-ae}
\end{equation}
\vspace{-3ex}

The counterfactuals $\bY_{\calD}(0)$ can then be imputed by $\widehat{Y}_{it}(0) = \widehat{\phi}_i (\widehat\rho(\widetilde{\bY}_{\cdot t})) + \widehat{g}_t(\bX_i)$ for any $(i,t) \in \calD$. The complete steps of imputing counterfactuals in the four-block design are summarized as Algorithm \ref{alg:four-block} to facilitate further generalizations to the staggered design.

\begin{rmk}[Rationale of Sequential Estimation]
    We adopt a two-step estimation strategy over a joint estimation of covariate and factor effects for multiple reasons. It is not merely a procedural convenience, but rather a natural alignment with the data structure that improves stability and interpretability. We provide a detailed justification and supporting numerical evidence in Appendix~\ref{append:seq-vs-joint}.
\end{rmk}

\begin{algorithm}[h]
\caption{Estimating Counterfactuals by Matrix Completion in the Four-Block Design}
\small
\label{alg:four-block}
\begin{algorithmic}[1]
\STATE \textbf{Input:} Observed outcomes \(\bY\), covariates $\bX$, the estimated covariate effects $\widehat{g}_t(\cdot)$ from \eqref{eq:g-hat}, treatment indicators $\bOmega$ with a four-block design in \eqref{eq:four-block}, latent factor dimension $K$
\STATE \textbf{Output:} Imputed outcome matrix $\widehat{\bY}_{\calD}(0)$.
\STATE Obtain the covariate-adjusted observed outcomes by $\widetilde{Y}_{it} = Y_{it} - \widehat{g}_t(\bX_i)$ for $i\in[N]$, $t\in[T]$.
\STATE Fit a multi-output AE using the entries in the untreated part $\{\widetilde{Y}_{it}(0): W_{it}=0\}$ and obtain $(\widehat{\phi}_1, \ldots, \widehat{\phi}_N, \widehat{\rho})$ from \eqref{eq:multi-output-ae}.
\STATE Estimate the missing elements %
by $\widehat{Y}_{it}(0) = \widehat{\phi}_i (\widehat\rho(\widetilde{\bY}_{\cdot t})) + \widehat{g}_t(\bX_i)$ for any $(i,t) \in \calD$. 
\STATE \textbf{Return:} $\widehat{\bY}_{\calD}(0)$.
\end{algorithmic}
\end{algorithm}
\vspace{-2ex}
\begin{algorithm}[h]
\caption{\textbf{CoDEAL}: Staggered Adoption Design}
\small
\label{alg:staggered}
\begin{algorithmic}[1]
\STATE \textbf{Input:} Observed outcomes \(\mathbf{Y}\), covariates \(\bX\), treatment indicators \(\boldsymbol{\Omega}\), latent factor dimension \(K\).
\STATE \textbf{Output:} Estimated unit-specific ATT $\widehat{\tau}_i$ for each treated $i$.
\STATE Obtain the estimated covariate effect $\widehat{g}_t(\cdot)$ by \eqref{eq:g-hat}
\STATE Initialize $\widehat{\bY}(0)\in \mathbb{R}^{N\times T}$ by setting $\widehat{Y}_{it}(0) = Y_{it}$ for the untreated $\{(i,t): W_{it} = 0\}$.
\STATE Extract block partitions \(\{N_\xi \}_{1\le \xi \le r}\) and \(\{T_\eta\}_{1\le \eta\le r}\) from \(\boldsymbol{\Omega}\)
\FOR{\(\xi_0 = 2, \dots, r\)}
    \FOR{\(\eta_0 = r+2-\xi_0, \dots, r\)}
        \STATE Construct the four-block submatrix \( \bY^{(\xi_0,\eta_0)}\) and \(\boldsymbol{\Omega}^{(\xi_0,\eta_0)}\) according to Equation~\eqref{eq:staggered_design}.
        \STATE Call Algorithm~\ref{alg:four-block} with inputs \( \bY^{(\xi_0,\eta_0)}\), $\bX^{(\xi_0)}$, $\widehat{g}_t(\cdot)$, \(\boldsymbol{\Omega}^{(\xi_0,\eta_0)}\), and \(K\) to obtain the imputed block $\widehat{\bY}_{\calD_{(\xi_0,\eta_0)}}(0)$, and extract the estimated block \(\widehat{\bY}_{(\xi_0,\eta_0)}(0)\) from $\widehat{\bY}_{\calD_{(\xi_0,\eta_0)}}(0)$.
        \STATE Update the matrix \(\widehat{\bY}(0)\) with the imputed block \(\widehat{\bY}_{(\xi_0,\eta_0)(0)}\).
    \ENDFOR
\ENDFOR
\STATE Compute unit-specific estimated ATT as $\widehat\tau_i = \sum_{t: W_{it} = 1} [Y_{it}(1)-\widehat{Y}_{it}(0)]/\sum_{t} W_{it}$.
\STATE \textbf{Return:} Estimated unit-specific ATT $\widehat{\tau}_i$.
\end{algorithmic}
\end{algorithm}

\vspace{-2ex}
\subsubsection{CoDEAL in Staggered Adoption Design} \label{subsec:CoDeal-Staggered-Design}
\vspace{-1ex}
The extension to SAD is a natural generalization of the four-block design. 
Under the SAD, we rearrange the panels by sorting the $N$ units based on the time at which they began receiving the treatment in a descending order (with never-treated units at the top, followed by the latest adopters, and then progressively earlier adopters towards the bottom). This sorting procedure yields a block structure in the treatment indicator matrix $\bOmega$. By grouping the units based on their adoption timing and segmenting the overall time periods into relevant intervals, we partition the indicator matrix $\bOmega$ and outcome $\bY$ into submatrices $\bOmega_{(\xi, \eta)} \in \mathbb{R}^{N_{\xi} \times T_{\eta}}$ and $\bY_{(\xi, \eta)} \in \mathbb{R}^{N_{\xi} \times T_{\eta}}$ for $\xi=1,\ldots, r$ and $\eta =1,\ldots, r$, where $\sum_{\xi=1}^{r} N_{\xi} =N $ and $\sum_{\eta=1}^{r} T_{\eta} = T $. 
Examples of block partitions in the staggered design are included in Appendix \ref{append:staggered-design}.
Let $\calI_{tr} = \{(\xi,\eta): \bOmega_{(\xi, \eta)} = 1\}$ denote the index set of the treated blocks. By the construction of these partitions, the indices in $\calI_{tr}$ satisfy $\xi+\eta > r+1$. 

To estimate the unit-specific ATT, it suffices to estimate the counterfactuals of $\bY_{(\xi,\eta)}$ for all $(\xi,\eta) \in \calI_{tr}$. The imputation of $\bY_{(\xi,\eta)}(0)$ can be reduced to a four-block design problem \citep{yan2024entrywise}, therefore the method proposed in the previous Section \ref{subsec:CoDeal-Four-Block} can be directly applied. Specifically, we construct a four-block data matrix $ \bY^{(\xi_0,\eta_0)}$ as follows to estimate $\bY_{(\xi_0, \eta_0)}(0)$ for some $(\xi_0, \eta_0) \in \calI_{tr}$:
\vspace{-2ex}

{
\small
\begin{equation}
\begin{aligned}
\bY^{(\xi_0,\eta_0)} &=
\begin{bmatrix}
\bY_{\calA_{(\xi_0,\eta_0)}} & \bY_{\calB_{(\xi_0,\eta_0)}} \\
\bY_{\calC_{(\xi_0,\eta_0)}} & \bY_{\calD_{(\xi_0,\eta_0)}}
\end{bmatrix}
=
\begin{bmatrix}
\bY_{\calA_{(\xi_0,\eta_0)}}(0) & \bY_{\calB_{(\xi_0,\eta_0)}}(0) \\
\bY_{\calC_{(\xi_0,\eta_0)}}(0) & \bY_{\calD_{(\xi_0,\eta_0)}}(1)
\end{bmatrix}
\end{aligned}
\label{eq:staggered_design}
\end{equation}
}
\vspace{-2ex}

where 
$\calA_{(\xi_0,\eta_0)} = \{(\xi,\eta): 1 \leq \xi \leq k_1, 1 \leq \eta \leq k_2 \}$, $\calB_{(\xi_0,\eta_0)} = \{(\xi,\eta):1\leq \xi \leq k_1, k_2+1 \leq \eta \leq \eta_0 \}$, $\calC_{(\xi_0,\eta_0)} = \{(\xi,\eta): k_1+1 \leq  \xi \leq  \xi_0, 1 \leq j \leq k_2 \}$, and $\calD_{(\xi_0,\eta_0)} = \{(\xi,\eta): k_1+1 \leq  \xi \leq  \xi_0, k_2+1 \leq j \leq \eta_0 \}$
with  \( k_1 = r + 1 - \eta_0 \) and \( k_2 = r + 1 - \xi_0 \).
The associated treatment indicator matrix $\bOmega^{(\xi_0,\eta_0)}$ is defined analogously and the associated covariate matrix $\bX^{(\xi_0)}$ can be obtained by retaining the units included in $\bY^{(\xi_0,\eta_0)}$.
The missing potential outcomes $\bY_{\calD_{(\xi_0,\eta_0)}}(0)$ can be directly estimated using Algorithm \ref{alg:four-block}. 
By construction, $\bY_{\calD_{(\xi_0,\eta_0)}} \supseteq \bY_{(\xi_0,\eta_0)}$. In the case when $\bY_{\calD_{(\xi_0,\eta_0)}} \supset \bY_{(\xi_0,\eta_0)}$, we treat all the blocks of $\bY_{\calD_{(\xi_0,\eta_0)}}$ as missing and impute their values, but only extract $\widehat{\bY}_{(\xi_0,\eta_0)}(0)$ from the imputed $\widehat{\bY}_{\calD_{(\xi_0,\eta_0)}}(0)$; see Figure \ref{fig:eg-construction-of-4-block-design1} for an example. %

\begin{figure}[!tbh]
\centering
\resizebox{0.85\linewidth}{!}{

\begin{tabular}{|c|c|c|c||c||c|}\hline

 & $T_1$ & $T_2$ & $T_3$ & $T_4$ & $T_5$ \\ \hline \hline 

$N_1$ & \cellcolor{green!20}$\bY_{(1,1)}$ & \cellcolor{green!20}$\bY_{(1,2)}$& \cellcolor{green!20}$\bY_{(1,3)}$& \cellcolor{red!15}$\bY_{(1,4)}$ & $\bY_{(1,5)}$\\
\hline
$N_2$ & \cellcolor{green!20}$\bY_{(2,1)}$& \cellcolor{green!20}$\bY_{(2,2)}$& \cellcolor{green!20}$\bY_{(2,3)}$& \cellcolor{red!15}$\bY_{(2,4)}$&  \\
\hline \hline
$N_3$ & \cellcolor{blue!15}$\bY_{(3,1)}$ & \cellcolor{blue!15}$\bY_{(3,2)}$& \cellcolor{blue!15}$\bY_{(3,3)}$& \cellcolor{orange!30}?& \\
\hline \hline
$N_4$ &$\bY_{(4,1)}$& $\bY_{(4,2)}$&& & \\
\hline
$N_5$ & $\bY_{(5,1)}$& & & & \\

\hline
\end{tabular}
\quad 
\begin{tabular}{|c||c||c|c||c|c|} \hline 

 & $T_1$ & $T_2$ & $T_3$ & $T_4$ & $T_5$ \\ \hline \hline
 
$N_1$ & \cellcolor{green!20}$\bY_{(1,1)}$ & \cellcolor{red!15}$\bY_{(1,2)}$& \cellcolor{red!15}$\bY_{(1,3)}$ & $\bY_{(1,4)}$& $\bY_{(1,5)}$\\ \hline

$N_2$ & \cellcolor{green!20}$\bY_{(2,1)}$ & \cellcolor{red!15}$\bY_{(1,2)}$& \cellcolor{red!15}$\bY_{(2,3)}$ & $\bY_{(2,4)}$&  \\ \hline

$N_3$ & \cellcolor{green!20}$\bY_{(3,1)}$& \cellcolor{red!15}$\bY_{(3,2)}$& \cellcolor{red!15}$\bY_{(3,3)}$& & \\ \hline \hline

$N_4$ & \cellcolor{blue!15}$\bY_{(4,1)}$ & \cellcolor{orange!30}$\bY_{(4,2)}$& \cellcolor{orange!30}& & \\ \hline

$N_5$ & \cellcolor{blue!15}$\bY_{(5,1)}$& \cellcolor{orange!30}& \cellcolor{orange!30}?& & \\ \hline

\end{tabular}
}
\caption{\small Examples of constructing a four-block submatrix \(\bY^{(\xi_0,\eta_0)}\) to estimate \(\bY_{(\xi_0,\eta_0)}(0)\). Here, \( r=5 \). The question mark denotes the block of interest, with $(\xi_0,\eta_0)$ = (3,4) (Left) and (5,3) (Right). The associated four blocks are designed by green (\(\bY_{\calA_{(\xi_0,\eta_0)}}\)), red (\(\bY_{\calB_{(\xi_0,\eta_0)}}\)), blue (\(\bY_{\calC_{(\xi_0,\eta_0)}} \)), and orange (\(\bY_{\calD_{(\xi_0,\eta_0)}}\)) as in \eqref{eq:staggered_design}. 
}
\label{fig:eg-construction-of-4-block-design1}
\end{figure} 
\vspace{-1ex}
With the four-block construction in \eqref{eq:staggered_design} and the CoDEAL method in four-block design introduced in Section \ref{subsec:CoDeal-Four-Block}, we obtain the imputed values of $\bY_{(\xi, \eta)}(0)$ for all $(\xi,\eta) \in \calI_{tr}$. The unit-specific ATT can then be estimated by $\widehat\tau_i = \sum_{t: W_{it} = 1} [Y_{it}(1)-\widehat{Y}_{it}(0)]/\sum_{t} W_{it}$ for each treated unit $i$. The complete algorithm of CoDEAL in the SAD is summarized as Algorithm \ref{alg:staggered}.
\vspace{-1ex}

\vspace{-1ex}
\subsection{Theoretical Properties} \label{subsec:theoretical-results} 
\vspace{-1ex}

This section establishes theoretical guarantees. We derive an explicit convergence rate for counterfactual recovery that characterizes the effects of nonparametric estimation, latent factor recovery, and finite-sample approximation. The results imply the estimation consistency of the CoDEAL estimator.

\begin{assumption}\label{asmp:density}
Suppose $\bX_i$ is supported on $[0,1]^P$ and has a density function $f$ satisfying
$\sup_{x \in [0,1]^P} f(x) =: f_{\max} < \infty,$
where $f_{\max}$ is independent of dimension $P$. We assume there exists a constant $B > 0$ such that $\E \bF_t^\star=0$ and $\|\bF_t^\star\|_\infty \leq B$ holds almost surely. We further assume that $\varepsilon_{it}$ is sub-Gaussian with sub-Gaussian norm $\sigma_\epsilon^2$, and that $\varepsilon_{it}, \bF_t^\star,\bX_i$ are mutually independent. 
\end{assumption}

\vspace{-1ex}
\begin{assumption}\label{asmp:smooth}
The functions $\{g_t^\star,\phi_i^\star\}$ are assumed to be $(\beta, C)$-H\"older smooth, that is, each function is $\lfloor \beta \rfloor$ times continuously differentiable, and its $\lfloor \beta \rfloor$th derivative is H\"older continuous with exponent $\beta - \lfloor \beta \rfloor$ and constant $C$. Here, $\lfloor \beta \rfloor$ denotes the greatest integer less than $\beta$. 
\end{assumption}

\begin{assumption}[Pervasiveness] \label{asmp:factor}
There exist a matrix $M^\star \in \mathbb{R}^{K \times N}$ and a function $\rho^\star$, defined on the image of the mapping $M^\star \phi^\star$, denoted by $M^\star \phi^\star([-B,B]^K)$. Assume that $\rho^\star $ is also $(\beta, C)$-H\"older smooth, and $\|M^\star\|_\infty \lesssim n^{-1}$, $\|M^\star\|_0 \asymp n$ for some diverging integer $n>0$. We further assume that
\vspace{-1ex}
{\small
\begin{equation}\label{eq:factor}
\sup_{x \in [-B,B]^K} \| \rho^\star(M^\star \phi^\star(x)) - x \|^2 + \|M^\star \|^2 \lesssim n^{-1}.
\end{equation}
}
\end{assumption}

\vspace{-2ex}
\begin{assumption}[Sequential exogeneity/no anticipation]\label{asmp:exogeneity}
Let $\mathcal{F}_t=\sigma(\bX,\{\bF_s^\star,\varepsilon_{\cdot s}\}_{s\le t})$. Assume
(i) $W_{it}$ is $\mathcal{F}_{t-1}$-measurable (adoption time is a stopping time);
(ii) $\varepsilon_{\cdot t}\perp\mathcal{F}_{t-1}$ and $\mathbb{E}\varepsilon_{it}=0$.
\end{assumption}
\vspace{-1ex}

Assumption \ref{asmp:density} is a standard assumption in the literature of nonparametric regression. Note that assuming $\bX_i$ has compact support is equivalent to assuming that the support is $[0,1]^P$ via centering and scaling. 
The smoothness condition is standard in the nonparametric regression literature (cf. \cite{tsybakov2009introduction}, Chapter 1), as it governs the complexity of the underlying function class. A formal definition is provided in Appendix \ref{section:proofs}.
Assumption \ref{asmp:factor} %
effectively enables approximate recovery of the latent factors from the observed input: when $x$ denotes the latent factors, condition \eqref{eq:factor} %
captures the tradeoff between approximation accuracy and model complexity, with $n$ reflecting the number of encoder parameters. By construction, it is monotonically non-increasing in $n$, measuring how gains in encoder expressiveness translate into improved factor recovery. The $n^{-1}$ rate could be generalized but is chosen for analytical simplicity. The constraint $\|M^\star\|_0\asymp n$ regulates the necessary model complexity. %
To illustrate, consider the linear factor model in \cite{bai2003}, where $\phi^\star(x) = A x$ with $\|A\|_\infty \lesssim 1$ and $\|A\|_0 \asymp N$ indicating that each factor affects nearly all observed variables. Assumption~\ref{asmp:factor} is satisfied by setting $M^\star = N^{-1} A^\top$ and $\rho^\star(x) = N (A^\top A)^{-1} x$ with $n = N$. For a nonlinear illustration, consider $\phi^\star(x)=\exp(Ax)$. Taking $M^\star=n^{-1} A^\top$, $M^\star\exp(Ax)$ has a positive definite Jacobian, ensuring local invertibility by the inverse function theorem. Hence there exists a smooth inverse $\rho^\star(M^\star \phi^\star(x))=x$, yielding the same $n^{-1}$ approximation error bound. 
Assumption \ref{asmp:exogeneity} allows adoption to depend on covariates $\bX_i$, the full latent factor path $\{\bF_s^\star\}_{s<t}$ (already absorbed by the model), and lagged outcomes and shocks, covering design-based rollouts and policies responding to realized history. Excluded is dependence on contemporaneous or future shocks.

The following theorem establishes the convergence of the estimated counterfactuals under a four-block design. The result can be readily extended to general SADs.
\begin{theorem}\label{thm1}
Suppose Assumptions \ref{asmp:density}-\ref{asmp:factor} hold, and let $K_1\ge K$, with $K_1$ defined in \eqref{eq:ae-class}. Assume that $N_1\asymp N$ and $T_1\asymp T_2$. Then, under a suitable choice of parameters for the DNN function class (specified in Appendix~\ref{section:proofs}), we have
\vspace{-1ex}
{\small
\begin{align*}
& \max_{N_1+1\le i \le N} \frac{1}{T_2} \sum\nolimits_{t=T_1+1}^{T} \mathbb{E} \left( \widehat Y_{it}(0)  - \phi_i^\star(\bF_{t}^\star) - g_t^\star(\bX_i) \right)^2   \nonumber = O_{\P} (\gamma),
\\ & \gamma=  \left( \frac{(N+n)T^{-\frac{2\beta}{2\beta + K}}}{n} + \frac{n}{T} + \frac{nK_1+N}{n^2} \right) \log^4(N T)  +N^{-\frac{2\beta}{2\beta + k}}  \log^5 N .
\end{align*}
}
\end{theorem}
\vspace{-2ex}
Theorem~\ref{thm1} establishes the statistical consistency of the counterfactual estimator and guarantees the algorithm's approximation quality.
Writing $r=\tfrac{2\beta}{2\beta+K}$, the four terms of $\gamma$ are: (i)~$\tfrac{N+n}{n}T^{-r}$, the Stone-optimal rate for regressing on the $K$-dimensional latent space with $T$ time points (learning the $N$ decoders), inflated by $\tfrac{N+n}{n}$ because factors are recovered through an $n$-parameter encoder; (ii)~$n/T$, encoder variance ($n$ effective parameters, $T$ columns); (iii)~$K_1/n+N/n^2$, factor-recovery bias from the $n^{-1}$ approximation in Assumption~\ref{asmp:factor} (equation \eqref{eq:factor}); (iv)~the Stone
rate $N^{-2\beta/(2\beta+P)}$ for $\widehat g_t$ on the $P$-dimensional covariate interactions with $N$ cross-sectional observations. Terms (i),(iii) decrease and (ii) increases in $n$: a bias--variance tradeoff in encoder capacity. The curse of dimensionality enters only through the fixed, small $K$ and $P$ (never through $N,T$), which is precisely the benefit of the factor structure.
\begin{rmk}[Linear Pervasive Factors]
Let $\phi^\star(x)=Ax$ with $\|A\|_\infty\lesssim1$ and subset pervasiveness $\lambda_{\min}(A_S^\top A_S)\asymp|S|$ (e.g., i.i.d.
loadings). Then Assumption~\ref{asmp:factor} holds for \emph{every} diverging $n\le N$ via $M^\star=n^{-1}A_S^\top$, $\rho^\star(z)=n(A_S^\top A_S)^{-1}z$, $|S|=n$ (exact inversion; $\|M^\star\|^2\lesssim n^{-1}$). Optimizing $n\asymp(NT)^{1/3}$
gives
$\gamma\asymp (N^{1/3}T^{-2/3}+N^{2/3}T^{-4/3})\log^4(NT)
+N^{-\frac{2\beta}{2\beta+P}}\log^5N$, which is consistent iff $N=o(T^2)$ (up to logs). For $N\asymp T$, we have
$\gamma\asymp T^{-1/3}\log^4T+T^{-\frac{2\beta}{2\beta+P}}\log^5T$.    
\end{rmk}

Writing $\mathcal{D}_i=\{t: W_{it}=1\}$, the unit-specific ATT estimation error satisfies
\[
\widehat\tau_i-\tau_i^\star
=\tfrac{1}{T_2}\textstyle\sum_{t\in\mathcal{D}_i}\varepsilon_{it} + \tfrac{1}{T_2}\textstyle\sum_{t\in\mathcal{D}_i}\big[\phi_i^\star(\bF_t^\star)+g_t^\star(\bX_i)-\widehat Y_{it}(0)\big].
\]
By Assumption~\ref{asmp:exogeneity}, $\{W_{it}\varepsilon_{it},\mathcal{F}_t\}$ is a martingale difference
sequence, so $\mathbb{E}[\varepsilon_{it}\mid W_{it}=1]=0$ holds as a consequence. By Theorem~\ref{thm1} and the condition $\gamma=o(1)$, Azuma's inequality gives the uniform bound
$\max_i|\widehat\tau_i-\tau_i^\star|
=O_{\mathbb P}(\sqrt{\log N/T_2}+\sqrt\gamma)$,
which establishes consistency of the unit-specific ATT estimator.
Proofs and more technical details are in Appendix~\ref{section:proofs}. 

\begin{rmk}[Extension to Time-Varying Treatment Effects] \label{rmk:time-varying-treatment}
In principle, CoDEAL can produce estimates of unit-time-specific treatment effect $\tau_{it}$ using $\widehat{\tau}_{it} = Y_{it}(1) - \widehat{Y}_{it} (0)$. 
  The static-effect assumption enters only the interpretation of the averaged estimator $\widehat{\bar\tau}_i := \ T_2^{-1}\sum_{t:W_{it}=1}\widehat{\tau}_{it}$, not the imputation machinery. CoDEAL's counterfactual imputation $\widehat{Y}_{it}(0)$ is trained exclusively on untreated outcomes, so Theorem~\ref{thm1}'s guarantee on 
  $
  \max_{N_1+1\le i\le N}\ \frac{1}{T_2}\sum_{t=T_1+1}^{T}
  \mathbb{E} (\widehat{Y}_{it}(0)-\phi_i^\star(\mathbf{F}_t^\star)-g_t^\star(\mathbf{X}_i) )^2 
  $
  is entirely unaffected by whether the treatment effect is static or dynamic.
  When the treatment effect is static ($\tau_{it} \equiv \tau_i$), 
  this result establishes the estimation consistency of $\widehat{\bar\tau}_i$ for $\tau_i$. 
  When the treatment effect $\tau_{it}$ varies over time, Theorem \ref{thm1} implies that $\widehat{\bar\tau}_i$ consistently estimates the unit-wise average treatment effect  $\bar{\tau}_i=T_2^{-1}\sum_{t:W_{it}=1}\tau_{it}$. 
  Thus, CoDEAL permits the contemporaneous treatment effect to vary across time and units, while preserving consistency for the average effect.
  A pointwise guarantee of $\widehat{\tau}_{it}$ for ${\tau}_{it}$, however, is not attainable by the current theory.
\end{rmk}

\vspace{-2ex}

\section{Simulation Studies} 
\vspace{-2ex}
\label{sec:simulation}
We use simulation to demonstrate the finite-sample performance of CoDEAL\footnote{The code for implementation is available at \texttt{https://github.com/chenezandie/CoDEAL-Code}. }. %
Complete simulation studies are provided in Appendix \ref{append:simulation}.
We consider various benchmark methods, including a single-output AE (\textbf{Single AE}) (details in Appendix \ref{append:simu-benchmark}), neuron-enhanced AE (\textbf{AEMC-NE}) \citep{fan2024neuron}, matrix completion with nuclear norm minimization (\textbf{MC-NNM}) \citep{athey2021matrix}, vertical regression (\textbf{Vert-Reg}) \citep{doudchenko2016balancing}, difference-in-differences (\textbf{DiD}) \citep{imai2021use}, and kernel principal component analysis (\textbf{KernelPCA}) \citep{scholkopf1997kernel}. %
We also compare to three synthetic-control-style baselines, including the classic synthetic control \citep{abadie2010synthetic}, generalized synthetic control \citep{xu2017generalized}, and synthetic difference-in-differences \citep{arkhangelsky2021synthetic}, with results reported in the Appendix.
We evaluate %
using mean squared error (MSE) and mean absolute error (MAE). %

\noindent \textbf{Data Generating Process and Simulation Configurations} (with details in Appendix \ref{append:simulation-DGP}). We carry out extensive simulations under various linear and nonlinear generating mechanisms (4 variants for generating factor effects and 6 variants for generating covariate effects), with different setups of unit numbers $N$, time periods $T$, latent factor dimension $K$, %
  and treatment assignments. %
We consider configurations 1-4: \textbf{Four-Block Design}: (1) $(N,T,N_1,T_1,P,K)=$ $(100,200,50,100,3,4)$, (2) $(N,T,N_1,T_1,P,K)=$ $(200,120,100,60,5,3)$, \textbf{SAD}: (3) $(N,T,K)=(100,200,4)$, $r=\{5,10\}$, and (4) $(N,T,K)=(200,120,3)$, $r=\{5,10\}$. 
Hyperparameter settings and sensitivity analyses are presented in Appendix~\ref{appen: hyper}. CoDEAL is robust to the learning rate and network depth over a wide range; the bottleneck size $K_1$ is the one choice that matters materially, and under-specifying it below the true factor dimension is far more damaging than over-specifying it. %
The hyperparameters may also be selected empirically via a masked-block validation strategy, which we detail in Remark~\ref{rmk:masked-block}.%

\noindent \textbf{CoDEAL in the removal of covariate effects} (with details in Appendix \ref{append:simu-covariate-removal}). We compare the performance of CoDEAL using different methods, specifically, DNN and linear regression (LR), to remove the covariate effects, against an oracle benchmark when no covariate effects are present. Six scenarios of covariate effects were considered. The results show that LR is
adequate for linear covariate effects but the DNN approach is essential in the presence of nonlinear covariate effects. %

\vspace{-0.2ex}

\noindent \textbf{CoDEAL in causal matrix completion} (with details in Appendix \ref{append:simu-CoDEAL-performance}). As shown in Table \ref{tab:simu-CoDeal-MAE}, CoDEAL consistently yields the lowest MAE across nearly all settings, and achieves substantial improvements relative to all competing methods. %
In SAD with {linear} factors, CoDEAL \textbf{{ties MC-NNM within 3\% while reducing MAE by 35–89\% over other benchmarks}}. In SAD with nonlinear factors, CoDEAL delivers evident advantages. Especially when under {ReLU}-based factors, CoDEAL delivers its largest margins, with \textbf{{MAE reductions of 15–60\% versus all competing approaches}}. As $r$ increases from 5 to 10, the missingness pattern grows more complex and extensive, so all methods show some drop in accuracy. CoDEAL’s errors only suffer a slight increase, whereas simpler methods like Vert-Reg show a dramatic rise in MAE when missingness increases.

\vspace{-0.2ex}
\noindent \textbf{In summary}, CoDEAL achieves performance comparable to MC-NNM in linear settings, while exhibiting substantial improvements over all considered methods in the presence of nonlinear effects.
By integrating DNN-based covariate adjustments with deep multi-output-AE–based nonlinear factor analysis, CoDEAL is robustly effective across a wide range of linear and nonlinear scenarios.

\begin{table*}[!t]
\small
\centering
\caption{\small Comparisons of prediction accuracy (measured by MAE, averaged over 50 replications) %
in Config.1–4. 
Results of MSE are reported in Table~\ref{tab:simu-CoDeal} of Appendix~\ref{append:simulation}. %
Numbers in parentheses are standard errors. 
}
\vspace{-2ex}
\resizebox{\textwidth}{!}{
\begin{tabular}{llcccccccc}
\toprule
\multirow{2}{*}{\textbf{Factor}} & \multirow{2}{*}{\textbf{Method}} 
& \multicolumn{2}{c}{Config. 1} 
& \multicolumn{2}{c}{Config. 2} 
& \multicolumn{2}{c}{Config. 3} 
& \multicolumn{2}{c}{Config. 4} \\ 
& & Linear & Nonlinear & Linear & Nonlinear & $r=5$ & $r=10$ & $r=5$ & $r=10$ \\
\midrule

\multirow{7}{*}{\textbf{Linear}}
& CoDEAL      & \textbf{0.646(0.009)} & \textbf{0.701(0.008)} & \textbf{0.609(0.004)} & \textbf{0.644(0.005)} & 0.614(0.007) & 0.625(0.005) & \textbf{0.532(0.003)} & \textbf{0.560(0.004)} \\
& Single AE   & 0.821(0.010) & 0.851(0.008) & 0.777(0.007) & 0.795(0.007) & 0.789(0.011) & 0.795(0.008) & 0.693(0.005) & 0.714(0.006) \\
& AEMC-NE     & 0.838(0.009) & 0.862(0.009) & 0.759(0.007) & 0.771(0.007) & 0.731(0.010) & 0.730(0.007) & 0.647(0.005) & 0.660(0.005) \\
& DiD         & 0.892(0.010) & 0.916(0.009) & 0.810(0.007) & 0.828(0.008) & 0.801(0.011) & 0.802(0.008) & 0.711(0.005) & 0.726(0.006) \\
& Vert-Reg    & 0.757(0.006) & 0.798(0.007) & 0.761(0.004) & 0.770(0.005) & 0.940(0.005) & 0.865(0.023) & 0.691(0.003) & 0.750(0.016) \\
& MC-NNM      & 0.699(0.006) & 0.735(0.005) & 0.650(0.005) & 0.697(0.006) & \textbf{0.599(0.007)} & \textbf{0.616(0.006)} & 0.537(0.003) & 0.565(0.004) \\
& KernelPCA   & 0.811(0.007) & 0.824(0.008) & 0.835(0.009) & 0.849(0.009) & 0.798(0.007) & 0.801(0.007) & 0.707(0.005) & 0.709(0.005) \\
\midrule
 
\multirow{7}{*}{\textbf{Sine}}
& CoDEAL      & \textbf{1.078(0.012)} & \textbf{1.117(0.011)} & \textbf{0.894(0.007)} & \textbf{0.945(0.008)} & \textbf{0.973(0.010)} & \textbf{0.996(0.010)} & \textbf{0.850(0.007)} & \textbf{0.869(0.008)} \\
& Single AE   & 1.223(0.013) & 1.250(0.012) & 1.142(0.009) & 1.185(0.009) & 1.087(0.011) & 1.122(0.011) & 1.057(0.008) & 1.061(0.008) \\
& AEMC-NE     & 1.209(0.013) & 1.235(0.012) & 1.128(0.009) & 1.167(0.010) & 1.055(0.012) & 1.078(0.011) & 1.011(0.008) & 1.008(0.009) \\
& DiD         & 1.244(0.013) & 1.272(0.012) & 1.191(0.009) & 1.234(0.009) & 1.110(0.012) & 1.146(0.012) & 1.091(0.008) & 1.088(0.009) \\
& Vert-Reg    & 1.169(0.012) & 1.208(0.011) & 1.012(0.007) & 1.054(0.008) & 1.549(0.090) & 1.524(0.123) & 0.974(0.009) & 1.047(0.013) \\
& MC-NNM      & 1.201(0.013) & 1.231(0.012) & 1.105(0.009) & 1.153(0.010) & 1.066(0.011) & 1.097(0.011) & 0.998(0.008) & 1.003(0.008) \\
& KernelPCA   & 1.195(0.009) & 1.231(0.009) & 1.170(0.013) & 1.190(0.013) & 1.109(0.012) & 1.115(0.012) & 1.075(0.009) & 1.079(0.009) \\
\midrule
 
\multirow{7}{*}{\textbf{Polynomial}}
& CoDEAL      & \textbf{0.908(0.013)} & \textbf{0.952(0.015)} & \textbf{0.772(0.010)} & \textbf{0.807(0.011)} & \textbf{0.785(0.011)} & \textbf{0.811(0.010)} & \textbf{0.662(0.007)} & \textbf{0.699(0.007)} \\
& Single AE   & 0.962(0.013) & 0.997(0.016) & 0.820(0.010) & 0.854(0.011) & 0.895(0.015) & 0.926(0.016) & 0.739(0.011) & 0.785(0.010) \\
& AEMC-NE     & 0.997(0.014) & 1.038(0.017) & 0.856(0.011) & 0.887(0.012) & 0.860(0.014) & 0.897(0.013) & 0.720(0.009) & 0.764(0.009) \\
& DiD         & 0.995(0.013) & 1.025(0.016) & 0.854(0.010) & 0.886(0.011) & 0.887(0.013) & 0.917(0.014) & 0.739(0.009) & 0.773(0.008) \\
& Vert-Reg    & 0.953(0.013) & 1.019(0.016) & 0.855(0.007) & 0.884(0.010) & 1.155(0.035) & 1.131(0.032) & 0.788(0.005) & 0.881(0.014) \\
& MC-NNM      & 0.951(0.013) & 0.989(0.015) & 0.819(0.009) & 0.851(0.011) & 0.839(0.012) & 0.865(0.012) & 0.703(0.008) & 0.735(0.007) \\
& KernelPCA   & 0.927(0.013) & 0.991(0.013) & 1.029(0.015) & 1.051(0.015) & 0.935(0.011) & 0.944(0.011) & 0.776(0.008) & 0.782(0.008) \\
\midrule
 
\multirow{7}{*}{\textbf{ReLU MLP}}
& CoDEAL      & \textbf{0.797(0.009)} & \textbf{0.838(0.010)} & \textbf{0.698(0.005)} & \textbf{0.743(0.006)} & \textbf{0.632(0.004)} & \textbf{0.733(0.004)} & \textbf{0.633(0.004)} & \textbf{0.646(0.004)} \\
& Single AE   & 0.989(0.009) & 1.026(0.011) & 0.891(0.007) & 0.930(0.008) & 0.876(0.005) & 0.913(0.006) & 0.797(0.005) & 0.809(0.005) \\
& AEMC-NE     & 1.045(0.010) & 1.088(0.011) & 0.959(0.007) & 0.982(0.008) & 0.934(0.006) & 0.968(0.007) & 0.860(0.006) & 0.878(0.006) \\
& DiD         & 0.930(0.008) & 0.963(0.010) & 0.842(0.005) & 0.870(0.007) & 0.817(0.004) & 0.842(0.004) & 0.742(0.004) & 0.748(0.005) \\
& Vert-Reg    & 0.888(0.010) & 0.958(0.013) & 0.847(0.005) & 0.894(0.006) & 1.045(0.031) & 0.948(0.026) & 0.755(0.003) & 1.070(0.253) \\
& MC-NNM      & 0.843(0.007) & 0.892(0.009) & 0.767(0.005) & 0.806(0.006) & 0.721(0.003) & 0.750(0.003) & 0.656(0.004) & 0.662(0.004) \\
& KernelPCA   & 1.105(0.008) & 1.126(0.009) & 1.156(0.011) & 1.172(0.011) & 1.111(0.008) & 1.119(0.008) & 1.005(0.006) & 1.009(0.006) \\
\bottomrule
\end{tabular}
}
\label{tab:simu-CoDeal-MAE}
\end{table*}

\vspace{-2.2ex}
\section{Real Data Applications} 
\label{sec:realdata}
\vspace{-2ex}
We illustrate our methodology using data from the Oxford COVID-19 Government Response Tracker (OxCGRT) \citep{Hale2021}. %
OxCGRT collects panel data on the timing and intensity of 24 different government interventions, such as school closures, travel restrictions, and vaccination mandates, across over 185 countries from January 2020 through 2022. We report two additional case studies in Appendix~\ref{append:additional-real-data}, using the US Election Day Registration dataset \citep{xu2017generalized} and the German reunification dataset \citep{pang2022bayesian}.

\vspace{-0.5ex}

\begin{figure*}[ht]
\centering
\includegraphics[width=0.88\textwidth]{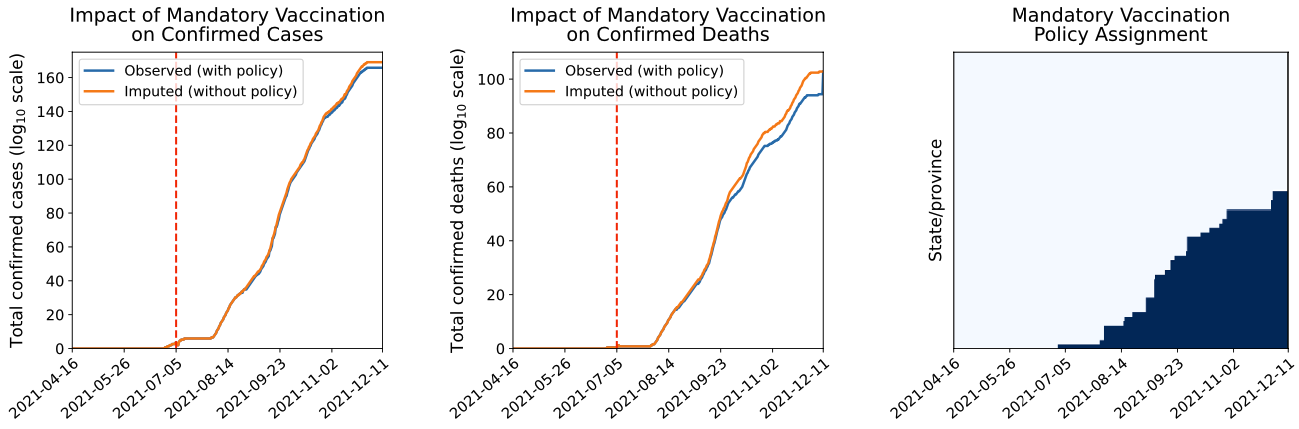}
\vspace{-1ex}
    \caption{%
    \small
    Left and Middle: Comparison of the total confirmed cases (Left) and deaths (Middle) across policy-in-effect states between observed results with policy executed and estimated counterfactuals if no policy. %
    Values are reported on a ${\log_{10}}$ scale, and plots are visualized with 14-day moving averages. 
    Right: Indicator matrix of observed policy assignment, %
    with darker color referring to policy in effect. Enlarged figures are in Appendix \ref{append: covid19}.} 
    \label{fig:MV1}
\end{figure*}

We examine the impact of two specific policy interventions:
(1) the \textbf{implementation of mandatory vaccination}, and (2) the \textbf{termination of internal travel restrictions} (results in Appendix \ref{append: covid19}). Our analysis covers their impacts on COVID-19 confirmed cases and deaths across 64 states and territories in the United States and Canada from April 16, 2021, to December 11, 2021. This time frame aligns with the period following the detection of the Delta variant and precedes the rapid emergence of the Omicron variant in North America, after vaccines became widely available; restricting to this window limits major regime shifts in epidemic dynamics during the analysis. We use %
four OxCGRT indices (overall government response, containment and health, stringency, and economic support), averaged over the time frame,  as covariates, resulting in data dimensions of $N=64$, $T=240$, and $P=4$. 
The policy is adopted in different states at various times, forming a SAD (as visualized in Figure \ref{fig:MV1}). 

\vspace{-1.5ex}

Figure \ref{fig:MV1} plots the comparisons of the total confirmed cases and deaths across policy-in-effect states between observed results (with policy executed) and estimated counterfactuals (if no policy). %
Results of other benchmarks besides CoDEAL are in the appendix.
These results represent estimated counterfactual trajectories rather than directly verifiable causal effects, which cannot be directly verified due to the unobservability of counterfactual outcomes in real-world settings. The estimated trajectories allow for time-varying contemporaneous policy effects, while the estimated treatment effects are interpreted as averages over the post-adoption period, as discussed in Remark \ref{rmk:time-varying-treatment}.
As shown in the plot, if the mandatory vaccination policy were not implemented, there would be an increase in both confirmed cases and deaths, with a notably larger impact on mortality. This observation aligns with recent research highlighting a greater reduction in mortality than in infection rates following vaccination \citep{hernandez2023impact,xu2023systematic,mesle2024estimated,wu2023long}. Although the visual differences in the log-transformed graphs appear modest, converting these back to their original scales reveals substantial impacts: on December 11, 2021, the actual cumulative counts of confirmed cases and deaths for states with mandatory vaccination policies were ${26.57\text{ }million, 0.39\text{ }million}$, respectively. In contrast, our estimates suggest that without these policies, the counts would have risen to ${29.81\text{ }million(\uparrow 12.2\%), 0.52\text{ }million(\uparrow 34.6\%)}$. 

\vspace{-2ex}
\section{Discussion of Limitations and Extensions} \label{sec:discussion}
\vspace{-2ex}

We propose CoDEAL, a deep-learning-based causal matrix completion method and a unifying framework for estimating heterogeneous causal effects in panel data models. The proposed framework largely extends existing causal panel literature in the capacity of accommodating heterogeneity and nonlinearity. However, the extent to which these features are present in empirical applications is generally unknown. To guide practical implementation, Remark~\ref{rmk:masked-block} discusses a masked-block reconstruction strategy that enables data-driven comparison of model specifications and hyperparameter selection using observed untreated outcomes, given that the true counterfactual outcomes are inherently unobservable.
We also clarify that several assumptions (e.g., time-invariant covariates, static treatment effects) are not intrinsic to the algorithm and can be relaxed to accommodate more general settings. Moreover, despite the additive model formulation, the proposed framework does not preclude interactions between covariates and latent factors, nor does it require them to be uncorrelated. 
The CoDEAL framework is amenable to further adaptation across data-driven contexts. Detailed discussions of these extensions are provided in Appendix~\ref{append:discussion}. 

CoDEAL fundamentally differs from standard MAR-assumed matrix completion methods. Counterfactual missingness is systematically tied to treatment assignment. CoDEAL addresses the missing-not-at-random scenario explicitly through (1) a dedicated block-partition-and-learning algorithm that systematically leverages before-/after- treatment observations, and (2) a nonlinear factor model that captures the inherent dependence for imputation. The decomposition into $\phi_i^\star()$ and $\bF_t^\star$ allows us to account for the underlying cross-sectional and temporal dependencies driving the outcome.

Methodologically, with different choices of factor models and advanced neural network architectures, the proposed CoDEAL framework can be further generalized to accommodate various data types with more complex data structures, such as tensors, images, and networks. For example, CoDEAL can be naturally extended to three-dimensional or higher-order tensors by involving tensor factorization methodologies \citep{han2022tensor, chen2026estimation, zhou2025factor,chen2025diffusion}, providing a more general approach compared to matrix factor analysis \citep{luo2022inverse,yu2022nonparametric,chen2025modewise,yu2024dynamic}. That being said, to extend to multiple treatments, how to handle the increased level of missingness brought by the treatment assignments in a higher-dimensional tensor remains challenging and requires careful modeling strategies. We leave the multiple treatment settings as future work.

\begin{ack}
We thank the anonymous reviewers for helpful suggestions. The work of Yuefeng Han is supported in part by National Science Foundation Grants DMS-2412578.  The work of Guanhao Zhou and Xiufan Yu is supported in part by National Institutes of Health grant R01GM152812. 
\end{ack}

{%
\bibliographystyle{plainnat}
\bibliography{ref}
 }

\newpage
\appendix

\setcounter{table}{0}
\setcounter{figure}{0}
\setcounter{algorithm}{0}
\setcounter{equation}{0}
\setcounter{rmk}{0}
\renewcommand{\thefigure}{S.\arabic{figure}}
\renewcommand{\thetable}{S.\arabic{table}}
\renewcommand{\thealgorithm}{S.\arabic{algorithm}}
\renewcommand{\theequation}{S.\arabic{equation}}
\renewcommand{\thermk}{S.\arabic{rmk}}

\section*{\Large Appendices}
The Appendices of this paper contain five parts. 
Appendix \ref{append:tech-notes} includes additional technical notes for clearer exposition on the methodological framework, including an introduction to staggered adoption design along with detailed illustrative examples, an illustrative DAG of the proposed model, as well as discussions on the rationale of adopting a sequential estimation regime. 
Appendix \ref{append:simulation} presents a complete report, along with full details, of the simulation studies. 
Appendix \ref{append: real} provides additional results of the real data analysis using the Oxford COVID-19 Government Response Tracker (OxCGRT), and two additional empirical studies.
Appendix \ref{append:discussion} provides further discussions on broader impacts and directions of extensions. 
Appendix \ref{section:proofs} provides additional theoretical properties and complete proofs of the theoretical guarantees of the convergence of CoDEAL.

\section{Technical Notes on Methodological Framework} 
\label{append:tech-notes}

\subsection{Staggered Adoption Design} \label{append:staggered-design}
In this section, we introduce the concept of staggered adoption design \citep{athey2022design}, which is a commonly encountered framework in causal panel data analysis \citep{athey2021matrix,shaikh2021randomization,wing2024designing,choi2024matrix,yan2024entrywise}.

\subsubsection{Four-Block Design and Staggered Adoption Design}

Throughout this section, we assume the treatment adoption is irreversible. That is, once the treatment is initiated for a unit, it remains exposed to the treatment afterward. 

\begin{figure}[h]
\centering
\includegraphics[width=0.5\textwidth]{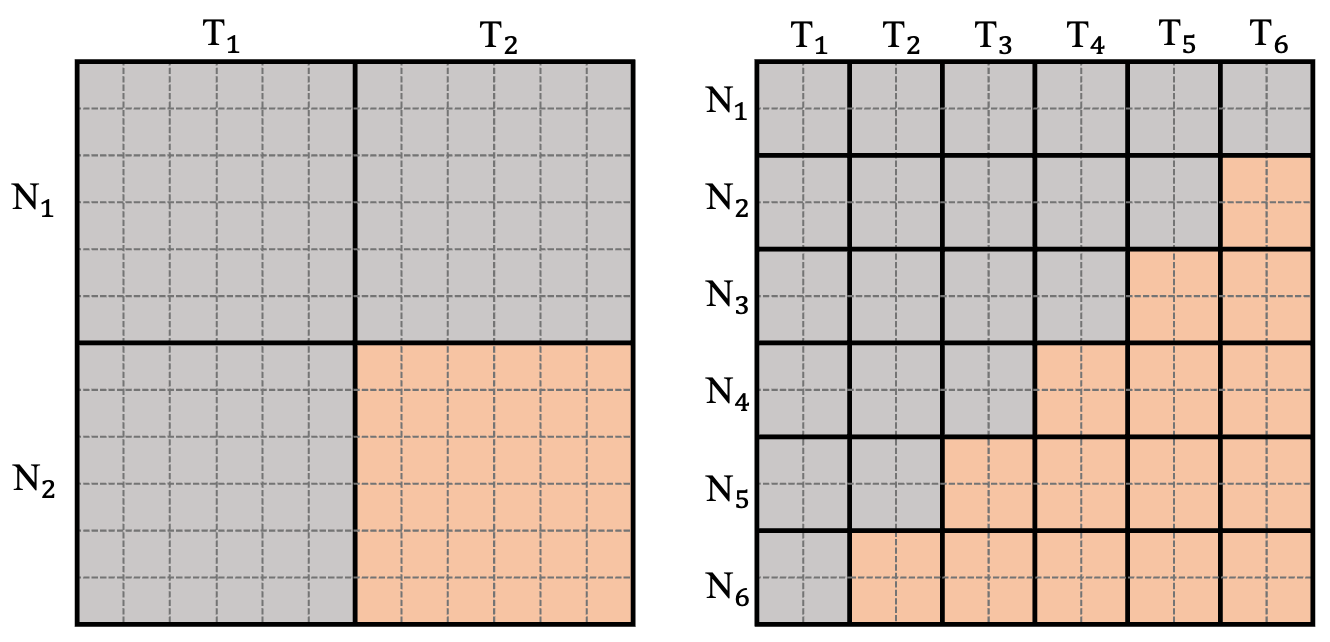}
\caption{Examples that illustrate the four-block design (Left) and staggered adoption design with $r=6$ (Right). Rows stand for units and columns stand for time. Orange blocks stand for treated.}
\label{fig:fbsad}
\end{figure}

We begin with the simplest design of treatment assignment in causal panel data analysis, when there is one control group and one treatment group. All units in the control group are never exposed to the treatment, and all units in the treatment group begin receiving the treatment at the same time. This design is considered the building block of more complex designs in causal panel models, often called the Four-Block design or Differences-in-Differences design in the literature. A graphical representation of the four-block design is plotted in the left panel of Figure \ref{fig:fbsad}.

Staggered adoption design accommodates more complex treatment assignments by allowing different units to get exposed to treatment at varying points in time. 
The four-block design can be regarded as a special case of $2\times 2$ staggered adoption. A graphical representation of the general staggered design is plotted in the right panel of Figure \ref{fig:fbsad}. Without loss of generality, we can arrange the panel units  by sorting the units based on the time at which they began receiving the treatment in a descending or ascending order, forming a block-wise missing pattern in the matrices of potential outcomes; see Section \ref{append:example_staggered_designs}
 for an illustrative example.

Figure \ref{fig:fbsad} visually illustrates the distinction between the four-block and staggered adoption designs. 
Each row in these panels represents an individual observational unit, while each column indicates a distinct time period. Orange blocks indicate periods during which units have adopted treatment, thus resulting in missing counterfactual control outcomes. Gray blocks represent observable periods under control conditions.
The left panel presents the four-block design, in which one group of units simultaneously transitions from untreated (gray cells) to treated status (orange cells) at a single shared time point. The right panel depicts the staggered adoption scenario, characterized by multiple adoption times and varying durations of observed control outcomes.

\subsubsection{Illustrative Example of Staggered Adoption Designs}\label{append:example_staggered_designs}

We use an explicit example to clearly illustrate the block-wise missing patterns of staggered adoption structure. 
In a staggered adoption design, various units receive irreversible treatment at different times. 
Without loss of generality, we arrange the panel by sorting the units based on the time at which they began receiving the treatment in a descending order (with never-treated units at the top, followed by the latest adopters, and then progressively earlier adopters towards the bottom).

We consider the following example with $N=6$ units over $T=5$ time periods. For treatment adoption designs, we suppose 
\begin{itemize}[leftmargin = 2ex]
\item \textbf{Unit 1 (Never adopter):} never receives treatment during the study periods, so all control outcomes are observed throughout the study.
\item \textbf{Unit 2 (Late adopter):} adopts treatment towards the end of the observation period, resulting in observed control outcomes at earlier time points.
\item \textbf{Unit 3 and Unit 4 (Medium adopters):} adopt treatment midway through the study, leading to missing control outcomes for periods following the adoption.
\item \textbf{Unit 5 and Unit 6 (Early adopters):} adopt treatment early in the observation period, causing most subsequent outcomes to become unobserved control outcomes.
\end{itemize}

The matrix of control outcome $\bY(0)$ is shown in Equation \eqref{eq:staggered-example}, where question marks denote the entries whose control outcomes are missing as this unit is under treatment at the time. The corresponding matrix of treatment indicators and its block-wise representation are presented in Equation \eqref{eq:staggered-example-omega}, with 1 for treated and 0 otherwise.

As revealed by Equations \eqref{eq:staggered-example} and \eqref{eq:staggered-example-omega}, there is naturally a block structure in the treatment indicator matrix $\bOmega$, or equivalently speaking, a block-wise missing pattern in the matrix of control outcome $\bY(0)$. Though it is possible to impute the missing elements one by one, it is generally computationally expensive, especially if either $T$ or $N$ is large. For efficient imputation, as illustrated in Section \ref{subsec:CoDeal-Staggered-Design}, we partition the indicator matrix (as well as the control outcome matrix) into submatrices 
$\bOmega_{(\xi, \eta)} \in \mathbb{R}^{N_{\xi} \times T_{\eta}}$ and $\bY_{(\xi, \eta)} \in \mathbb{R}^{N_{\xi} \times T_{\eta}}$ for $\xi=1,\ldots, r$ and $\eta =1,\ldots, r$, where $\sum_{\xi=1}^{r} N_{\xi} =N $ and $\sum_{\eta=1}^{r} T_{\eta} = T $. The matrix imputation can subsequently be conducted using the block structured submatrices, following the procedures introduced in Algorithm \ref{alg:staggered} of Section \ref{subsec:CoDeal-Staggered-Design}. In this example, details of the partitions are provided in Equation \eqref{eq:staggered-example-block}, resulting in $r$ taking the value of 4.

\begin{align}
\bY(0) & =\left[
\begin{array}{cccccc}
Y_{11} &Y_{12} &Y_{13} &Y_{14} & Y_{15} &\text{never adopter}\\
Y_{21} &Y_{22} &Y_{23} &Y_{24} & ? & \text{late adopter}\\
Y_{31} &Y_{32} &? & ? & ? & \text{medium adopter}\\
Y_{41} &Y_{42} &? & ? & ? & \text{medium adopter}\\
Y_{51} & ? & ? & ? & ? & \text{early adopter}\\
Y_{61} & ? & ? & ? & ? & \text{early adopter}
\end{array}
\right] \label{eq:staggered-example}
\\ \medskip 
\label{eq:staggered-example-block}
& =\left[
\begin{array}{c||c||cc||c}
Y_{11} &Y_{12} &Y_{13} &Y_{14} & Y_{15} \\ \hline\hline
Y_{21} &Y_{22} &Y_{23} &Y_{24} & ? \\ \hline\hline
Y_{31} &Y_{32} &? & ? & ? \\
Y_{41} &Y_{42} &? & ? & ? \\\hline\hline
Y_{51} & ? & ? & ? & ? \\
Y_{61} & ? & ? & ? & ? 
\end{array}
\right] 
: = 
\left[
\begin{array}{c||c||c||c}
\bY_{(1,1)} & \bY_{(1,2)} & \bY_{(1,3)} & \bY_{(1,4)}  \\ \hline\hline
\bY_{(2,1)} & \bY_{(2,2)} & \bY_{(2,3)} & \textbf{?}   \\ \hline\hline
\bY_{(3,1)} & \bY_{(3,2)} & \textbf{?} & \textbf{?} \\\hline\hline
\bY_{(4,1)} & \textbf{?} & \textbf{?} & \textbf{?}  \\
\end{array}
\right] \\ \medskip
\label{eq:staggered-example-omega} 
\text{and } \
\boldsymbol{\Omega} & =\left[
\begin{array}{ccccc}
0 & 0 & 0 & 0 & 0 \\
0 & 0 & 0 & 0 & 1 \\
0 & 0 & 1 & 1 & 1 \\
0 & 0 & 1 & 1 & 1 \\
0 & 1 & 1 & 1 & 1 \\
0 & 1 & 1 & 1 & 1 
\end{array}
\right] 
= \left[
\begin{array}{c||c||cc||c}
0 & 0 & 0 & 0 & 0 \\\hline\hline
0 & 0 & 0 & 0 & 1 \\\hline\hline
0 & 0 & 1 & 1 & 1 \\
0 & 0 & 1 & 1 & 1 \\\hline\hline
0 & 1 & 1 & 1 & 1 \\
0 & 1 & 1 & 1 & 1 
\end{array}
\right] 
= 
\left[
\begin{array}{c||c||c||c}
\textbf{0} & \textbf{0} & \textbf{0} & \textbf{0} \\\hline\hline
\textbf{0} & \textbf{0} & \textbf{0} & \textbf{1} \\\hline\hline
\textbf{0} & \textbf{0} & \textbf{1} & \textbf{1} \\\hline\hline
\textbf{0} & \textbf{1} & \textbf{1} & \textbf{1} \\
\end{array}
\right] 
\end{align}

\subsection{An Illustrative DAG of Model \eqref{eq:CoDEAL-Model}}\label{append:DAG}

\begin{figure}[h!]
\centering
\includegraphics[width=1.0\textwidth]{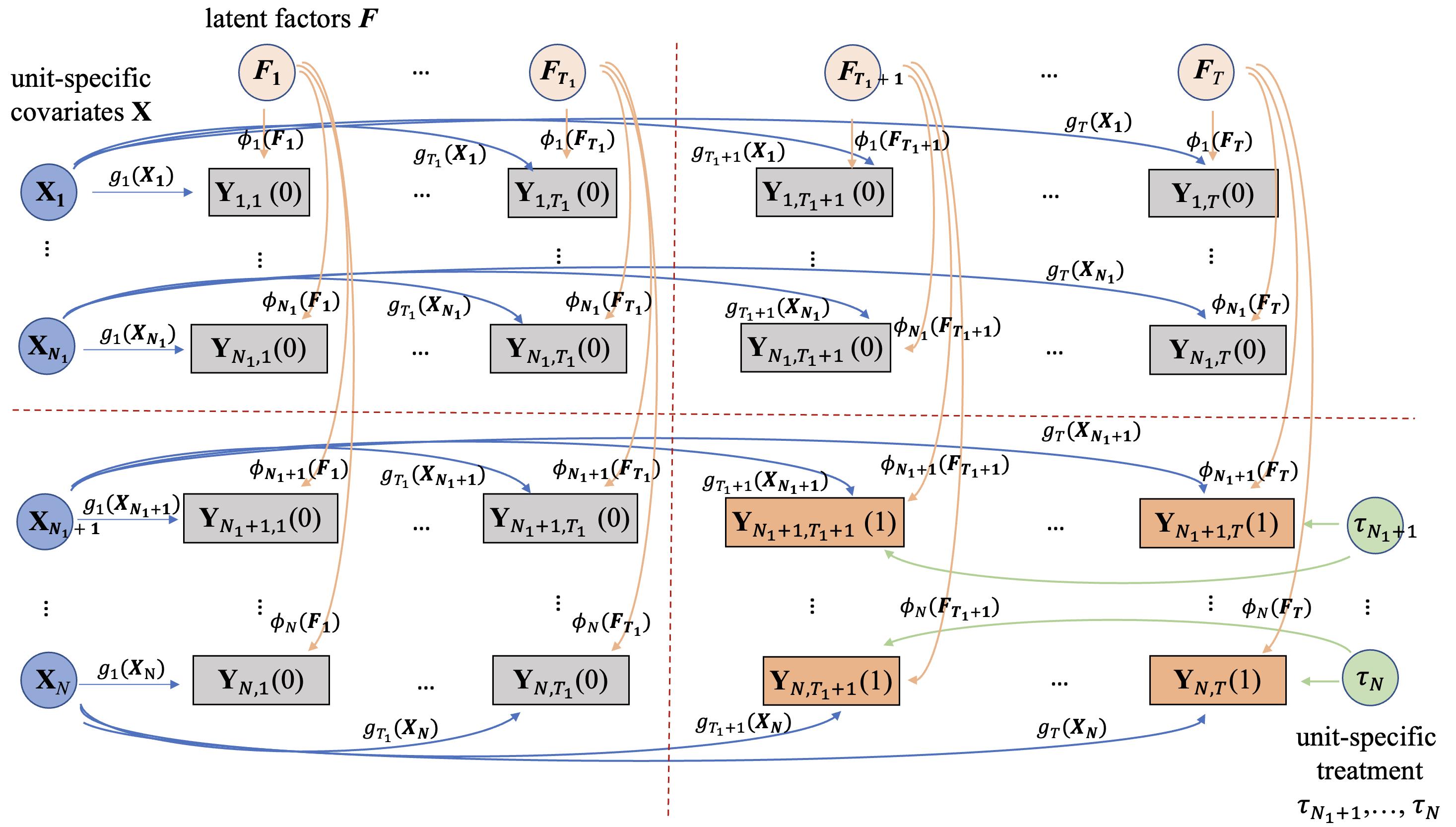}
\caption{Illustrative diagram of the CoDEAL model. Observed covariates $\bX_i$ (left) affect outcomes through $g_t(\bX_i)$, while latent time-varying factors $\bF_t$ influence outcomes through $\phi_i(\bF_t)$. The pre-treatment period (left of the vertical dashed line) contains only control outcomes, while the post-treatment period (right) includes treated units with unit-specific effects $\tau_i$. %
}
\label{fig:fbsad2}
\end{figure} 

Figure \ref{fig:fbsad2} plots an illustrative DAG of model \eqref{eq:CoDEAL-Model}:  $$ Y_{it}= \phi_i^\star(\bF^\star_{t}) + g^\star_{t}(\bX_i) + \tau_i^\star \cdot \mathds{1}\{W_{it}=1\} + \varepsilon_{it}. $$
For clarity, the illustrative DAG emphasizes the key features of our model: (i) $\bF^\star_{t}$ acts as a time-varying confounder, (ii) treatment has a contemporaneous effect on outcomes, (iii) time-varying covariate effects can be accommodated in the proposed framework, and (iv) temporal dependence operates through $\bF^\star_{t}$ rather than through explicit lagged outcomes or treatments. %

While an illustrative DAG offers an intuitive visualization, a fully rigorous causal DAG would be nontrivial. Despite the high-level mathematical equivalence between Rubin's potential outcome framework \citep{rubin1974estimating,imbens2015causal} and Pearl's causal DAG framework \citep{pearl1995causal,pearl2009causality}, mapping the causal panel data model onto a faithful graphical structure presents significant challenges. 
In our setting, temporal dependence is governed through time-varying latent factors (i.e., $\bF^\star_{t}$), which act as unobserved confounders; representing these in a causal DAG requires introducing unobserved nodes with specific structural assumptions about their evolution and connections across time. 
Notably, in the relevant literature on causal panel models (e.g., \citep{athey2019estimating,bai2021matrix,choi2024matrix}, including the recent literature review by \citep{arkhangelsky2024causal}), there has been no formal statement framing causal panel models within a causal graphical framework in rigorous terms. Establishing a complete and rigorous equivalence between the two formulations in the causal panel setting is beyond the scope of this paper.

\subsection{Rationale of Sequential Estimation and Comparisons with Joint Estimation}
\label{append:seq-vs-joint}

For CoDEAL, we employ a two-step estimation strategy: first, we model covariate effects to obtain adjusted observed outcomes; second, we estimate counterfactuals using a multi-output autoencoder (AE). We adopt this sequential approach over joint estimation that estimates covariate and factor effects simultaneously for several strategic reasons.

First, sequential estimation improves stability and interpretability. Joint estimation risks entangling the effects of latent factor $\bF_t^\star$ and observed covariates $\bX_i^\star$. If optimized jointly, the two neural networks may compete to explain the same signal, making it difficult to disentangle observed covariate effects from latent structure, and may potentially cause identification issues. Sequential estimation enforces a clean separation: covariate effects are estimated first from observed information, and only residual variation is attributed to the latent structure. This prevents the autoencoder (AE) from ``absorbing'' covariate effects in an uninterpretable way.

Moreover, sequential design is not merely a procedural convenience, but is a natural alignment with the data structure. $\bX_i$ is observed and time-invariant, whereas $\bF_t^\star$ is unobserved and time-varying. This asymmetry motivates a sequential ordering. It also aligns with the spirit of double/debiased machine learning, where nuisance functions are partialled out in a first stage so that the second stage operates on a less confounded signal.

To provide empirical grounding for this choice, we perform simulation experiments to evaluate the relative performance of sequential (SEQ) versus joint (JOINT) estimation strategies. Following the data-generating process detailed in Section \ref{sec:simulation}, we generate data with
\[
g_t^\star(X_i)
=
1+\left[\tanh\left(|X_i^\top w_t|\right)\right]^{1/2}+b_t,
\]
and use ReLU-MLP factors for $\phi_i^\star(F_t^\star)$, following the setup in Table~S.1. Across 20 replications, the average ground-truth component values are
$g^\star = 1.7968, 
\phi^\star = 0.1515, $ 
and $ g^\star+\phi^\star = 1.9483.$
Table~\ref{tab:seq_joint_summary} reports the summary statistics of the estimated covariate and latent-factor components.

\begin{table}[htbp]
\centering
\caption{Comparison between sequential estimation (SEQ) and joint estimation (JOINT) over 20 replications.}
\label{tab:seq_joint_summary}
\small
\setlength{\tabcolsep}{6pt}
\begin{tabular}{lcccccc}
\hline
& \multicolumn{3}{c}{SEQ} & \multicolumn{3}{c}{JOINT} \\
\cline{2-4} \cline{5-7}
& $\hat g$ & $\hat\phi$ & $\hat g+\hat\phi$
& $\hat g$ & $\hat\phi$ & $\hat g+\hat\phi$ \\
\hline
Mean & 1.9010 & 0.0409 & 1.9417 & 0.2814 & 1.6465 & 1.9279 \\
SD   & 1.0508 & 0.1043 & 1.0934 & 0.1433 & 1.0283 & 1.1063 \\
MSE  & 0.9832 & 0.9573 & 0.0033 & 2.3166 & 2.2593 & 0.0069 \\
\hline
\end{tabular}
\end{table}

The results show that SEQ yields smaller MSEs than JOINT for both component-wise estimation and aggregate estimation. 
At the aggregate level, both methods estimate $g^\star+\phi^\star$ reasonably well. However, the component-wise decomposition differs substantially. JOINT assigns most of the signal to the latent factor network, whereas SEQ provides a decomposition closer to the intended model structure. In particular, JOINT has much larger component-wise MSE for $\hat g$ and $\hat\phi$, despite producing a comparable estimate of their sum. This suggests that joint optimization can obscure the separation between observed covariate effects and latent factor effects.

To further reveal the entanglement phenomenon between $\widehat{g}_t(X_i)$ and  $\widehat\phi_i(\widehat{F}_t)$ in JOINT, Table~\ref{tab:seq_joint_replication} reports the per-replication decomposition. In JOINT, the constant component in $g_t^\star(X_i)$ is frequently absorbed by the latent factor network $\hat\phi$. This is likely due to the flexibility of the autoencoder, which can absorb constant or shared signals into the latent representation. In contrast, SEQ removes the constant component during the covariate-removal step, leaving the remaining residual signal for factor extraction.

\begin{table}[!h]
\centering
\caption{Per-replication comparison between SEQ and JOINT.}
\label{tab:seq_joint_replication}
\small
\setlength{\tabcolsep}{3pt}
\begin{tabular}{c|ccc|ccc|ccc}
\hline
Sim
& \multicolumn{3}{c|}{True value}
& \multicolumn{3}{c|}{SEQ}
& \multicolumn{3}{c}{JOINT} \\
\cline{2-10}
& $g^\star$ & $\phi^\star$ & $g^\star+\phi^\star$
& $\hat g$ & $\hat\phi$ & $\hat g+\hat\phi$
& $\hat g$ & $\hat\phi$ & $\hat g+\hat\phi$ \\
\hline
1  & 1.6327 & -1.1332 & 0.4995  & 0.4476  & -0.0081 & 0.4395  & 0.0671 & 0.2076  & 0.2748 \\
2  & 1.8617 & 1.2469  & 3.1086  & 2.8624  & 0.2424  & 3.1048  & 0.1713 & 2.9729  & 3.1442 \\
3  & 1.8485 & -0.5985 & 1.2500  & 1.3981  & -0.1415 & 1.2567  & 0.1582 & 1.2026  & 1.3609 \\
4  & 1.8246 & 0.0237  & 1.8483  & 1.8629  & 0.0875  & 1.9504  & 0.3391 & 1.5944  & 1.9335 \\
5  & 1.8295 & 0.0629  & 1.8924  & 1.8260  & -0.0135 & 1.8125  & 0.4222 & 1.4780  & 1.9002 \\
6  & 2.0058 & 1.1126  & 3.1184  & 3.0681  & 0.1774  & 3.2456  & 0.3539 & 2.6920  & 3.0459 \\
7  & 1.7880 & -0.7071 & 1.0809  & 1.1180  & -0.0462 & 1.0718  & 0.0906 & 1.0832  & 1.1738 \\
8  & 1.5881 & -1.8680 & -0.2799 & -0.2799 & -0.0166 & -0.2965 & 0.0122 & -0.4029 & -0.3907 \\
9  & 1.8263 & 0.5298  & 2.3561  & 2.2064  & 0.1415  & 2.3480  & 0.3307 & 1.9936  & 2.3242 \\
10 & 1.8187 & 0.1177  & 1.9363  & 1.8625  & 0.1091  & 1.9717  & 0.1847 & 1.7044  & 1.8891 \\
11 & 1.8074 & 0.4165  & 2.2238  & 2.1950  & 0.0999  & 2.2949  & 0.2813 & 1.8516  & 2.1329 \\
12 & 1.8108 & 0.0605  & 1.8713  & 1.8906  & -0.0706 & 1.8199  & 0.3713 & 1.5437  & 1.9150 \\
13 & 1.8646 & -1.4704 & 0.3942  & 0.3745  & -0.0495 & 0.3251  & 0.3569 & -0.0833 & 0.2736 \\
14 & 1.8628 & -0.1993 & 1.6635  & 1.5455  & 0.0408  & 1.5862  & 0.1689 & 1.4396  & 1.6085 \\
15 & 1.7500 & 1.9254  & 3.6753  & 3.5302  & 0.1655  & 3.6956  & 0.3782 & 3.2931  & 3.6713 \\
16 & 1.7984 & 1.3016  & 3.1000  & 3.0414  & -0.0008 & 3.0405  & 0.2314 & 2.8129  & 3.0444 \\
17 & 1.8398 & 0.6253  & 2.4651  & 2.3634  & 0.1366  & 2.5000  & 0.3930 & 2.1074  & 2.5005 \\
18 & 1.7183 & -0.8995 & 0.8188  & 0.7065  & 0.0899  & 0.7963  & 0.2918 & 0.4728  & 0.7647 \\
19 & 1.5954 & 1.0235  & 2.6189  & 2.6049  & -0.0050 & 2.5999  & 0.4415 & 2.2092  & 2.6506 \\
20 & 1.8643 & 1.4608  & 3.3251  & 3.3918  & -0.1201 & 3.2717  & 0.5840 & 2.7569  & 3.3409 \\
\hline
\end{tabular}%
\end{table}

This experiment provides empirical evidence for the entanglement issue in joint estimation: JOINT risks making the decomposition into latent-factor and observed-covariate contributions less stable and less interpretable. While JOINT can still produce a reasonable estimate of the aggregate signal $\hat g+\hat\phi$, it may yield a misleading decomposition between observed covariate effects and latent factor effects. This can compromise the interpretability of the learned factors and deviate from the intended model design, where $g_t(\cdot)$ captures observed covariate effects and $\phi_i(\cdot)$ captures latent effects.

More generally, when two flexible neural networks jointly model an additive structure of the form $g(\cdot)+\phi(\cdot)$, the decomposition is not identifiable without additional restrictions. For any overlapping component $h(\cdot)$, the same aggregate function can be represented as
\[
g(\cdot)+\phi(\cdot)
=
\{g(\cdot)-h(\cdot)\}
+
\{\phi(\cdot)+h(\cdot)\},
\]
which leaves the prediction unchanged but reallocates signal between the two components. We therefore interpret the above results not as a formal identification failure, but as evidence that joint optimization can absorb shared signals into different channels in a non-robust way. This supports the use of sequential estimation in CoDEAL.

\section{Supplement to Simulation Studies in Section \ref{sec:simulation}} \label{append:simulation}

In this section, we use extensive simulation studies to demonstrate the finite-sample performance of the proposed CoDEAL. The context in this appendix supplements Section \ref{sec:simulation} of the main context. 

\subsection{Benchmark Methods} \label{append:simu-benchmark}
We consider a variety of benchmark methods, including a single-output AE (\textbf{Single AE}) (detailed below), neuron-enhanced AE (\textbf{AEMC-NE}) \citep{fan2024neuron}, matrix completion with nuclear norm minimization (\textbf{MC-NNM}) \citep{athey2021matrix}, vertical regression (\textbf{Vert-Reg}) \citep{doudchenko2016balancing}, difference-in-differences (\textbf{DiD}) \citep{imai2021use}, and %
kernel principal component analysis (\textbf{KernelPCA}) \citep{scholkopf1997kernel}. %
In the assessment of performance in causal matrix completion, we further compare to three synthetic-control-style baselines, including the classic synthetic control (\textbf{SCM})\citep{abadie2010synthetic}, generalized synthetic control (\textbf{GSC}) \citep{xu2017generalized}, and synthetic difference-in-differences (\textbf{SDID}) \citep{arkhangelsky2021synthetic}.

Additionally, we evaluate the algorithm's performance in terms of covariate effect removal. To this end, we compare the performance of CoDEAL using different methods to remove the covariate effects, specifically, DNN-based removal (as illustrated in Section \ref{subsec:method-CoDEAL}), and linear-regression(LR)-based removal (detailed below). 

We evaluate the algorithm's performance using two evaluation metrics: 
the mean absolute error
$MAE=(\sum_{(i, t): W_{it} = 1}|Y_{it}(1)-\widehat{{Y}}_{it}(0)-\tau_{i}^\star|)/(\sum_{(i,t)} W_{it})$ 
and the mean squared error $MSE=(\sum_{(i,t): W_{it} = 1}(Y_{it}(1)-\widehat{{Y}}_{it}(0)-\tau_{i}^\star)^2)/(\sum_{(i,t)} W_{it})$.

\noindent {\textbf{Matrix Completion by Single-Output AE (as a benchmark method).}}
Our proposed CoDEAL utilizes a multi-output autoencoder (as illustrated in Figure \ref{fig:method}) to recover nonlinear latent factors while explicitly modeling the unit-specific heterogeneity through multiple decoders.
The multi‐output autoencoder uses a shared encoder network \(\rho(\cdot)\) to map each column \(\widetilde{\mathbf{Y}}_{\!\cdot t}\) into a \(K\)-dimensional latent space, followed by \(N\) separate decoder networks \(\{\phi_1,\dots,\phi_N\}\) to reconstruct each unit’s value. 
To demonstrate CoDEAL's ability to account for the unit-specific heterogeneity, we include an additional benchmark that replaces the multi-output autoencoder with a single-output autoencoder in the CoDEAL algorithm. This benchmark is referred to as the Single AE in later discussions. 
A single-output autoencoder uses the same encoder \(\rho(\cdot)\) but only one decoder \(\phi(\cdot)\) (i.e. $\phi=\phi_1=\dots=\phi_N$) for all units. By design, the common decoder makes the single-output AE inherently unable to model the unit-specific heterogeneity effectively. For implementation,  DNNs in the single-output AE are estimated by
\begin{equation*}
\widehat{\phi}\circ\widehat{\rho}(\cdot)=\underset{\phi\circ{\rho}(\cdot)\in \mathcal{G}_{\text{AE}}^{K_1}  }{\arg \min}\sum_{(i,t):W_{it}=0}\Bigr(\phi(\rho(\widetilde{\bY}_{\cdot t}))-\widetilde{Y}_{it}\Bigr)^2  .
\end{equation*}

{\textbf{Covariate Effect Removal by Linear Regression (as a benchmark method).}}
Our proposed CoDEAL uses a DNN to remove the potentially nonlinear, heterogeneous, and time-varying effects brought by covariates. To evaluate the algorithm's performance in terms of the covariate effect removal, we include a linear-regression(LR)-based removal as a benchmark. The LR-based removal assumes that covariates have a linear impact on the outcome.
With covariates $\mathbf{X}_i\in\mathbb{R}^{P}$, $i\in [N]$, we denote the effect of covariates on the outcome by a coefficient matrix $\bH = [\bH_1, \ldots, \bH_T] \in\mathbb{R}^{P\times T}$, and we estimate $\bH_t \in \mathbb{R}^P$ separately for each period. 
For each pre-treatment time $t\in\{1,\dots,T_1\}$, we estimate the covariate effect vector $\widehat{\bH}_t\in\mathbb{R}^{P}$ using OLS on all units:
$\widehat{\bH}_t = \arg\min_{\bH_t} \sum_{i=1}^{N}\left(Y_{it}-\bX_i^\top\bH_t\right)^2.$
For each post-treatment period $t\in\{T_1+1,\dots,T\}$, we use only control units to estimate the covariate effect:
$\widehat{\bH}_t = \arg\min_{\bH_t} \sum_{i: W_{it}=0}\left(Y_{it}-\bX_i^\top\bH_t\right)^2.$
With the estimated $\widehat{\bH}_t, t\in [T]$, we obtain the LR-based covariate-adjusted
observed outcomes by 
$\widetilde{Y}_{it} = Y_{it} - \bX_i^\top\widehat{\bH}_t$ for each $i \in [N]$ and $t \in [T]$.

\subsection{Data Generating Process and Simulation Configurations } \label{append:simulation-DGP}

We perform extensive simulation studies under various linear and nonlinear generating mechanisms, with different setups of unit numbers $N$, time periods $T$, matrix ranks $K$, and treatment assignments. Specifically, we consider the following configurations 1-4: 
\begin{description}
    \item[] 1. Four-Block Design: $(N,T,N_1,T_1,P,K)=(100,200,50,100,3,4)$
    \item[] 2. Four-Block Design: $(N,T,N_1,T_1,P,K)=(200,120,100,60,5,3)$
    \item[] 3. Staggered Adoption Design: $(N,T,K)=(100,200,4)$, $r=\{5,10\}$
    \item[] 4. Staggered Adoption Design: $(N,T,K)=(200,120,3)$, $r=\{5,10\}$.
\end{description}

For the generation of factor effects and covariate effects, we first generate covariates 
$\bX = (X_{it}) \in \mathbb{R}^{N\times P}$, factors $\bF = (F_{it}) \in \mathbb{R}^{T\times K}$, factor loadings $\boldsymbol{\Lambda} = (\Lambda_{it})\in \mathbb{R}^{N\times K}$ with $X_{it},F_{it}, \Lambda_{it}\overset{\text{i.i.d.}}{\sim}N(0,1)$. With these building blocks, we consider various linear and nonlinear generating mechanisms to generate factor effects $\phi_i^\star(\bF^\star_{t})$ and covariate effects $g^\star_t(\mathbf{X}_i)$. Specifically, we experiment with 4 variants for generating factor effects and 6 variants for generating covariate effects, as listed in Table \ref{tab:data_gen}. The unit-specific causal effect $\tau_i^\star \overset{i.i.d}{\sim}N(12,5) $ is designed to be added to treated units in treated periods. The observed outcome matrix $\bY = (Y_{it}) \in \mathbb{R}^{N \times T}$ is then obtained by 
$Y_{it}= Y^\star_{it} + \tau^\star_i \cdot \mathcal{I}\{W_{it}=1\} + \varepsilon_{it}$
where $W_{it}$ are determined based on the treatment assignments according to the simulation configuration, and the idiosyncratic error $\varepsilon_{it}$ follows $\varepsilon_{it}\overset{i.i.d}{\sim}\mathcal{N}(0,0.5^2)$.

\begin{table}[h]
\centering
\small
\caption{Data Generation Processes for Latent Factor Effects and Covariate Effects}
\resizebox{\textwidth}{!}{
\begin{tabular}{ll|ll}
\toprule
\multicolumn{2}{c|}{\textbf{Factor Effect Generation} ($\phi_i^\star(\bF^\star_{t})$)} & \multicolumn{2}{c}{\textbf{Covariate Effect  Generation} ($g^\star_t(\mathbf{X}_i)$)} \\
\midrule
\textbf{Linear} & $0.5C_1\boldsymbol{\Lambda}_i^\top \mathbf{F}_t$,  $C_1\overset{\text{i.i.d.}}{\sim}N(0,1)$& \textbf{Matrix Linear} & $\bX_i^\top\bU_{t}$, $\bU = (\bU_1, \ldots, \bU_{t})\overset{\text{i.i.d.}}{\sim}N_{P\times T}(1,1)$\\
\textbf{Sine} & $2C_1\sin(\boldsymbol{\Lambda}_i^\top \mathbf{F}_t)$ & \textbf{Vector Linear} & $\bX_i^\top \bU$, $\bU \overset{\text{i.i.d.}}{\sim}N_P(1,1)$\\
\textbf{Polynomial} & $0.2C_1\boldsymbol{\Lambda}_i^\top \mathbf{F}_t + 0.2C_2\mathbf{F}_t^\top \mathbf{F}_t$, $C_2\overset{\text{i.i.d.}}{\sim}N(0,1)$& \textbf{tanh} & $[\tanh(|\mathbf{X}_i^\top \boldsymbol{w}_t|)]^{1/2}+b_t$, $\boldsymbol{w}_t\sim N_P(0,1)$\\
\textbf{ReLU MLP} & $\bR_i^{(2)}\text{ReLU}(\bR_i^{(1)}\mathbf{F}_t+\mathbf{b}_i^{(1)})+b_i^{(2)}$ & \textbf{poly} & $|\mathbf{X}_i^\top \boldsymbol{w}_t|^{1/2}+b_t$ \\
& $b_i^{(2)}\overset{\text{i.i.d.}}{\sim}N(0,0.5^2)$, $\bR_i^{(1)}\overset{\text{i.i.d.}}{\sim}N_{h_1\times r}(0,0.5^2)$& \textbf{log} & $\log(|\mathbf{X}_i^\top \boldsymbol{w}_t+b_t|)$, $b_{t}\overset{\text{i.i.d.}}{\sim}N(0,1)$ \\
& $\mathbf{b}_i^{(1)},\bR_i^{(2)}\overset{\text{i.i.d.}}{\sim}N_{h_1}(0,0.5^2)$, $h_1=10$& \textbf{ReLU} & $\text{ReLU}(\bX_i\bR_1^{(t)}+\mathbf{b}_1^{(t)})\bR_2^{(t)}+b_2^{(t)}, b_2^{(t)} \overset{\text{i.i.d.}}{\sim}N(0,1)$\\
 & & &$\bR_1^{(t)}\overset{\text{i.i.d.}}{\sim}N_{P \times h_2}(0,\frac{2}{P}), \mathbf b_1^{(t)} \overset{\text{i.i.d.}}{\sim}N_{h_2}(0,1)$\\
 & & &$\bR_2^{(t)} \overset{\text{i.i.d.}}{\sim}N_{h_2}(0,\frac{2}{h_2})$, $h_2=32$\\
\bottomrule
\label{tab:data_gen}
\end{tabular}}

Note: The vector linear covariate effect generation setting mimics a situation in which covariate effects are heterogeneous across panel units but remain time-invariant over time. The matrix linear covariate effect generation setting, however, allows for heterogeneous and time-varying covariate effects. 
\end{table}

\subsection{Evaluation on the Performance of CoDEAL in the Removal of Covariate Effects}
\label{append:simu-covariate-removal}
To evaluate the algorithm’s performance in terms of the covariate effect removal, we compare the performance of CoDEAL using different methods, specifically, DNN (as illustrated in Section \ref{subsec:method-CoDEAL}) and LR (detailed in Section \ref{append:simu-benchmark}), to remove the covariate effects. 
Six scenarios of covariate effects are considered as detailed in Table \ref{tab:data_gen}. We also include an oracle benchmark, when no covariate effects are present and no covariate-adjusted operations are needed.

\begin{table}[!bt]
\small
\centering
\caption{Comparisons of prediction accuracy (measured by MAE and MSE) with covariate effects removed by \textbf{LR} and \textbf{DNN} under various covariate generations in Config.1.  Results are averaged over 50 replications. Numbers in parentheses are standard errors.}
\resizebox{\textwidth}{!}{
\begin{tabular}{llccccccll}
\toprule
\multirow{2}{*}{\textbf{Covariate}} & \multirow{2}{*}{\textbf{Method}} & \multicolumn{2}{c}{\textbf{LR}, Linear Factor}&\multicolumn{2}{c}{\textbf{NN}, Linear Factor}& \multicolumn{2}{c}{\textbf{LR}, Sine Factor}& \multicolumn{2}{c}{\textbf{NN}, Sine Factor}\\
& &MAE& MSE&MAE& MSE&MAE& MSE& \text{ }\text{ }\text{ }\text{ }MAE&\text{ }\text{ }\text{ }\text{ }\text{ }MSE\\
\midrule

\multirow{7}{*}{\textbf{No covariate}}
& CoDEAL & \textbf{0.560(0.009)} &0.578(0.028)&\textbf{0.560(0.009)} &0.578(0.028)&\textbf{0.933(0.015)} & \textbf{1.722(0.066)} & \textbf{0.933(0.015)} &\textbf{1.722(0.066)} \\
& Single AE   & 0.724(0.009)&1.081(0.034)&0.724(0.009)&1.081(0.034)&1.071(0.013)& 2.131(0.056)& 1.071(0.013)&2.131(0.056)\\
& AEMC-NE   & 0.734(0.009)&1.142(0.039)&0.734(0.009)&1.142(0.039)&1.090(0.013)& 2.172(0.056)& 1.090(0.013)&2.172(0.056)\\
& DiD         & 0.804(0.010)&1.364(0.043)&0.804(0.010)&1.364(0.043)&1.141(0.013)& 2.343(0.059)& 1.141(0.013)&2.343(0.059)\\
& Vert-Reg    & 0.601(0.002)&\textbf{0.574(0.004)} &0.601(0.002)&\textbf{0.574(0.004)} &1.024(0.012)& 2.228(0.069)& 1.024(0.012)&2.228(0.069)\\
& MC-NNM      & 0.580(0.010)&0.593(0.030)&0.580(0.010)&0.593(0.030)&1.093(0.013)& 2.147(0.054)& 1.093(0.013)&2.147(0.054)\\
& KernelPCA     &0.773(0.009)&1.252(0.037)&0.773(0.009)&1.252(0.037)&1.099(0.009)&2.184(0.056)&1.099(0.009)&2.184(0.056) \\
\midrule

\multirow{7}{*}{\textbf{Matrix Linear}}
& CoDEAL & \textbf{0.585(0.013)}&\textbf{0.615(0.024)}&\textbf{0.646(0.009)} &\textbf{0.722(0.023)} &\textbf{0.983(0.011)}& \textbf{1.889(0.046)}& \textbf{1.079(0.012)}&\textbf{2.135(0.054)} \\
& Single AE   & 0.751(0.008)&1.131(0.032)&0.821(0.010)&1.320(0.039) &1.123(0.014)& 2.283(0.060)& 1.226(0.013)&2.626(0.062)\\
& AEMC-NE   & 0.779(0.010)&1.244(0.041)&0.838(0.009)&1.394(0.041) &1.145(0.013)& 2.336(0.057)& 1.209(0.013)&2.564(0.060) \\
& DiD         & 0.837(0.010)&1.444(0.045)&0.892(0.010)&1.587(0.045) &1.187(0.013)& 2.490(0.060)& 1.244(0.013)&2.696(0.063)\\
& Vert-Reg    & 0.686(0.006)&0.766(0.015)&0.757(0.006)&0.957(0.019) &1.103(0.012)& 2.498(0.066)& 1.169(0.012)&2.647(0.064)\\
& MC-NNM      & 0.604(0.009)&0.664(0.025)&0.699(0.006)&0.895(0.020) &1.140(0.013)& 2.293(0.055)& 1.201(0.013)&2.510(0.059)\\
& KernelPCA    &0.830(0.008)&1.413(0.037)  &0.811(0.007) &1.260(0.030)&1.179(0
012)&2.433(0.049)&1.195(0.009) &2.496(0.039) \\
\midrule

\multirow{7}{*}{\textbf{Vector Linear}}
& CoDEAL & \textbf{0.577(0.010)}&\textbf{0.592(0.026)}&\textbf{0.652(0.006)}&\textbf{0.715(0.017)}&\textbf{0.964(0.016)}& \textbf{1.775(0.060)}& \textbf{1.053(0.011)}&\textbf{2.035(0.045)} \\
& Single AE   & 0.746(0.007)&1.106(0.026)&0.841(0.008)&1.373(0.030) &1.106(0.012)& 2.226(0.054)& 1.205(0.012)&2.549(0.053)\\
& AEMC-NE   & 0.778(0.010)&1.205(0.032)&0.823(0.008)&1.317(0.029)&1.135(0.013)& 2.298(0.053)& 1.191(0.012)&2.497(0.053)\\
& DiD         & 0.825(0.008)&1.373(0.031)&0.872(0.009)&1.489(0.033)&1.173(0.012)& 2.437(0.053)& 1.224(0.012)&2.620(0.056) \\
& Vert-Reg    & 0.669(0.005)&0.727(0.012)&0.733(0.006)&0.887(0.017) &1.085(0.011)& 2.413(0.057)& 1.145(0.011)&2.520(0.053) \\
& MC-NNM      & 0.591(0.008)&0.617(0.020)&0.678(0.007)&0.826(0.020) &1.106(0.017)& 2.219(0.099)& 1.175(0.012)&2.412(0.053)\\
& KernelPCA    &0.819(0.009)&1.402(0.035)  &0.806(0.008) &1.208(0.029)&1.156(0
011)&2.398(0.047)&1.189(0.008) &2.437(0.038) \\
\midrule

\multirow{7}{*}{\textbf{tanh}}
& CoDEAL & 0.778(0.008)&1.168(0.029)&\textbf{0.701(0.008)}&\textbf{0.871(0.031)}&\textbf{1.142(0.013)}& 2.366(0.059)& \textbf{1.080(0.013)}&\textbf{2.175(0.059)} \\
& Single AE   & 0.889(0.008)&1.554(0.035)&0.851(0.008)&1.383(0.033) &1.299(0.014)& 2.910(0.067)& 1.257(0.014)&2.748(0.066) \\
& AEMC-NE   & 0.988(0.008)&1.822(0.036)&0.862(0.009)&1.445(0.036) &1.305(0.013)& 2.925(0.064)& 1.239(0.014)&2.677(0.065) \\
& DiD         & 0.855(0.008)&1.473(0.035)&0.916(0.009)&1.642(0.037) &1.199(0.013)& 2.536(0.061)& 1.274(0.014)&2.817(0.067) \\
& Vert-Reg    & \textbf{0.721(0.005)} &\textbf{0.844(0.015)} &0.798(0.007)&1.075(0.024) &1.146(0.012)& 2.660(0.071)& 1.212(0.013)&2.787(0.072) \\
& MC-NNM      & 0.837(0.007)&1.131(0.021)&0.735(0.005)&1.007(0.017)&1.153(0.013)& \textbf{2.339(0.056)} & 1.234(0.013)&2.636(0.063) \\
& KernelPCA      &0.856(0.008)&1.471(0.036)&0.824(0.008)&1.292(0.033) &1.196(0.011)&2.496(0.049)&1.231(0.009)&2.626(0.041)  \\
\midrule

\multirow{7}{*}{\textbf{log}}
& CoDEAL & \textbf{1.020(0.006)} &\textbf{1.873(0.026)} &\textbf{0.905(0.010)} &\textbf{1.517(0.030} &\textbf{1.345(0.012)}& \textbf{3.233(0.065)} & \textbf{1.253(0.013)}&\textbf{2.849(0.064)} \\
& Single AE   & 1.119(0.006)&2.266(0.031)&1.013(0.008)&1.943(0.033) &1.427(0.012)& 3.502(0.065)& 1.386(0.013)&3.332(0.068) \\
& AEMC-NE   & 1.137(0.007)&2.386(0.036)&1.024(0.008)&2.007(0.036) &1.421(0.012)& 3.483(0.064)& 1.372(0.014)&3.281(0.069) \\
& DiD         & 1.127(0.007)&2.371(0.037)&1.069(0.008)&2.182(0.038) &1.408(0.013)& 3.437(0.066)& 1.401(0.014)&3.402(0.071) \\
& Vert-Reg    & 1.100(0.006)&2.087(0.024)&1.055(0.007)&1.940(0.028) &1.485(0.012)& 4.054(0.077)& 1.439(0.012)&3.737(0.074) \\
& MC-NNM      & 1.044(0.006)&1.998(0.028)&0.939(0.005)&1.649(0.020) &1.373(0.012)& 3.266(0.061)& 1.366(0.013)&3.233(0.067)\\
& KernelPCA   & 1.136(0.008)&2.426(0.040)&1.064(0.009)&2.174(0.041) &1.417(0.010)& 3.455(0.050)& 1.391(0.011)& 3.326(0.052)\\
\midrule

\multirow{7}{*}{\textbf{poly}}
& CoDEAL & 0.833(0.007)&1.290(0.028)&\textbf{0.715(0.007)} &\textbf{0.913(0.035)} &\textbf{1.182(0.013)}& \textbf{2.507(0.059)} & \textbf{1.024(0.013)}&\textbf{2.086(0.062)} \\
& Single AE   & 0.946(0.008)&1.704(0.036)&0.854(0.008)&1.395(0.033) &1.345(0.014)& 3.088(0.068)& 1.261(0.014)&2.760(0.066) \\
& AEMC-NE   & 1.060(0.008)&2.038(0.037)&0.868(0.008)&1.461(0.036) &1.348(0.012)& 3.088(0.063)& 1.245(0.014)&2.701(0.065) \\
& DiD         & 0.906(0.008)&1.596(0.036)&0.921(0.009)&1.653(0.037) &1.235(0.013)& 2.659(0.062)& 1.279(0.014)&2.832(0.067) \\
& Vert-Reg    & \textbf{0.756(0.006)} &\textbf{0.934(0.017)} &0.800(0.007)&1.078(0.023) &1.281(0.012)& 2.983(0.072)& 1.216(0.013  &2.790(0.070) \\
& MC-NNM      & 0.847(0.006)&1.305(0.032)&0.747(0.005)&1.014(0.017) & 1.290(0.013)& 2.763(0.057)& 1.238(0.013)&2.652(0.063) \\
& KernelPCA   & 0.924(0.008)&1.637(0.037)&0.920(0.009)&1.563(0.038)& 1.274(0.012)& 2.785(0.053) &1.243(0.011)& 2.659(0.049)\\

\midrule

\multirow{7}{*}{\textbf{ReLU}}
& CoDEAL & 0.946(0.009)&1.702(0.042)&\textbf{0.717(0.008)} &\textbf{0.916(0.025)} &1.205(0.011)&2.645(0.056)& \textbf{1.121(0.012)} &\textbf{2.269(0.054)} \\
& Single AE   & 1.020(0.010)&1.971(0.046)&0.858(0.010)&1.407(0.039) &1.317(0.011)&2.976(0.050)& 1.222(0.014)&2.617(0.062) \\
& AEMC-NE   & 1.288(0.011)&2.982(0.053)&0.868(0.010)&1.471(0.043) &1.544(0.012)&4.003(0.062)& 1.235(0.013)&2.660(0.062) \\
& DiD         & 0.960(0.009)&1.777(0.046)&0.922(0.010)&1.670(0.047) &1.274(0.011)&2.802(0.050)& 1.269(0.014)&2.791(0.064) \\
& Vert-Reg    & \textbf{0.779(0.006)} &\textbf{1.013(0.020)} &0.782(0.006)&1.027(0.021) &\textbf{1.195(0.010)}&2.786(0.059)& 1.199(0.012)&2.753(0.067) \\
& MC-NNM      & 0.967(0.009)&1.893(0.035)&0.764(0.009)&1.073(0.024) &1.233(0.011)&\textbf{2.620(0.048)} & 1.227(0.013)&2.608(0.060)\\
& KernelPCA   & 0.955(0.008)&1.746(0.038)&0.919(0.009)&1.645(0.039) &1.270(0.011)&2.773(0.049)& 1.264(0.012)& 2.747(0.052)\\

\bottomrule
\label{tab:lrnn1}
\end{tabular}
}
\end{table}

\begin{table}[!bt]
\small
\centering
\caption{Comparisons of prediction accuracy (measured by MAE and MSE) with covariate effects removed by \textbf{LR} and \textbf{DNN} under various covariate generations in Config.2.  Results are averaged over 50 replications. Numbers in parentheses are standard errors.}
\resizebox{\textwidth}{!}{
\begin{tabular}{llccccccll}
\toprule
\multirow{2}{*}{\textbf{Covariate}} & \multirow{2}{*}{\textbf{Method}} & \multicolumn{2}{c}{\textbf{LR}, Linear Factor}&\multicolumn{2}{c}{\textbf{NN}, Linear Factor}& \multicolumn{2}{c}{\textbf{LR}, Sine Factor}& \multicolumn{2}{c}{\textbf{NN}, Sine Factor}\\
& &MAE& MSE&MAE& MSE&MAE& MSE& \text{ }\text{ }\text{ }\text{ }MAE&\text{ }\text{ }\text{ }\text{ }\text{ }MSE\\
\midrule

\multirow{7}{*}{\textbf{No covariate}}
& CoDEAL & \textbf{0.507(0.004)} &\textbf{0.425(0.010)} &\textbf{0.507(0.004)} &\textbf{0.425(0.010)} &\textbf{0.779(0.007)} & \textbf{1.173(0.027)} & \textbf{0.779(0.007)} &\textbf{1.173(0.027)} \\
& Single AE   & 0.622(0.006)&0.761(0.021)&0.622(0.006)&0.761(0.021)&0.963(0.010)& 1.772(0.038)& 0.963(0.010)&1.772(0.038)\\
& AEMC-NE   & 0.647(0.006)&0.850(0.023)&0.647(0.006)&0.850(0.023)&1.004(0.009)& 1.884(0.036)& 1.004(0.009)&1.884(0.036)\\
& DiD         & 0.713(0.007)&1.043(0.028)&0.713(0.007)&1.043(0.028)&1.085(0.010)& 2.163(0.041)& 1.085(0.010)&2.163(0.041)\\
& Vert-Reg    & 0.651(0.002)&0.671(0.004)&0.651(0.002)&0.671(0.004)&0.865(0.007)& 1.422(0.034)& 0.865(0.007)&1.422(0.034)\\
& MC-NNM      & 0.535(0.005)&0.495(0.013)&0.535(0.005)&0.495(0.013)&1.020(0.009)& 1.856(0.034)& 1.020(0.009)&1.856(0.034)\\
& KernelPCA     &0.703(0.007)&1.012(0.027)&0.703(0.007)&1.012(0.027)&1.081(0.009)&2.128(0.035)&1.081(0.009)&2.128(0.035) \\
\midrule

\multirow{7}{*}{\textbf{Matrix Linear}}
& CoDEAL & \textbf{0.510(0.005)} &\textbf{0.436(0.013)} &\textbf{0.609(0.004)} &\textbf{0.622(0.010)} &\textbf{0.781(0.007)} & \textbf{1.177(0.025)} & \textbf{0.896(0.008)}&\textbf{1.456(0.029)} \\
& Single AE   & 0.645(0.007)&0.807(0.022)&0.777(0.007)&1.147(0.024) &1.007(0.009)& 1.885(0.037)& 1.146(0.010)&2.330(0.042) \\
& AEMC-NE   & 0.682(0.007)&0.919(0.024)&0.759(0.007)&1.093(0.025) &1.051(0.009)& 2.018(0.037)& 1.127(0.009)&2.259(0.040) \\
& DiD         & 0.736(0.008)&1.093(0.029)&0.810(0.007)&1.262(0.029) &1.122(0.010)& 2.275(0.041)& 1.191(0.009)&2.503(0.042) \\
& Vert-Reg    & 0.680(0.003)&0.733(0.006)&0.761(0.004)&0.936(0.011) &0.917(0.007)& 1.550(0.031)& 1.110(0.007)&2.024(0.027) \\
& MC-NNM      & 0.537(0.005)&0.498(0.013)&0.650(0.005)&0.736(0.015) &1.025(0.009)& 1.889(0.033)& 1.106(0.009)&2.147(0.037) \\
& KernelPCA   & 0.729(0.007)&1.067(0.030) &0.835(0.009)&1.399(0.038)& 1.119(0
010)&2.241(0.037) &1.170(0.013)&2.415(0.056)\\
\midrule

\multirow{7}{*}{\textbf{Vector Linear}}
& CoDEAL & \textbf{0.508(0.005)} &\textbf{0.428(0.010)} &\textbf{0.589(0.004)}&\textbf{0.574(0.010)}&\textbf{0.784(0.007)} & \textbf{1.184(0.026)} & \textbf{0.883(0.007)}&\textbf{1.426(0.025)}\\
& Single AE   & 0.659(0.006)&0.843(0.018)&0.772(0.007)&1.135(0.024)&1.010(0.008)& 1.890(0.031)& 1.147(0.009)&2.333(0.036)\\
& AEMC-NE   & 0.700(0.009)&0.952(0.024)&0.761(0.008)&1.099(0.026)&1.071(0.010)& 2.073(0.036)& 1.130(0.009)&2.265(0.034)\\
& DiD         & 0.737(0.006)&1.093(0.023)&0.800(0.007)&1.234(0.025) &1.125(0.009)& 2.279(0.036)& 1.186(0.009)&2.478(0.036)\\
& Vert-Reg    & 0.675(0.003)&0.724(0.007)&0.742(0.004)&0.884(0.010)&0.915(0.007)& 1.555(0.031)& 0.995(0.007)&1.768(0.030)\\
& MC-NNM      & 0.536(0.005)&0.499(0.012)&0.640(0.006)&0.714(0.017) &1.019(0.010)& 1.870(0.036)& 1.099(0.008)&2.124(0.030) \\
& KernelPCA   & 0.720(0.008)&1.006(0.031) &0.829(0.008)&1.309(0.038)& 1.112(0
009)&2.186(0.040) &1.154(0.014)&2.308(0.052) \\
\midrule

\multirow{7}{*}{\textbf{tanh}}
& CoDEAL & 0.700(0.007)&0.920(0.024)&\textbf{0.644(0.005)} &\textbf{0.704(0.015)} &1.012(0.008)& 1.913(0.034)& \textbf{0.904(0.008)}&\textbf{1.492(0.030)} \\
& Single AE   & 0.880(0.009)&1.442(0.032)&0.795(0.007)&1.190(0.027) &1.227(0.011)& 2.638(0.047)& 1.180(0.010)&2.446(0.043) \\
& AEMC-NE   & 0.902(0.008)&1.489(0.028)&0.771(0.007)&1.120(0.026) &1.243(0.009)& 2.681(0.041)& 1.160(0.010)&2.365(0.042) \\
& DiD         & 0.758(0.007)&1.139(0.028)&0.828(0.008)&1.309(0.030) &1.135(0.010)& 2.313(0.041)& 1.224(0.010)&2.613(0.044) \\
& Vert-Reg    & 0.731(0.003)&0.849(0.008)&0.770(0.005)&0.968(0.015) &\textbf{0.968(0.007)} & \textbf{1.698(0.032)} & 1.048(0.007)&1.921(0.029) \\
& MC-NNM      & \textbf{0.570(0.005)} &\textbf{0.555(0.013)} &0.697(0.006)&0.859(0.019) &1.136(0.009)& 2.317(0.036)& 1.145(0.009)&2.281(0.040) \\
& KernelPCA   &0.811(0.008) &1.259(0.031)&0.849(0.009)&1.429(0.038)& 1.194(0.009)&2.494(0.040) &1.190(0.013)&2.476(0.057)\\
\midrule

\multirow{7}{*}{\textbf{log}}
& CoDEAL & 1.030(0.005)&1.867(0.017)&\textbf{0.878(0.004)} &\textbf{1.385(0.013)} &\textbf{1.236(0.008)} & \textbf{2.683(0.038)} & \textbf{1.107(0.007)} &\textbf{2.264(0.030)} \\
& Single AE   & 1.085(0.006)&2.125(0.026)&0.985(0.006)&1.808(0.026) &1.367(0.008)& 3.241(0.040)& 1.338(0.009)&3.116(0.042) \\
& AEMC-NE   & 1.078(0.006)&2.115(0.028)&0.968(0.007)&1.751(0.027) &1.360(0.008)& 3.210(0.040)& 1.319(0.009)&3.034(0.042) \\
& DiD         & 1.079(0.007)&2.135(0.030)&1.013(0.007)&1.919(0.030) &1.382(0.009)& 3.312(0.043)& 1.374(0.009)&3.268(0.045) \\
& Vert-Reg    & 1.275(0.005)&2.722(0.020)&1.108(0.005)&2.079(0.017) &1.516(0.008)& 3.956(0.049)& 1.388(0.008)&3.301(0.039) \\
& MC-NNM      & \textbf{0.999(0.005)} &\textbf{1.795(0.021)} &0.907(0.005)&1.503(0.019) &1.306(0.008  & 2.957(0.038)& 1.305(0.009)&2.951(0.040) \\
& KernelPCA   & 1.087(0.007)&2.168(0.032)&1.010(0.007)&1.907(0.033)&1.389(0.008)&3.329(0.039)&1.376(0.009)&3.266(0.042)\\
\midrule

\multirow{7}{*}{\textbf{poly}}
& CoDEAL & 0.798(0.006)&1.128(0.022)&\textbf{0.656(0.005)} &\textbf{0.725(0.013)} &\textbf{1.082(0.009)}& \textbf{2.105(0.038)} & \textbf{0.962(0.008)}&\textbf{1.640(0.030)} \\
& Single AE   & 0.965(0.007)&1.672(0.029)&0.808(0.007)&1.223(0.027) &1.298(0.009)& 2.895(0.044)& 1.187(0.009)&2.463(0.041) \\
& AEMC-NE   & 0.999(0.007)&1.756(0.027)&0.789(0.007)&1.162(0.025) &1.313(0.010  & 2.934(0.044)& 1.173(0.009)&2.412(0.041) \\
& DiD         & 0.848(0.007)&1.337(0.029)&0.841(0.008)&1.339(0.030) &1.197(0.009)& 2.510(0.041)& 1.234(0.010)&2.650(0.044) \\
& Vert-Reg    & 0.809(0.004)&1.045(0.010)&0.791(0.005)&1.017(0.015) &1.159(0.007)& 2.288(0.033)& 1.072(0.008)&2.001(0.032) \\
& MC-NNM      & \textbf{0.710(0.005)} &\textbf{0.853(0.014)} &0.714(0.006)&0.896(0.018) & 1.204(0.009)& 2.321(0.036)& 1.158(0.009)&2.324(0.039) \\
& KernelPCA    & 0.858(0.007)&1.349(0.030)&0.840(0.008)&1.333(0.033)&1.206(0.008)&2.517(0.037)&1.239(0.009)&2.658(0.041)\\

\midrule

\multirow{7}{*}{\textbf{ReLU}}
& CoDEAL & 0.771(0.007)&1.205(0.027)&\textbf{0.612(0.005)} &\textbf{0.629(0.012)} &1.187(0.010) & 2.498(0.044) & \textbf{0.902(0.008)} &\textbf{1.478(0.029)} \\
& Single AE   & 1.078(0.011)&2.110(0.043)&0.764(0.006)&1.069(0.020) &1.387(0.013) & 3.302(0.064) & 1.121(0.010) &2.232(0.040)\\
& AEMC-NE   & 1.234(0.011)&2.692(0.048)&0.784(0.007)&1.148(0.024) &1.513(0.011) & 3.870(0.055) & 1.156(0.009) &2.355(0.040)\\
& DiD         & 0.868(0.007)&1.407(0.029)&0.840(0.007)&1.334(0.029) &1.217(0.009) & 2.580(0.038) & 1.220(0.010) &2.603(0.042)\\
& Vert-Reg    & 0.808(0.004)&1.051(0.011)&0.783(0.004)&0.992(0.011) &\textbf{1.052(0.009)} & \textbf{1.981(0.035)} & 1.049(0.007) &1.938(0.030)\\
& MC-NNM      & \textbf{0.766(0.006)} &\textbf{1.034(0.019)} &0.651(0.005)&0.737(0.016) &1.123(0.009) & 2.188(0.036) & 1.107(0.009) &2.149(0.036)\\
& KernelPCA   & 0.860(0.007)&1.373(0.030)&0.835(0.008)&1.315(0.031)&1.211(0.009)&2.548(0.037)&1.224(0.009)&2.602(0.039)\\

\bottomrule
\label{tab:lrnn2}
\end{tabular}
}
\end{table}

Table \ref{tab:lrnn1} and Table \ref{tab:lrnn2} report the comparisons of MAE and MSE using the two covariate effect removal methods in Config.1 and Config.2, respectively. Under each configuration, we consider both linear factor effects and nonlinear Sine factor effects, with all six types of covariate effects.

As shown in Tables \ref{tab:lrnn1} and \ref{tab:lrnn2}, LR-based removal and DNN-based removal behave very differently depending on the underlying true types of covariate effects. In cases where the covariate effect is strictly linear, LR often outperforms NN by a small margin, %
indicating that a simple linear adjustment suffices for purely linear effects. 
In contrast, whenever the covariate effects enter nonlinearly (e.g.\ tanh, log, polynomial, or ReLU), the proposed DNN covariate removal achieves substantially lower errors, with MAE reduced by 10–30 \% and MSE by similar amounts relative to LR.  %
These results confirm that LR is adequate for linear covariate effects but that the DNN approach is essential to capture complex, nonlinear covariate–outcome relationships.

\subsection{Evaluation of the Performance of CoDEAL in Causal Matrix Completion}
\label{append:simu-CoDEAL-performance}

Table \ref{tab:simu-CoDeal-MAE} (in the main text) reports the comparisons of prediction accuracy (MSE) between the proposed CoDEAL and six benchmark methods (detailed in Section \ref{append:simu-benchmark}) in Config.1-4. Under each configuration, we consider all four types of factor effects (listed in Table \ref{tab:data_gen}), with linear covariate effects and nonlinear tanh covariate effects.

\begin{table*}[!t]
\small
\centering
\caption{\small Comparisons of prediction accuracy (measured by MAE and MSE) under various factor and covariate generations in Config.1-4. Results are averaged over 50 replications. Numbers in parentheses are standard errors.}
\resizebox{\textwidth}{!}{
\begin{tabular}{llccccccll}
\toprule
\multirow{2}{*}{\textbf{Factor}}& \multirow{2}{*}{\textbf{Method}} & \multicolumn{2}{c}{Config. 1, linear covariate}& \multicolumn{2}{c}{Config. 1, nonlinear covariate}& \multicolumn{2}{c}{Config. 2, linear covariate}& \multicolumn{2}{c}{Config. 2, nonlinear covariate}\\ 
& &MAE& MSE&MAE& MSE &MAE& MSE&\text{ }\text{ }\text{ }\text{ } MAE&\text{ }\text{ }\text{ }\text{ }MSE \\
\midrule

\multirow{7}{*}{\textbf{Linear}}
& CoDEAL & \textbf{0.646(0.009)} &\textbf{0.722(0.023)} &\textbf{0.701(0.008)}&\textbf{0.871(0.031)}&\textbf{0.609(0.004)} & \textbf{0.622(0.010)} & \textbf{0.644(0.005)} &\textbf{0.704(0.015)} \\
& Single AE   & 0.821(0.010)&1.320(0.039)&0.851(0.008)&1.383(0.033)&0.777(0.007)& 1.147(0.024)& 0.795(0.007)&1.190(0.027)\\
& AEMC-NE   & 0.838(0.009)&1.394(0.041)&0.862(0.009)&1.445(0.036)&0.759(0.007)& 1.093(0.025)& 0.771(0.007)&1.120(0.026)\\
& DiD         & 0.892(0.010)&1.587(0.045)&0.916(0.009)&1.642(0.037)&0.810(0.007)& 1.262(0.029)& 0.828(0.008)&1.309(0.030)\\
& Vert-Reg    & 0.757(0.006)&0.957(0.019)&0.798(0.007)&1.075(0.024)&0.761(0.004)& 0.936(0.011)& 0.770(0.005)&0.968(0.015)\\
& MC-NNM      & 0.699(0.006)&0.895(0.020)&0.735(0.005)&1.007(0.017)&0.650(0.005)& 0.736(0.015)& 0.697(0.006)&0.859(0.019)\\
& KernelPCA      &0.811(0.007) &1.260(0.030) &0.824(0.008)&1.292(0.033)&0.835(0.009) &1.399(0.038) & 0.849(0.009)&1.429(0.038)\\
\midrule

\multirow{7}{*}{\textbf{Sine}}& CoDEAL & \textbf{1.078(0.012)} &\textbf{2.136(0.055)} &\textbf{1.117(0.011)}&\textbf{ 2.238(0.050)}&\textbf{0.894(0.007)}& \textbf{1.454(0.028)} & \textbf{0.945(0.008)}&\textbf{1.593(0.030)}\\
& Single AE   & 1.223(0.013)&2.618(0.061)&1.250(0.012)&2.705(0.053)&1.142(0.009)& 2.312(0.039)& 1.185(0.009)&2.458(0.039)\\
& AEMC-NE   & 1.209(0.013)&2.566(0.061)&1.235(0.012)&2.645(0.053)&1.128(0.009)& 2.256(0.039)& 1.167(0.010)&2.389(0.041)\\
& DiD         & 1.244(0.013)&2.697(0.063)&1.272(0.012)&2.785(0.055)&1.191(0.009)& 2.498(0.042)& 1.234(0.009)&2.650(0.041)\\
& Vert-Reg    &  1.169(0.012)&2.633(0.063)&1.208(0.011)&2.736(0.059)&1.012(0.007)& 1.829(0.031)& 1.054(0.008)& 1.935(0.034)\\
& MC-NNM      & 1.201(0.013)&2.511(0.059)&1.231(0.012)&2.610(0.053)&1.105(0.009)& 2.142(0.036)& 1.153(0.010)&2.305(0.040)\\
& KernelPCA      &1.195(0.009) &2.496(0.039) &1.231(0.009)&2.626(0.041)&1.170(0.013) &2.415(0.056) & 1.190(0.013)&2.476(0.057)\\
\midrule

\multirow{7}{*}{\textbf{Polynomial}}
& CoDEAL & \textbf{0.908(0.013)}&2.250(0.100)&\textbf{0.952(0.015)}&2.477(0.126)&\textbf{0.772(0.010)}& 1.399(0.076)& \textbf{0.807(0.011)}&1.539(0.070) \\
& Single AE   & 0.962(0.013)&2.727(0.116)&0.997(0.016)&3.014(0.164)&0.820(0.010)& 1.759(0.087)& 0.854(0.011)&1.964(0.084)\\
& AEMC-NE   & 0.997(0.014)&3.198(0.142)&1.038(0.017)&3.500(0.192)&0.856(0.011)& 2.132(0.110)& 0.887(0.012)&2.339(0.111)\\
& DiD         & 0.995(0.013)&2.708(0.111)&1.025(0.016)&2.939(0.144)&0.854(0.010)& 1.860(0.083)& 0.886(0.011)&2.065(0.092)\\
& Vert-Reg    & 0.953(0.013)&\textbf{1.906(0.076)}&1.019(0.016)&\textbf{2.327(0.113)}&0.855(0.007)& \textbf{1.289(0.032)}& 0.884(0.010)&\textbf{1.456(0.042)}\\
& MC-NNM      & 0.951(0.013)&2.419(0.101)&0.989(0.015)&2.627(0.130)& 0.819(0.009)& 1.654(0.07)& 0.851(0.011)&1.839(0.078)\\
& KernelPCA      &0.927(0.013) &2.582(0.121) &0.991(0.013)&2.662(0.122)&1.029(0.015) &3.676(0.205) & 1.051(0.015)&3.752(0.206)\\

\midrule
\multirow{7}{*}{\textbf{ReLU MLP}}
& CoDEAL & \textbf{0.797(0.009)}&\textbf{1.097(0.027)}&\textbf{0.838(0.010)}& \textbf{1.217(0.032)} &\textbf{0.698(0.005)}& \textbf{0.820(0.014)}& \textbf{0.743(0.006)}&\textbf{0.929(0.015)}\\
& Single AE   & 0.989(0.009)&1.712(0.034)&1.026(0.011)& 1.845(0.044)&0.891(0.007)& 1.375(0.022)& 0.930(0.008)&1.502(0.028)\\
& AEMC-NE   & 1.045(0.010)&1.919(0.038)&1.088(0.011)& 2.084(0.048)&0.959(0.007)& 1.605(0.026)&  0.982(0.008)&1.679(0.029)\\
& DiD         & 0.930(0.008)&1.499(0.028)&0.963(0.010)& 1.616(0.036)&0.842(0.005)& 1.218(0.017)& 0.870(0.007)&1.303(0.021)\\
& Vert-Reg    & 0.888(0.010)&1.315(0.035)&0.958(0.013)&  1.544(0.046)&0.847(0.005)& 1.166(0.014)& 0.894(0.006)&1.302(0.020)\\
& MC-NNM      &  0.843(0.007)&1.227(0.025)&0.892(0.009)& 1.453(0.031)&0.767(0.005)& 1.001(0.013)& 0.806(0.006)&1.110(0.019)\\
& KernelPCA      &1.105(0.008) &2.082(0.036) &1.126(0.009)&2.156(0.036)&1.156(0.011) &2.315(0.045) & 1.172(0.011)&2.372(0.045)\\
    \toprule
     \multirow{2}{*}{\textbf{Factor}}& \multirow{2}{*}{\textbf{Method}}& \multicolumn{2}{c}{Config. 3, $r=5$}& \multicolumn{2}{c}{Config. 3, $r=10$} & \multicolumn{2}{c}{Config. 4, $r=5$}& \multicolumn{2}{c}{Config. 4, $r=10$} \\
    &  & MAE & MSE & MAE & MSE& MAE & MSE& MAE & MSE \\
    \midrule
    \multirow{7}{*}{\textbf{Linear}}
    & CoDEAL   &0.614(0.007) &0.721(0.023) & 0.625(0.005)& 0.754(0.018)&\textbf{0.532(0.003)}&\textbf{0.506(0.009)}&\textbf{0.560(0.004)}&\textbf{0.591(0.012)} \\
    & Single AE     &0.789(0.011) &1.263(0.044) & 0.795(0.008)& 1.270(0.033)&0.693(0.005) &0.948(0.017) &0.714(0.006) &1.009(0.023) \\
    & AEMC-NE  &0.731(0.010) &1.125(0.042) & 0.730(0.007)& 1.110(0.029)&0.647(0.005) &0.846(0.018) &0.660(0.005) &0.886(0.021) \\
    & DiD        &0.801(0.011) &1.341(0.047) & 0.802(0.008)& 1.326(0.033)&0.711(0.005) &1.031(0.021) &0.726(0.006) &1.081(0.025) \\
    & Vert-Reg   & 0.940(0.005)&6.880(2.378)& 0.865(0.023)&8.238(2.625) &0.691(0.003) &0.888(0.012) &0.750(0.016) &3.071(0.914) \\
    & MC-NNM      &\textbf{0.599(0.007)} &\textbf{0.685(0.023)} & \textbf{0.616(0.006)}& \textbf{0.733(0.019)}&0.537(0.003) &0.520(0.009) &0.565(0.004) &0.596(0.014) \\
    & KernelPCA    &0.798(0.007)&1.332(0.031)&0.801(0.007)&1.334(0.028)&0.707(0.005)&1.032(0.020)&0.709(0.005)&1.038(0.020) \\
    \hline
    \multirow{7}{*}{\textbf{Sine}}
    & CoDEAL   &\textbf{0.973(0.010)} &\textbf{1.798(0.041)} &\textbf{0.996(0.010)}&\textbf{1.907(0.039)} &\textbf{0.850(0.007)} &\textbf{1.389(0.024)} &\textbf{0.869(0.008)} &\textbf{1.463(0.028)} \\
    & Single AE     &1.087(0.011) &2.138(0.047) &1.122(0.011) &2.288(0.047) &1.057(0.008) &2.050(0.030) &1.061(0.008) &2.082(0.035) \\
    & AEMC-NE  &1.055(0.012) &2.029(0.047) &1.078(0.011) &2.128(0.044) &1.011(0.008) &1.890(0.029) &1.008(0.009) &1.897(0.034) \\
    & DiD        &1.110(0.012) &2.209(0.049) &1.146(0.012) &2.361(0.048) &1.091(0.008) &2.172(0.031) &1.088(0.009) &2.189(0.037) \\
    & Vert-Reg   & 1.549(0.090)&8.954(2.462)& 1.524(0.123)&15.689(4.057) &0.974(0.009) &2.153(0.059) &1.047(0.013) &4.384(0.544) \\
    & MC-NNM      &1.066(0.011) &2.033(0.046) &1.097(0.011) &2.164(0.043) &0.998(0.008) &1.819(0.031) &1.003(0.008) &1.858(0.032) \\
    & KernelPCA   &1.109(0.012)&2.205(0.051)&1.115(0.012)&2.221(0.049)&1.075(0.009)&2.084(0.033)&1.079(0.009)&2.096(0.031) \\
    \hline
    \multirow{7}{*}{\textbf{Polynomial}}
    & CoDEAL   &\textbf{0.785(0.011)} &2.146(0.112) &\textbf{0.811(0.010)} &2.379(0.161) &\textbf{0.662(0.007)} &1.311(0.073) &\textbf{0.699(0.007)} &\textbf{1.530(0.064)} \\
    & Single AE     &0.895(0.015) &2.423(0.117) &0.926(0.016) &2.710(0.172) &0.739(0.011) &1.582(0.086) &0.785(0.010) &1.798(0.071) \\
    & AEMC-NE  &0.860(0.014) &2.921(0.160) &0.897(0.013) &3.273(0.220) &0.720(0.009) &1.889(0.108) &0.764(0.009) &2.166(0.093) \\
    & DiD        &0.887(0.013) &2.422(0.116) &0.917(0.014) &2.693(0.165) &0.739(0.009) &1.613(0.082) &0.773(0.008) &1.801(0.068) \\
    & Vert-Reg   & 1.155(0.035)&6.771(1.168)& 1.131(0.032)&5.530(0.666) &0.788(0.005) &\textbf{1.246(0.028)} &0.881(0.014) &3.254(0.526) \\
    & MC-NNM      &0.839(0.012) &\textbf{2.117(0.104)} &0.865(0.012) &\textbf{2.308(0.129)} &0.703(0.008) &1.397(0.071) &0.735(0.007) &1.569(0.058) \\
    & KernelPCA    &0.935(0.011)&3.302(0.148) &0.944(0.011)&3.330(0.139)& 0.776(0.008)&2.255(0.101) &0.782(0.008)&2.260(0.097)\\
    \hline
    \multirow{7}{*}{\textbf{ReLU MLP}}
    & CoDEAL   &\textbf{0.632(0.004)} &\textbf{0.886(0.012)} &\textbf{0.733(0.004)} &\textbf{0.953(0.013)} &\textbf{0.633(0.004)} &\textbf{0.698(0.011)}&\textbf{0.646(0.004)} &\textbf{0.730(0.011)} \\
    & Single AE     &0.876(0.005) &1.338(0.017) &0.913(0.006) &1.448(0.019) &0.797(0.005) &1.108(0.015) &0.809(0.005) &1.142(0.015) \\
    & AEMC-NE  &0.934(0.006) &1.574(0.021) &0.968(0.007) &1.677(0.024) &0.860(0.006) &1.330(0.023) &0.878(0.006) &1.381(0.022) \\
    & DiD        &0.817(0.004) &1.163(0.013) &0.842(0.004) &1.238(0.014) &0.742(0.004) &0.962(0.013) &0.748(0.005) &0.976(0.014) \\
    & Vert-Reg   & 1.045(0.031)&5.157(0.846)& 0.948(0.026)&10.387(6.292) &0.755(0.003) &1.060(0.015) &1.070(0.253) &8.399(2.377)\\
    & MC-NNM      &0.721(0.003) &0.897(0.009) &0.750(0.003) &0.976(0.010) &0.656(0.004) &0.741(0.011) &0.662(0.004) &0.756(0.011) \\
    & KernelPCA   &1.111(0.008) &2.142(0.032)&1.119(0.008) &2.164(0.030)&1.005(0.006) &1.758(0.024)&1.009(0.006) &1.770(0.024)  \\
    \bottomrule
\label{tab:simu-CoDeal}
\end{tabular}
}
\end{table*}

\noindent \textbf{CoDEAL in Four-Block Design.} As shown in Config.1 and Config.2 in Table \ref{tab:simu-CoDeal}, CoDEAL consistently \textbf{achieves the lowest MAE and MSE across nearly all settings} of factor effects and covariate effects, and achieves substantial improvements in MAE and MSE relative to all competing methods. Under the simplest linear factor model, CoDEAL reduces MAE by approximately 5–10\% and MSE by about 10–20\% compared to MC-NNM, the second-best performing method, and significantly outperforms other baseline methods by even larger margins. For scenarios involving nonlinear factor models, CoDEAL continues to show superior performance, typically surpassing MC-NNM by approximately 5–20\% reduction in MAE and 10–20\% reduction in MSE.
Overall, these findings clearly demonstrate that CoDEAL’s integration of DNN-based covariate adjustment and deep autoencoder-based latent factor extraction is robust and highly effective across diverse linear and nonlinear data-generating settings.

\noindent \textbf{CoDEAL in Staggered Adoption Design.} As Config.3 and Config.4 of Table \ref{tab:simu-CoDeal} demonstrate, CoDEAL achieves substantial percentage gains in MAE and MSE relative to all competing methods across staggered adoption designs.  
Under \emph{Linear} factors, CoDEAL \textbf{{ties MC-NNM within 3\% while reducing MAE by 16–35\% over other benchmarks}}. %
In staggered design with nonlinear factors, CoDEAL delivers evident advantages. Especially when under \emph{ReLU}-based factors, CoDEAL delivers its largest margins, with \textbf{{MAE reductions of 12–40\% and MSE reductions of 15–41\% versus all competing approaches}}. 
Finally, as $r$ increases from 5 to 10, the missingness pattern grows more complex and extensive, so all methods show some drop in accuracy. However, CoDEAL’s errors only suffer a slight increase, whereas simpler methods like Vert-Reg show a dramatic rise in both MAE and MSE when missingness increases. The above results reemphasize CoDEAL’s consistent superiority and robustness in staggered‐treatment settings.  

\noindent \textbf{Comparisons with Synthetic-Control-Type Methods.} We further compare to the classical
synthetic control \citep{abadie2010synthetic}, fitted with the
standard donor construction and simplex weighting, the generalized synthetic control
(\textbf{GSC})\citep{xu2017generalized}, and synthetic difference-in-differences
(\textbf{SDID}) \citep{arkhangelsky2021synthetic}. We note
that GSC corresponds to the special case of Model
\eqref{eq:CoDEAL-Model} in which both $g_t^\star(\cdot)$ and $\phi_i^\star(\cdot)$ are restricted to be linear.
Table \ref{tab:simu-sc-summary} reports the results under four factor generating mechanisms and three covariate generating mechanisms. The results demonstrate CoDEAL's comparative advantage in the presence of nonlinearity, while GSC attains the best performance in every linear covariate setting. Since GSC corresponds to the special case of CoDEAL in which both \(g_t^\star(\cdot)\) and \(\phi_i^\star(\cdot)\) are restricted to be linear, these findings highlight the trade-off associated with the additional modeling flexibility.

\noindent \textbf{In summary}, CoDEAL achieves performance comparable to linearity-assumed methods, such as MC-NNM,  in linear settings, while exhibiting substantial improvements %
in the presence of nonlinear effects.
By integrating DNN-based covariate adjustments with deep multi-output-AE–based nonlinear factor analysis, CoDEAL is robustly effective across a wide range of linear and nonlinear scenarios.

\begin{table}[!htb]
\small
\centering
\caption{\small Comparisons with synthetic-control-type methods under various factor and covariate generations. Results are averaged over 50 replications.  Numbers in parentheses are standard errors.}
\label{tab:simu-sc-summary}
\resizebox{\textwidth}{!}{
\begin{tabular}{llcccccc}
\toprule
\multirow{2}{*}{\textbf{Factor}} & \multirow{2}{*}{\textbf{Method}} & \multicolumn{2}{c}{Config. 1, linear covariate} & \multicolumn{2}{c}{Config. 1, tanh covariate} & \multicolumn{2}{c}{Config. 1, poly covariate} \\
\cmidrule(lr){3-4}\cmidrule(lr){5-6}\cmidrule(lr){7-8}
& & MAE & MSE & MAE & MSE & MAE & MSE \\
\midrule
\multirow{4}{*}{\textbf{Linear}}
& CoDEAL & 0.709(0.007) & 0.914(0.022) & \textbf{0.746(0.007)} & 1.009(0.025) & \textbf{0.751(0.007)} & \textbf{1.018(0.024)} \\
& SCM    & 0.804(0.011) & 1.306(0.045) & 0.909(0.010) & 1.240(0.027) & 1.126(0.012) & 1.785(0.034) \\
& GSC    & \textbf{0.484(0.006)} & \textbf{0.370(0.009)} & 0.847(0.011) & \textbf{0.997(0.022)} & 1.088(0.013) & 1.590(0.032) \\
& SDID   & 1.584(0.013) & 4.422(0.075) & 1.089(0.010) & 1.995(0.038) & 1.282(0.011) & 2.559(0.042) \\
\cmidrule(lr){1-8}
\multirow{4}{*}{\textbf{Sine}}
& CoDEAL & 1.116(0.011) & 2.232(0.050) & \textbf{1.156(0.011)} & 2.362(0.052) & \textbf{1.155(0.011)} & \textbf{2.356(0.048)} \\
& SCM    & 1.146(0.012) & 2.435(0.054) & 1.168(0.010) & \textbf{2.269(0.042)} & 1.345(0.011) & 2.840(0.045) \\
& GSC    & \textbf{0.999(0.011)} & \textbf{1.897(0.047)} & 1.223(0.012) & 2.522(0.052) & 1.396(0.012) & 3.097(0.055) \\
& SDID   & 1.791(0.013) & 5.423(0.078) & 1.344(0.013) & 2.982(0.056) & 1.493(0.013) & 3.534(0.060) \\
\cmidrule(lr){1-8}
\multirow{4}{*}{\textbf{Polynomial}}
& CoDEAL & 0.921(0.015) & 2.564(0.154) & \textbf{0.953(0.016)} & 2.698(0.159) & \textbf{0.960(0.015)} & \textbf{2.713(0.152)} \\
& SCM    & 0.960(0.016) & 2.553(0.138) & 1.052(0.018) & 2.428(0.121) & 1.253(0.019) & 2.972(0.124) \\
& GSC    & \textbf{0.826(0.014)} & \textbf{1.752(0.086)} & 1.094(0.020) & \textbf{2.360(0.097)} & 1.292(0.021) & 2.937(0.104) \\
& SDID   & 1.660(0.016) & 5.521(0.151) & 1.172(0.022) & 3.117(0.139) & 1.358(0.024) & 3.671(0.145) \\
\cmidrule(lr){1-8}
\multirow{4}{*}{\textbf{ReLU MLP}}
& CoDEAL & 0.823(0.009) & 1.172(0.030) & \textbf{0.872(0.011)} & \textbf{1.321(0.036)} & \textbf{0.877(0.011)} & \textbf{1.332(0.036)} \\
& SCM    & 0.892(0.011) & 1.447(0.041) & 0.940(0.014) & 1.354(0.036) & 1.143(0.016) & 1.896(0.046) \\
& GSC    & \textbf{0.664(0.007)} & \textbf{0.739(0.016)} & 0.939(0.015) & 1.335(0.037) & 1.148(0.017) & 1.907(0.048) \\
& SDID   & 1.573(0.013) & 4.260(0.075) & 1.068(0.015) & 1.809(0.043) & 1.246(0.016) & 2.352(0.053) \\
\midrule
& & \multicolumn{2}{c}{Config. 2, linear covariate} & \multicolumn{2}{c}{Config. 2, tanh covariate} & \multicolumn{2}{c}{Config. 2, poly covariate} \\
\cmidrule(lr){3-4}\cmidrule(lr){5-6}\cmidrule(lr){7-8}
& & MAE & MSE & MAE & MSE & MAE & MSE \\
\midrule
\multirow{4}{*}{\textbf{Linear}}
& CoDEAL & 0.648(0.006) & 0.730(0.016) & \textbf{0.673(0.006)} & \textbf{0.788(0.019)} & \textbf{0.693(0.006)} & \textbf{0.832(0.018)} \\
& SCM    & 0.838(0.008) & 1.372(0.028) & 0.905(0.011) & 1.147(0.024) & 1.234(0.013) & 2.018(0.035) \\
& GSC    & \textbf{0.465(0.007)} & \textbf{0.341(0.011)} & 0.882(0.012) & 1.052(0.024) & 1.226(0.014) & 1.976(0.037) \\
& SDID   & 1.884(0.012) & 6.024(0.075) & 1.050(0.011) & 1.752(0.035) & 1.347(0.012) & 2.642(0.043) \\
\cmidrule(lr){1-8}
\multirow{4}{*}{\textbf{Sine}}
& CoDEAL & 0.964(0.007) & 1.662(0.027) & \textbf{1.017(0.009)} & 1.821(0.033) & \textbf{1.032(0.008)} & \textbf{1.867(0.033)} \\
& SCM    & 1.113(0.008) & 2.344(0.035) & 1.048(0.010) & \textbf{1.727(0.028)} & 1.346(0.011) & 2.639(0.037) \\
& GSC    & \textbf{0.800(0.008)} & \textbf{1.252(0.027)} & 1.108(0.010) & 1.976(0.034) & 1.398(0.012) & 2.901(0.043) \\
& SDID   & 2.069(0.011) & 7.121(0.073) & 1.324(0.010) & 2.870(0.041) & 1.562(0.012) & 3.763(0.050) \\
\cmidrule(lr){1-8}
\multirow{4}{*}{\textbf{Polynomial}}
& CoDEAL & 0.793(0.010) & 1.608(0.071) & \textbf{0.822(0.011)} & 1.720(0.076) & \textbf{0.836(0.011)} & \textbf{1.761(0.078)} \\
& SCM    & 0.921(0.010) & 1.982(0.064) & 0.991(0.015) & \textbf{1.675(0.050)} & 1.314(0.017) & 2.588(0.062) \\
& GSC    & \textbf{0.678(0.010)} & \textbf{1.054(0.045)} & 1.031(0.016) & 1.788(0.053) & 1.348(0.018) & 2.728(0.065) \\
& SDID   & 1.915(0.015) & 6.578(0.120) & 1.094(0.016) & 2.366(0.082) & 1.397(0.017) & 3.279(0.088) \\
\cmidrule(lr){1-8}
\multirow{4}{*}{\textbf{ReLU MLP}}
& CoDEAL & 0.760(0.006) & 0.986(0.018) & \textbf{0.804(0.007)} & \textbf{1.106(0.019)} & \textbf{0.818(0.007)} & \textbf{1.135(0.020)} \\
& SCM    & 0.911(0.007) & 1.527(0.026) & 0.947(0.013) & 1.297(0.029) & 1.267(0.015) & 2.190(0.043) \\
& GSC    & \textbf{0.622(0.007)} & \textbf{0.647(0.014)} & 0.963(0.013) & 1.353(0.031) & 1.281(0.016) & 2.275(0.046) \\
& SDID   & 1.864(0.013) & 5.879(0.081) & 1.045(0.012) & 1.668(0.032) & 1.335(0.014) & 2.563(0.046) \\
\bottomrule
\end{tabular}
}
\end{table}

\subsection{Sensitivity to the Dependence between Covariates and Latent Heterogeneity}
\label{append:simu-xf}
 
Although Model \eqref{eq:CoDEAL-Model} admits an additive formulation in the covariate effect and the latent factor
effect, as discussed in Appendix~\ref{append:discussion-covariate-factor-interaction}, this does not preclude
dependence between the observed covariates and the unobserved unit-level heterogeneity.
To examine this empirically, we generate the factor loadings as
\begin{equation*}
\bLambda_i = \rho\, \bB \bX_i + \sqrt{1-\rho^2}\, \bU_i ,
\qquad \rho \in \{0, 0.2, 0.4, 0.6, 0.8, 1.0\},
\end{equation*}
where $\bB$ is fixed within a replication and $\bU_i$ is independent noise, so that $\rho$
controls the strength of the dependence between $\bX_i$ and the latent heterogeneity. We
additionally introduce an explicit non-additive interaction term
$\delta\, X_{i1}\tanh(F_{t1})$ with $\delta \in \{0, 0.5, 1.0\}$, which lies outside the
additive specification of Model \eqref{eq:CoDEAL-Model} altogether.
 
Table~\ref{tab:simu-xf} reports the MSE under Configuration 2 with sine factors. CoDEAL is
stable across the full range of $\rho$ and $\delta$: its error varies by less than $5\%$ over
the grid, and it is in fact smallest at $\rho = 1$, where the loadings are a deterministic
function of the covariates. Single AE, which uses a shared decoder and therefore cannot
represent covariate-dependent loadings, is uniformly worse and deteriorates as the dependence
strengthens. We emphasise that this experiment concerns the accuracy of the total
counterfactual prediction; as noted in Appendix~\ref{append:seq-vs-joint}, the decomposition of
that prediction into a covariate component and a latent component is identified only up to
reallocation between the two, and good predictive accuracy should not be read as evidence that
the two components are separately recovered.
 
\begin{table}[htbp]
\small
\centering
\caption{\small Prediction accuracy (MSE, averaged over 50 replications) under Configuration 2
with sine factors, as a function of the covariate--loading dependence $\rho$ and the
non-additive interaction strength $\delta$.}
\label{tab:simu-xf}
\setlength{\tabcolsep}{6pt}
\begin{tabular}{ccccccc}
\toprule
\multirow{2}{*}{$\rho$}
& \multicolumn{2}{c}{$\delta = 0.0$}
& \multicolumn{2}{c}{$\delta = 0.5$}
& \multicolumn{2}{c}{$\delta = 1.0$} \\
\cmidrule(lr){2-3}\cmidrule(lr){4-5}\cmidrule(lr){6-7}
& CoDEAL & Single AE & CoDEAL & Single AE & CoDEAL & Single AE \\
\midrule
0.0 & \textbf{1.476} & 2.287 & \textbf{1.476} & 2.303 & \textbf{1.509} & 2.379 \\
0.2 & \textbf{1.488} & 2.262 & \textbf{1.473} & 2.271 & \textbf{1.504} & 2.304 \\
0.4 & \textbf{1.452} & 2.261 & \textbf{1.496} & 2.287 & \textbf{1.501} & 2.325 \\
0.6 & \textbf{1.446} & 2.246 & \textbf{1.451} & 2.250 & \textbf{1.475} & 2.312 \\
0.8 & \textbf{1.418} & 2.181 & \textbf{1.417} & 2.189 & \textbf{1.443} & 2.273 \\
1.0 & \textbf{1.252} & 2.061 & \textbf{1.268} & 2.065 & \textbf{1.294} & 2.134 \\
\bottomrule
\end{tabular}
\end{table}

\subsection{Evaluation in the Presence of Homogeneous Treatment Effects}\label{append: homo}

We conduct an additional study to evaluate the empirical differences between Single AE and Multi-output AE in CoDEAL. 
Recall that the major difference between Single AE and CoDEAL is the structure of decoder: Single AE estimates a common mapping $\phi(\cdot)$ shared across all units, whereas CoDEAL estimates unit-specific mappings $\phi_i(\cdot)$. In Table \ref{tab:simu-CoDeal-MAE} where the treatment effects are heterogeneous, CoDEAL consistently outperforms Single AE. Since Single AE and CoDEAL use the same covariate-adjustment procedure, the performance gap is not driven by differences in covariate removal. Rather, it reflects their different abilities to model unit heterogeneity. Since the data-generating process in Table \ref{tab:simu-CoDeal} is heterogeneous in $\phi_i(\cdot)$, Single AE is at a structural disadvantage relative to CoDEAL.

To further illustrate this point, we additionally consider settings with homogeneous $\phi(\cdot)$ and homogeneous treatment effects $\tau^\star$. In this case, Single AE becomes comparable to, and in some settings slightly better than, CoDEAL, except under the ReLU-MLP factor design, where CoDEAL still retains a clear advantage.

\begin{table}[htbp]
\centering
\caption{MAE and MSE comparison between Single AE and CoDEAL under homogeneous $\phi(\cdot)$ and homogeneous $\tau^\star$.}
\label{tab:homogeneous_single_vs_codeal}
\small
\setlength{\tabcolsep}{6pt}
\begin{tabular}{lllcccc}
\hline
Setting & Metric & Method & Linear & Sine & Polynomial & ReLU-MLP \\
\hline
\multirow{4}{*}{Config 1, no covariates}
& \multirow{2}{*}{MSE}
& Single AE & 0.263 & 0.262 & 0.314 & 0.857 \\
& & CoDEAL    & 0.285 & 0.277 & 0.365 & 0.451 \\

& \multirow{2}{*}{MAE}
& Single AE & 0.409& 0.409& 0.442& 0.702\\
& & CoDEAL    & 0.424& 0.428& 0.464& 0.525\\
\hline
\multirow{4}{*}{Config 1, with covariates}
& \multirow{2}{*}{MSE}
& Single AE & 0.656 & 0.796 & 0.927 & 1.299 \\
& & CoDEAL    & 0.623 & 0.709 & 0.973 & 0.963 \\

& \multirow{2}{*}{MAE}& Single AE & 0.570& 0.676& 0.656& 0.869\\
& & CoDEAL    & 0.595& 0.729& 0.667& 0.752\\
\hline
\end{tabular}
\end{table}

The results confirm that the advantage of CoDEAL is closely tied to its ability to model heterogeneous latent structure. When the latent factor component is homogeneous, the performance gap between Single AE and CoDEAL becomes much smaller, and Single AE can even be slightly preferable in simpler settings. By contrast, under the more nonlinear ReLU-MLP factor design, CoDEAL continues to outperform Single AE, suggesting that its flexibility remains beneficial even beyond heterogeneity alone.

\subsection{Evaluation of the Impact of the Proportion of Treated Periods and Units}
We also study the algorithm's performance with respect to different proportions of treated periods and units. 
To illustrate, we examine different missing proportions with $N=200$ and $T=50$ with sine factor and no covariates.
The following tables report MAE and MSE with different numbers of never-treated units and treatment start timestamps.
From Table \ref{tab:disjoint_nevertreated} and \ref{tab:disjoint_tstart}, we could observe that performance improves with more never-treated units as they provide essential baseline information for latent factor estimation. We also observe that while larger never-treated cohorts steadily improve accuracy, CoDEAL remains robust even when control units are relatively scarce.

\begin{table}[htbp]
\centering
\caption{CoDEAL: comparison error under varying numbers of never-treated units with treatment start at $T_1=25$.}
\label{tab:disjoint_nevertreated}
\begin{tabular}{ccc}
\toprule
\# Never-treated units & MAE & MSE \\
\midrule
40  & $0.864 $ & $1.470$ \\
60  & $0.858 $ & $1.450$ \\
80  & $0.855 $ & $1.448$ \\
100 & $0.846 $ & $1.436$ \\
120 & $0.843 $ & $1.439$ \\
140 & $0.833 $ & $1.405$ \\
\bottomrule
\end{tabular}
\end{table}
\begin{table}[htbp]
\centering
\caption{CoDEAL: comparison error under varying treatment start timestamps with $N_1=100$ never-treated units.}
\label{tab:disjoint_tstart}
\begin{tabular}{ccc}
\toprule
Treatment start time & MAE & MSE \\
\midrule
15 & $0.953$ & $1.869$ \\
20 & $0.887$ & $1.584$ \\
25 & $0.846$ & $1.436$ \\
30 & $0.809$ & $1.310$ \\
35 & $0.780$ & $1.208$ \\
\bottomrule
\end{tabular}
\end{table}

\subsection{Hyperparameter Setting and Sensitivity Analysis}\label{appen: hyper}
The simulation experiments in Section \ref{sec:simulation} and Appendix \ref{append:simu-covariate-removal} were repeated 50 times for each setup. %
For DNN-based covariate effect removal, we used a single hidden‐layer neural network of width 32, trained with the Adam optimizer (learning rate $\eta=0.001$), Xavier initialization, for 400 epochs. Both single‐output AE and multi‐output AE for matrix completion employed three hidden layers in their encoders and decoders; these were trained with Adam (learning rate $\eta=0.002$) for 50 epochs. The AEMC-NE model featured layer‐wise hidden sizes $[30,30]$ and element‐wise sizes $[20,20]$, was trained with AdamW (learning rate $\eta=0.001$) for 30 epochs. 

We also conduct a sensitivity analysis for CoDEAL under Configuration 2 with sine factors, where the true latent factor dimension is $K=3$. We vary the learning rate $\eta$, the number of DNN hidden layers $L$, and the AE bottleneck size $K_1$. The results are reported in Table~\ref{tab:sensitivity_analysis}.

\begin{table}[htbp]
\centering
\caption{Sensitivity analysis of CoDEAL under Configuration 2 with sine factors $(K=3)$.}
\label{tab:sensitivity_analysis}
\small
\setlength{\tabcolsep}{5pt}
\begin{tabular}{c|cccccc}
\toprule
Learning rate $\eta$ & 0.0005 & 0.001 & 0.002 & 0.003 & 0.004 & 0.005 \\
\midrule
MSE                 & 1.624  & 1.284 & 1.173 & 1.078 & 1.098 & 1.122 \\
\midrule
Hidden layers $L$   & 1      & 2     & 3     & 4     & 5     &       \\
\midrule
MSE                 & 1.268  & 1.189 & 1.173 & 1.269 & 1.413 &       \\
\midrule
Bottleneck $K_1$    & 2      & 3     & 4     & 5     & 6     &       \\
\midrule
MSE                 & 1.464  & 1.173 & 1.108 & 1.063 & 1.023 &       \\
\bottomrule
\end{tabular}
\end{table}

There are three main findings from Table \ref{tab:sensitivity_analysis}. First, CoDEAL exhibits stable performance across a broad range of learning rates, suggesting that its accuracy is relatively insensitive to this choice, provided the learning rate is not excessively small. Second, moderate network depth generally yields better performance, indicating that additional layers do not necessarily improve accuracy. Third, and most consequential in practice, the bottleneck dimension $K_1$ is the one choice that materially affects the performance. In particular, under-specifying the bottleneck dimension leads to substantial accuracy deterioration, whereas over-specification does not adversely affect performance and may even yield modest improvements. This suggests that $K_1$ should be chosen to be at least as large as the anticipated number of latent factors. Overall, these findings demonstrate that CoDEAL is relatively robust to the choice of learning rate and network depth, with the bottleneck dimension being the most important hyperparameter to consider in practice.

Beyond the sensitivity analysis presented above, hyperparameters can also be selected in a data-driven manner by minimizing the out-of-sample reconstruction error on held-out blocks of the observed untreated panel. We further discuss this approach in Remark~\ref{rmk:masked-block} as a stand-alone empirical evaluation strategy, particularly useful for hyperparameter tuning and assessing performance in real-world applications where ground-truth counterfactual outcomes are inherently unobservable.

\begin{rmk}[An empirical strategy for out-of-sample evaluation and hyperparameter selection]\label{rmk:masked-block}
    One may construct pseudo-treatment blocks within the observed untreated portion of the panel, mask these blocks, fit competing methods, and evaluate their reconstruction errors on the held-out observations. Since the true untreated outcomes are observed for these masked entries, the resulting prediction error provides an out-of-sample measure of counterfactual reconstruction accuracy. This masked-block evaluation offers a practical approach to assessing counterfactual reconstruction performance in empirical applications, where the true counterfactual outcomes are inherently unobservable. The same strategy can also be used for data-driven hyperparameter selection.
\end{rmk}

\subsection{Compute Resources and Computation Time}\label{appen:Experimental Setting}

All simulations were executed on a 16‐core CPU. Over 50 replications, Configurations 1–2 (four‐block design) took roughly 1 hour and 500 MB max memory for every combination of factor and covariate generation, while Configurations 3–4 (staggered‐adoption design) required about 3 hours running time and max memory used 1GB for each factor generation type.

We report the per-replication runtime comparisons in Table \ref{tab:runtime_compare1}. The results show that although the inclusion of separate decoders in CoDEAL increases computational cost, the overall overhead remains modest, even when N is large.

\begin{table}[htbp]
\centering
\small
\caption{Runtime (in seconds) comparison in Config. 1 and 2 across methods with linear/sine factors and nonlinear (tanh) covariates}
\label{tab:runtime_compare1}
\begin{tabular}{lcc|cc}
\hline
\footnotesize
\multirow{2}{*}{}& \multicolumn{2}{c|}{Config. 1} & \multicolumn{2}{c}{Config. 2} \\
 & Linear&  Sine& Linear&  Sine\\
\hline
CoDEAL& 1.295& 1.286& 2.429&  2.495\\
Single AE& 0.025& 0.026& 0.025&  0.027\\
AEMC-NE& 0.115& 0.118& 0.125&  0.129\\
DiD        & 0.000& 0.001& 0.002&  0.001\\
Vert-Reg& 0.034& 0.037& 0.083&  0.083\\
MC-NNM& 1.421& 1.408& 3.876&  3.820\\
KernelPCA& 0.006&  0.004& 0.004& 0.003\\
\hline
\end{tabular}
\end{table}

\section{Supplement to Real Data Analysis in Section \ref{sec:realdata}}\label{append: real}
\subsection{Additional Results for OxCGRT dataset}
\label{append: covid19}
\begin{figure}[H]
\centering
\includegraphics[width=0.85\textwidth]{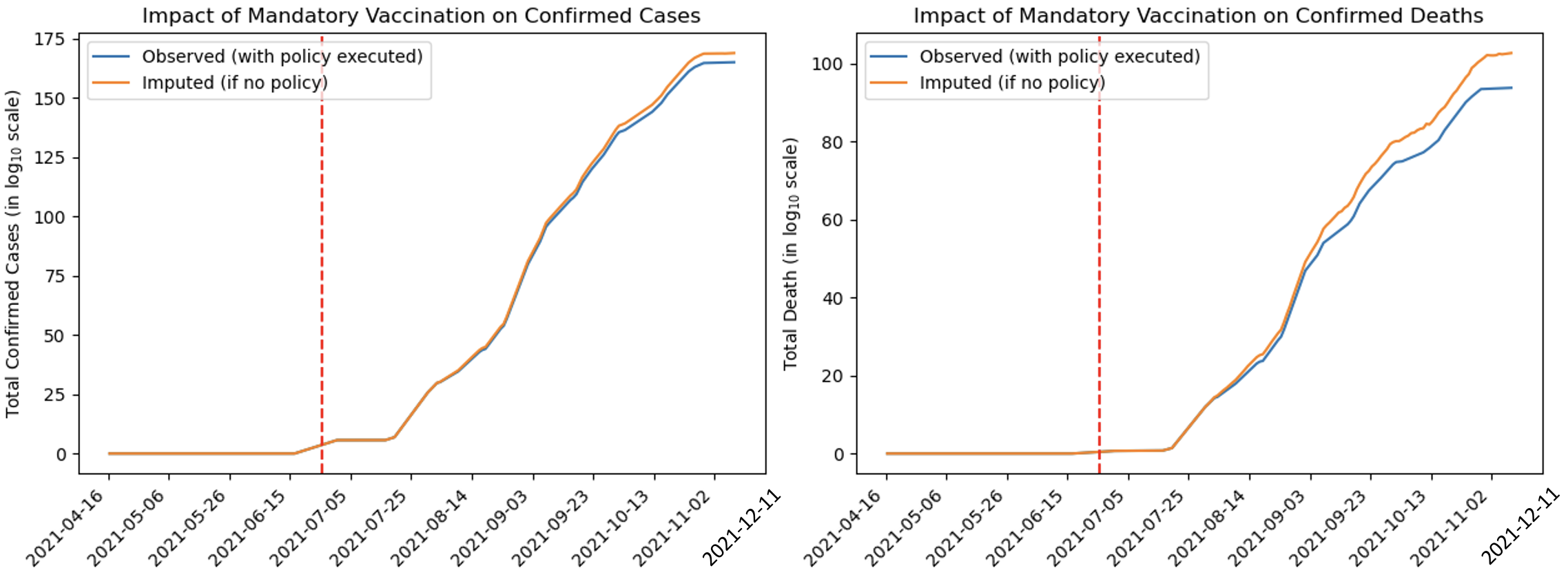}
    \caption{\small Comparison of the total confirmed cases (Left) and deaths (Right) across policy-in-effect states between observed results with the mandatory vaccination policy executed and estimated counterfactuals if no policy.
    Values are reported on a ${\log_{10}}$ scale, and plots are visualized with 14-day moving averages.} 
    \label{fig:large_MV}
\end{figure}

\begin{figure}[p]
\centering
    \includegraphics[width=0.75\textwidth]{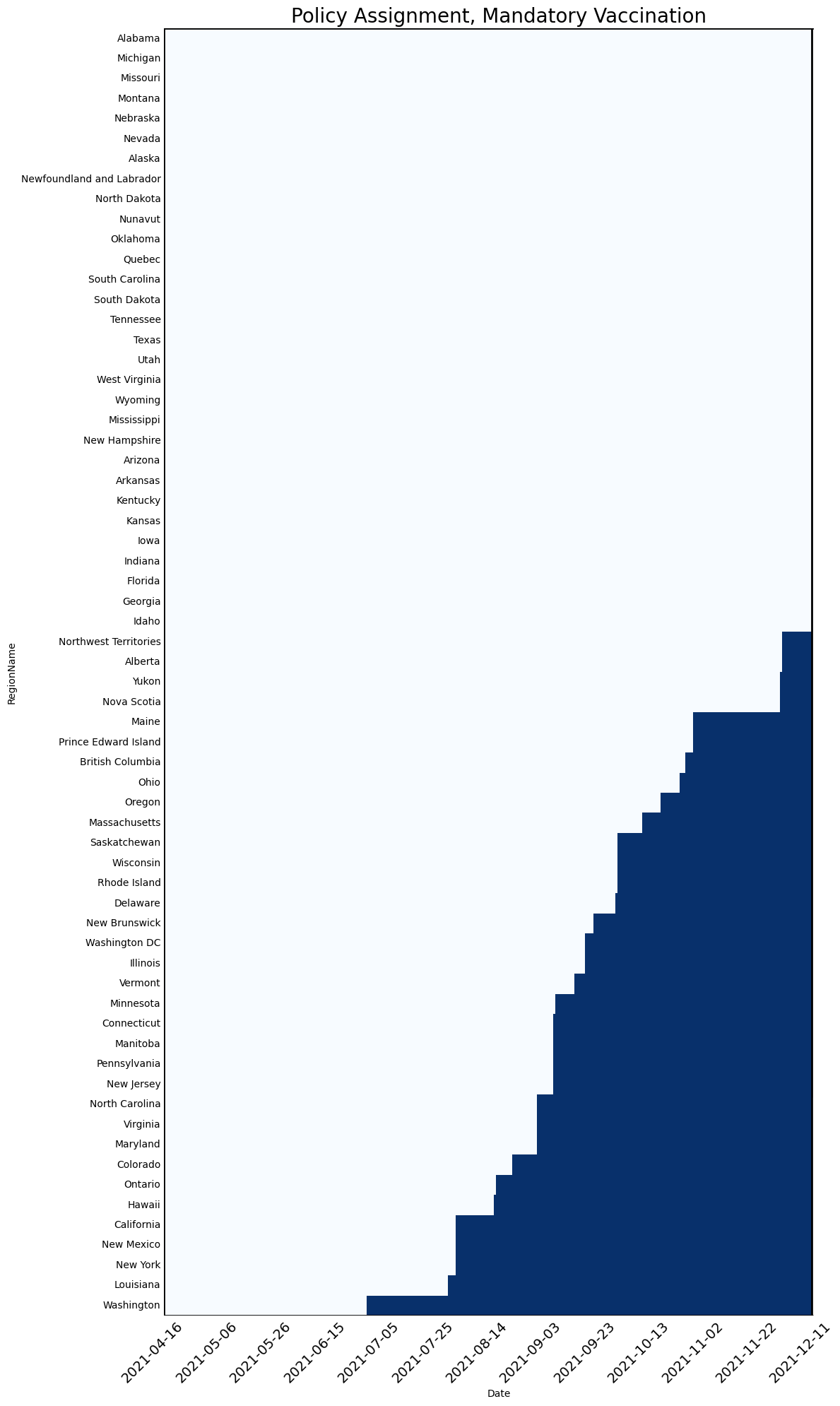}
    \captionsetup{justification=centering}
    \caption{Indicator matrix of the implementation of the mandatory vaccination policy. \\ Blocks with darker color refer to policy executed.} 
    \label{fig:MV_mask}
\end{figure}

This section reports supplementary results of the OxCGRT analysis. Figure \ref{fig:large_MV} and Figure \ref{fig:MV_mask} are enlarged plots of Figure \ref{fig:MV1} of the main text for clarity, showing the estimated impact of the mandatory vaccination policy and an indicator matrix of the observed policy assignment. 

\begin{figure}[H]
\centering
\includegraphics[width=0.85\textwidth]{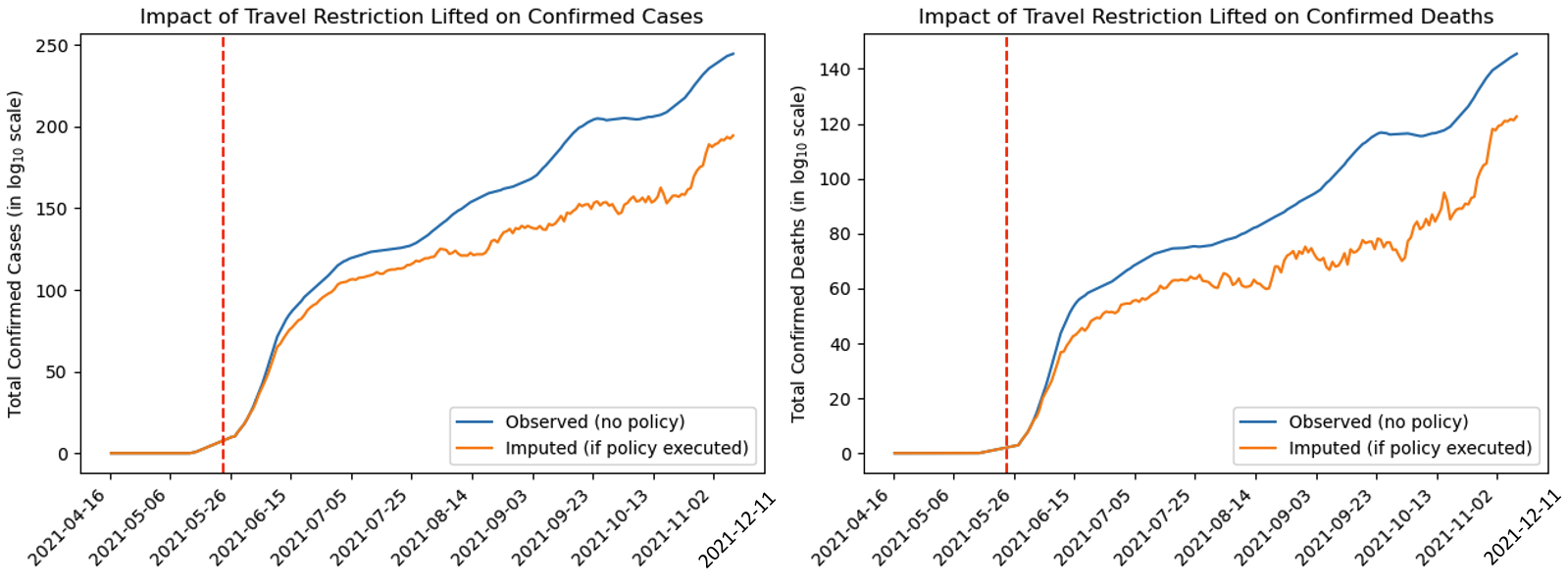}
    \caption{\small Comparison of the total confirmed cases (Left) and deaths (Right) across restriction-lifted states between observed results with the travel restrictions lifted and estimated counterfactuals if not.
    Values are reported in ${\log_{10}}$ scale, and plots are visualized with 14-day moving averages.}
    \label{fig:TR}
\end{figure}

\begin{figure}[p]
\centering
\includegraphics[width=0.75\textwidth]{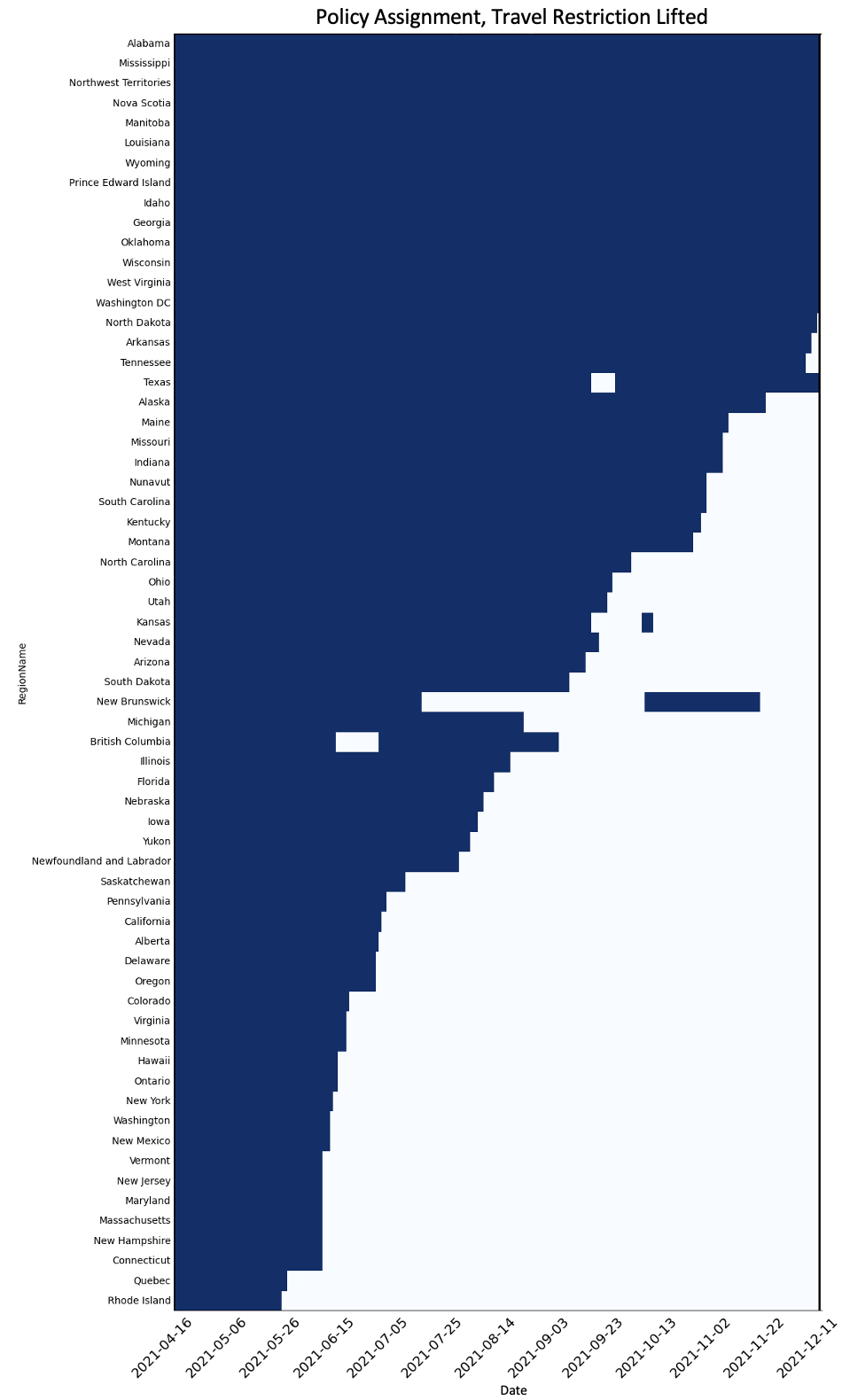}
    \captionsetup{justification=centering}
    \caption{Indicator matrix of the termination of interval travel restrictions. \\ Blocks with darker color refer to travel restriction lifted.} 
    \label{fig:TR_mask}
\end{figure}

We also examine the impact of the termination of internal travel restrictions, with an indicator matrix of the observed restriction-lifting assignments shown in Figure \ref{fig:TR_mask}. %
Figure \ref{fig:TR} plots the comparisons of the total confirmed cases and deaths across policy-lifted states between observed results (with travel restrictions lifted) and estimated counterfactuals (had the restrictions not been lifted). Results in Figure \ref{fig:TR} show that imposing internal travel restrictions led to similar proportional declines in both COVID-19 cases and deaths. This pattern is supported by recent studies \citep{chinazzi2020effect,murano2021impact}: by limiting movement, restrictions reduce virus transmission (and thus new cases), which in turn lowers the number of deaths, even though restricting trips does not directly affect individual mortality risk.

In Table \ref{tab:runtime_compare2}, we also report the comparisons of the estimated counterfactual confirmed cases and deaths (in millions) for other benchmark methods besides CoDEAL. The results are similar in terms of the magnitude, but differ in the exact numbers. The vertical regression estimates for deaths are not reported because the staggered adoption design yields very short pre-treatment windows and a small size of control units, which makes the regression unstable in this case. This leads to highly sensitive coefficient estimates and severe overextrapolation, producing implausibly large counterfactual results.

However, because of the fundamental problem of causal inference, we are unable to observe the ground-truth counterfactuals in real-world datasets. Therefore, while we can report and compare the predicted values by each method, it is not possible to objectively determine which method performs best in terms of prediction accuracy in this dataset.

\begin{table}[H]
\centering
\caption{Comparisons of the estimated counterfactual confirmed cases and deaths (in millions) on the last day, if no mandatory vaccination policy were implemented}
\label{tab:runtime_compare2}
\small
\begin{tabular}{lcc}
\hline
Method & Confirmed Cases&  Deaths\\
\hline
Ground Truth&26.6 &0.39 \\
CoDEAL&29.8 &0.52 \\
Single AE&34.1 &0.61 \\
AEMC-NE&25.3 &0.35 \\
DiD &31.5 &0.45 \\
Vert-Reg&28.3 &-- \\
MC-NNM&27.5 &0.41 \\
KernelPCA&15.9 &0.33  \\
\hline
\end{tabular}

\end{table}

\subsection{Additional Real-World Case Studies}
\label{append:additional-real-data}

We further evaluate CoDEAL on two additional real-world datasets that are commonly used in the causal panel data literature. Across both examples, CoDEAL produces estimates that are broadly consistent with existing findings.%

\paragraph{US Election Day Registration (EDR).}
We use the dataset in \citet{xu2017generalized} to estimate the effect of Election Day Registration (EDR) laws on voter turnout. CoDEAL estimates that EDR increases turnout by \(6.84\%\). For comparison, the estimated effects are \(+1.70\%\) for Single AE, \(+9.23\%\) for AEMC-NE, \(+3.60\%\) for MC-NNM, \(+1.26\%\) for DiD, \(+4.30\%\) for Vert-Reg, and \(+12.39\%\) for KernelPCA. Overall, CoDEAL yields a positive and moderate estimate, lying within the range of competing methods and supporting the conclusion that EDR has a favorable effect on voter participation.

\paragraph{German Reunification and GDP.}
We use the dataset in \citet{pang2022bayesian} to estimate the effect of German reunification on log-GDP per capita in Western Germany. CoDEAL estimates a decrease of \(20.14\%\). The corresponding estimates from the benchmarks are \(-21.71\%\) for Single AE, \(-24.81\%\) for AEMC-NE, \(-25.94\%\) for MC-NNM, \(-19.34\%\) for DiD, \(-21.86\%\) for Vert-Reg, and \(-22.12\%\) for KernelPCA. In this example, CoDEAL again produces an estimate aligned with the broader literature and close to the competing methods, suggesting a substantial negative impact of reunification on Western Germany’s log-GDP per capita in the post-treatment period.

These two additional case studies provide further evidence that CoDEAL yields empirically plausible estimates on real panel datasets and remains stable across applications with different treatment mechanisms and substantive interpretations.

\section{Further Discussions}
\label{append:discussion}

\subsection{Discussions on Broader Impacts}\label{append:broader-impacts}

In this paper, we propose the CoDEAL framework and demonstrate its effectiveness for estimating heterogeneous treatment effects in causal panel data models. Though we presented our methods within the context of panel data, the proposed CoDEAL does not have to be confined to panel data but can accommodate a variety of data types. The main idea of CoDEAL is to integrate nonlinear covariate effect components with nonlinear factor structures to effectively capture the inherent data dependencies and model the data heterogeneity. In panel data models (in the matrix form) with unit-specific covariates (in the vector form), we employ a feed-forward deep neural network to model the covariate effects and a multi-output autoencoder to characterize the nonlinear factor model. Methodologically, with different choices of factor models and advanced neural network architectures, the proposed CoDEAL framework can be further generalized to accommodate various data types with more complex data structures, such as tensors, images, and networks. The broad potential for generalization of CoDEAL to diverse data types brings a broader impact across both research and practice, unlocking innovations well beyond traditional panel data settings. Such generalization can offer new insights and opportunities to the broader research community beyond the field of panel data modeling. In addition, the generalization to more complex data structures can also equip practitioners with a powerful new toolkit for tackling an expanded array of real‐world problems.

\subsection{Discussions on Further Extensions}\label{append:extension}

The proposed framework largely extends existing causal panel literature in the capacity of accommodating heterogeneity and nonlinearity. 
We also clarify that several assumptions (e.g., time-invariant covariates, static treatment effects) are not intrinsic to the algorithm and can be relaxed to accommodate more general settings. Moreover, despite the additive model formulation, the proposed framework does not preclude interactions between covariates and latent factors, nor does it require them to be uncorrelated. 
The CoDEAL framework is amenable to further adaptation across data-driven contexts.

\subsubsection{Dynamic Treatment Effects}

As discussed in Remark~\ref{rmk:time-varying-treatment}, the static treatment assumption is not intrinsic to the proposed algorithm, but can be extended to accommodate time-varying contemporaneous treatment effects.
In principle, CoDEAL can produce estimates of unit-time-specific treatment effect $\tau_{it}$ using $\widehat{\tau}_{it} = Y_{it}(1) - \widehat{Y}_{it} (0)$. 
  The static-effect assumption enters only the interpretation of the averaged estimator $\widehat{\bar\tau}_i := \ T_2^{-1}\sum_{t:W_{it}=1}\widehat{\tau}_{it}$, not the imputation machinery. CoDEAL's counterfactual imputation $\widehat{Y}_{it}(0)$ is trained exclusively on untreated outcomes, so Theorem~\ref{thm1}'s guarantee on 
  $
  \max_{N_1+1\le i\le N}\ \frac{1}{T_2}\sum_{t=T_1+1}^{T}
  \mathbb{E} (\widehat{Y}_{it}(0)-\phi_i^\star(\mathbf{F}_t^\star)-g_t^\star(\mathbf{X}_i) )^2 
  $
  is entirely unaffected by whether the treatment effect is static or dynamic.
  When the treatment effect is static ($\tau_{it} \equiv \tau_i$), 
  this result establishes the estimation consistency of $\widehat{\bar\tau}_i$ for $\tau_i$. 
  When the treatment effect $\tau_{it}$ varies over time, Theorem \ref{thm1} implies that $\widehat{\bar\tau}_i$ consistently estimates the unit-wise average treatment effect  $\bar{\tau}_i=T_2^{-1}\sum_{t:W_{it}=1}\tau_{it}$. 
  Thus, CoDEAL permits the contemporaneous treatment effect to vary across time and units, while preserving consistency for the average effect.

  A pointwise guarantee of $\widehat{\tau}_{it}$ for ${\tau}_{it}$, however, is not attainable by the current theory.
  Estimating dynamic effect trajectories with pointwise guarantees would require additional assumptions and substantially more sophisticated techniques, which we leave for future research. We also clarify that the current theoretical framework for CoDEAL does not cover general lagged or fading effects. Modeling such dynamic treatment responses would require potential outcomes indexed by event time or treatment history, together with additional identifying assumptions and theoretical analysis, and we leave these important extensions for future work.

\subsubsection{Time-Covariate Interaction and Time-Varying Covariates}

While we present the methodology with unit-specific time-invariant covariates $\bX_i$, $i \in [N]$, the time-covariate interaction is not absent from the modeling. The covariate effect is allowed to vary over time through {\small $g^\star_{t}(\bX_i)$}. The framework also extends straightforwardly to time-varying covariates by replacing {\small $g^\star_{t}(\bX_i)$} with  {\small $g^\star_{t}(\bX_{it})$}.

\subsubsection{Dependence between the Observed Covariates and Latent Factors} 
\label{append:discussion-covariate-factor-interaction}

Although Model \eqref{eq:CoDEAL-Model} admits an additive formulation in the covariate effect and the latent factor effect, the proposed framework does not preclude the dependence between the observed covariates and the unobserved unit-level heterogeneity.
Formally, if the truth is $Y_{it}=\psi(\bX_i,\bU_i,\bF_t^\star)+g_t^\star(\bX_i)+\tau_i^\star\mathbf 1\{W_{it}=1\} +\varepsilon_{it}$ for arbitrary unobserved heterogeneity $\bU_i$ (precisely the case where heterogeneity depends on covariates) then setting $\phi_i^\star(\cdot):=\psi(\bX_i,\bU_i,\cdot)$ recovers Model \eqref{eq:CoDEAL-Model} exactly. The theory applies verbatim. The multi-output decoder is exactly the estimator of this structure, as a single-output AE (i.e., SingleAE) cannot absorb $X$-dependent loadings. %
We also conduct experiments in which $\bX_i$ and $\bF_t$ are dependent, with details presented in Appendix \ref{append:simu-xf}. The results show that SingleAE shows deteriorating performance as dependence gets stronger, while CoDEAL is robust to dependence and heterogeneity. 

What additivity does restrict is when interactions may vary over time only (i) through the fully unrestricted $g_t(\bX_i)$ (arbitrary covariate-time interaction), or (ii) through the $K$-dimensional $\bF_t^\star$ via $\phi_i$. Excluded are components $\Psi(\bX_i,\bU_i,t)$ full-rank in $t$ outside the factor span, essentially the premise of any factor-based method, not an additional cost of additivity. Moreover, the $g$/$\phi$ split is identified only up to the reallocation discussed in Appendix \ref{append:seq-vs-joint}, and both $\widehat Y_{it}(0)$ and Theorem \ref{thm1} depend on the sum alone; misallocated interaction affects component interpretation but not the ATT.

\subsubsection{Potential Extensions to Multiple Treatment Settings} \label{append:limitation}
One limitation of CoDEAL is that it is only designed for binary treatment settings. Causal panel data with multiple treatments is an open research area with limited existing work in the literature \citep{abadie2021using}, yet this topic is gradually gaining attention due to its growing relevance in practical applications \citep{agarwal2020synthetic,squires2022causal}.
To extend CoDEAL to multiple treatment settings, one possible direction is to adopt a similar strategy to \citep{agarwal2020synthetic} 
in which the treatment type is incorporated as an additional data dimension. %
Together with the original panel data structures, the inclusion of treatment dimension forms a three-dimensional tensor. 

\section{Supplementary Theoretical Properties and Proofs of Main Theorem} \label{section:proofs}

\subsection{Preliminary}

The set of natural numbers is denoted by $\mathbb N$, and we define $\mathbb N_0 = \mathbb N \cup \{0\}.$ For any function $f:\R^d \to \R$, we define the supremum norm as $\|f\|_{\infty}=\sup_{\bx\in\R^d} |f(\bx)|$. For a matrix $A=(a_{ij})\in\R^{m\times n}$, we define the entrywise max norm as $\|A\|_{\infty}=\max_{i,j}|a_{ij}|$. Let $\P_{\bX}$ denote the distribution of the random vector $X$. Then, the $\mathcal L_2(\P_{\bX})$ norm of a function $f$ is defined as its $\mathcal L_2 $ norm with respect to the probability measure $\P_{\bX}$, i.e.,
\[
\|f\|_{\mathcal{L}_2(\mathbb{P}_{\bX})} := \left( \int |f(\by)|^2 \, d\mathbb{P}_{\bX}(\by) \right)^{1/2}.
\]
For two sequences of real numbers $\{a_n\}$ and $\{b_n\}$, write $a_n=O(b_n)$ (resp. $a_n\asymp b_n$) if there exists a constant $C$ such that $|a_n|\leq C |b_n|$ (resp. $1/C \leq a_n/b_n\leq C$) for all sufficiently large $n$, and write $a_n=o(b_n)$ if $\lim_{n\to\infty} a_n/b_n =0$. Write $a_n\lesssim b_n$ (resp. $a_n\gtrsim b_n$) if there exists a constant $C$ such that $a_n\le Cb_n$ (resp. $a_n\ge Cb_n$). Write $a\wedge b=\min\{a,b\}$ and $a\vee b=\max\{a,b\}$. We use $C, C_1,c,c_1,...$ to denote generic constants, whose actual values may vary from line to line. 

\begin{definition}
Let $\beta = q + s$ for some $q \in \mathbb{N}_0$ and $0 < s \leq 1$. A function $m : \mathbb{R}^d \rightarrow \mathbb{R}$ is called \((\beta, C)\)-H\"older smooth, if for every $\balpha = (\alpha_1, \dots, \alpha_d) \in \mathbb{N}_0^d$ with $\sum_{j=1}^d \alpha_j = q$, the partial derivative
\[
\frac{\partial^q m}{\partial x_1^{\alpha_1} \cdots \partial x_d^{\alpha_d}}
\]
exists and satisfies
\[
\left| \frac{\partial^q m}{\partial x_1^{\alpha_1} \cdots \partial x_d^{\alpha_d}}(\bx)
- \frac{\partial^q m}{\partial x_1^{\alpha_1} \cdots \partial x_d^{\alpha_d}}(\bz) \right|
\leq C \cdot \|\bx - \bz\|^s
\]
for all $x, z \in \mathbb{R}^d$, where $\|\cdot\|$ denotes the Euclidean norm.
    
\end{definition}

\subsection{Covariate Adjustment}
We begin by analyzing the estimation accuracy of the covariate function $g_t^\star$. For technical convenience, we assume that the covariate function $g_t^\star$ in model \eqref{eq:CoDEAL-Model} admits the following hierarchical decomposition
\begin{align}\label{eq:hierarchical}
g_t^\star(\bX_i)= \sum_{j=1}^P h_{0t,j} (X_{i,j})+ \sum_{j_1<j_2}  h_{0t,j_1 j_2} (X_{i,j_1},X_{i,j_2}) + \cdots +\sum_{j_1<\cdots <j_k}  h_{0t,j_1 j_2...j_k} (X_{i,j_1},...,X_{i,j_k})  ,
\end{align}
where $k\le P$, each $h_{0t,j}$ is a univariate function and each $h_{0t,j_1 \cdots j_k}$ is a $k$-variate interaction function.
Under this decomposition, the smoothness condition in Assumption~\ref{asmp:smooth} for $g_t^\star$ is equivalent to the same smoothness condition imposed on all component functions $h_{0t,j_1 \cdots j_k}$. To ensure identifiability and enable estimation of these nonparametric components, we impose additional structural constraints to avoid non-identifiable additive shifts. To that end, we make the following assumption.

\begin{assumption}[Identifiability and boundedness]\label{asmp:ident}
We assume the following conditions for identifiability purposes and boundedness:
\begin{enumerate}
\item All the $k$-variate interaction functions in \eqref{eq:hierarchical} have 0 marginals, that is,
    \[
    \int_0^1 h_{0t,j_1 \cdots j_k}(x_1,...,x_k) \, dx_1 =\cdots = \int_0^1 h_{0t,j_1 \cdots j_k}(x_1,...,x_k) \, dx_k = 0.
    \]
    
\item $\|g_t^\star\|_{\infty} \leq B$ for some $B > 0$.
\end{enumerate}
   
\end{assumption}
Note that the above assumption implies that the total integral of $h_{0t,j_1 \cdots j_k}$ is also 0. Assumption~\ref{asmp:ident} ensures identifiability of all the functions involved in \eqref{eq:hierarchical}.

We now briefly describe our estimation procedure for covariate adjustment. Consider the model in equation \eqref{eq:hierarchical}. For each time point $t$, we estimate the component functions using neural networks \((\psi_{t1}, ...,\psi_{tk})\) by minimizing the squared-error loss:
\begin{equation}\label{eq:x_erm}
(\hat{\psi}_{t1}, ..., \hat{\psi}_{tk}) = \arg\min_{\psi_{t1} \in \mathcal{F}^{1}_{NN}, ...,\psi_{tk} \in \mathcal{F}^{k}_{NN}} \frac{1}{N} \sum_{i=1}^N (1-W_{it}) (Y_{it} - \psi_{t1}(\bX_{i}) - \cdots - \psi_{tk}(\bX_i))^2 .
\end{equation}
We then define the aggregated estimate as \(\hat{g}_{t,\text{raw}} := \hat{\psi}_{t1} +\cdots+ \hat{\psi}_{tk}\), and obtain the final estimator $\widehat{g}_t$ by truncating $\hat{g}_{t,\text{raw}}$ at level $B$: 
\begin{align*}
\widehat{g}_t := \text{sgn}(\hat{g}_{t,\text{raw}})(|\hat{g}_{t,\text{raw}}| \wedge B)    
\end{align*}
where $B$ is chosen such that \(\|g_t^\star\|_\infty \leq B\), as specified in point 2 of Assumption~\ref{asmp:ident}. The function classes \(\mathcal{F}^{k}_{NN}, 1\le k\le p,\) are defined as
\begin{equation}
\mathcal{F}^{k}_{NN} = \mathcal{F}_{NN}(p, c_1 p^k M_k \log M_k, c_2 L_k \log L_k, c_3 p^k (M_k \log M_k)^2 L_k \log L_k, 1)
\end{equation}
for some constants \(c_1, c_2, c_3\). 
These are not fully connected networks over the entire covariate vector; rather, we use fully connected neural networks to estimate each $k$-variate component function separately, with width \(M_k \log M_k\) and depth \(L_k \log L_k\), and then sum the outputs. With slight abuse of notation, we continue to denote these specialized architectures by \(\mathcal{F}^{k}_{NN}\). The parameters \(M_k\) and \(L_k\) typically depend on the sample size $N$ and will be specified later.
As with any nonparametric estimator, the \textit{generalization error} of $\widehat{g}_t$ can be decomposed into two components:
\begin{equation}
\|\widehat{g}_t - g_t^\star \|_{\mathcal L_2(\P_{\bX})} \leq \underbrace{\|\widehat{g}_t - g_{0t}\|_{\mathcal L_2(\P_{\bX})}  }_{\text{statistical error}} +   \underbrace{\| g_{0t} - g_t^\star\|_{\mathcal L_2(\P_{\bX})}  }_{\text{approximation error}},
\end{equation}
where \( g_{0t}\) denotes the best approximation to \(g_t^\star\) in \(\mathcal L_2(\P_{\bX})\) norm within the specified class of neural networks. The following lemma presents a bound on the
overall generalization error, combining both statistical and approximation errors.

\begin{lemma}\label{lem:cov}
Under Assumptions \ref{asmp:density}, \ref{asmp:smooth}, \ref{asmp:ident}, the estimator $\widehat{g}_t$ of $g_t^\star$ defined in equation \eqref{eq:x_erm} satisfies
\[
\mathbb{E}[(\widehat{g}_t(\bX_i) - g_t^\star(\bX_i))^2 \mid \bX] = O_{\P} \left( P^{k} N^{-\frac{2\beta}{2\beta + k}}  \log^4 N \right).
\]
\end{lemma}

The proof of this lemma follows the same argument as Theorem 2.11 in \cite{bhattacharya2024deep} and is therefore omitted. A careful examination of that proof also yields the following proposition in a four-block design. 
\begin{proposition}\label{prop:cov}
Under Assumptions \ref{asmp:density}, \ref{asmp:smooth}, \ref{asmp:ident}, the estimator $\widehat{g}_t$ of $g_t^\star$ defined in equation \eqref{eq:x_erm} satisfies
\[
\max_{N_1+1\le i\le N }\frac{1}{T_2}\sum_{t=T_1+1}^{T_1+T_2} \mathbb{E}[(\widehat{g}_t(\bX_i) - g_t^\star(\bX_i))^2 \mid \bX] = O_{\P} \left( P^{k} N^{-\frac{2\beta}{2\beta + k}}  \log^5 N \right).
\]
\end{proposition}
When $P$ is fixed and a fully connected neural network is used to estimate $g_t^\star$, this is equivalent to setting $k=P$ in Lemma \ref{lem:cov} and Proposition \ref{prop:cov} above.

\subsection{Proof of Theorem \ref{thm1}}

Consider the covariate-adjusted observed outcomes as 
\begin{align*}
\widetilde{Y}_{it}(0) = Y_{it}(0) - \widehat{g}_t(\bX_i)=\phi_i^\star(\bF^\star_{t}) + g^\star_{t}(\bX_i) - \widehat{g}_t(\bX_i) +  \varepsilon_{it} =\phi_i^\star(\bF^\star_{t}) +  \varepsilon_{it} +o_{\P}(1).    
\end{align*}
In a four-block design, only within blocks $\cA,\cB, \cC$, we have untreated outcomes $\widetilde{\bY}(0)$. Let $\varepsilon_t=(\varepsilon_{1t},...,\varepsilon_{Nt})^\top$. Slightly abusing notation, we first consider the case without covariates, namely $\widetilde{Y}_{it}(0)=\phi_i^\star(\bF^\star_{t}) +\varepsilon_{it} $. Let $\widetilde{\bY}_{t}(0)=\bphi^\star(\bF^\star_{t}) +\varepsilon_{t} $, where $\bphi^\star=(\phi_1,...\phi_N)^\top$. We then can apply the triangle inequality to incorporate the additional estimation error arising from covariate adjustment.

Define the function class of the multi-output autoencoder is
\[
\mathcal{G}_{\text{AE}}^{K_1} := \left\{ (\phi_1, \ldots, \phi_N) \circ \rho : \rho \in \mathcal{G}^{K_1}_{N}(L_1, \bd_1, B, T^{5\beta + 5}), \, \phi_i \in \mathcal{G}^{1}_{K_1}(L_2, \bd_2, B, T^{5\beta + 5}), \, i \in [N] \right\},
\]
where $K_1\ge K$.
The proof of Theorem \ref{thm1} requires the following additional conditions
\begin{align}\label{thm1:eqa}
&K\le K_1 \le \min\{\max_{1\le \ell\le L_1} d_{1,\ell}, \max_{1\le \ell\le L_2} d_{2,\ell}\}   ,\, L_1\asymp L_2\asymp \log(T), \nonumber\\
& \max_{1\le \ell\le L_1} d_{1,\ell} \asymp  \max_{1\le \ell\le L_2} d_{2,\ell} \asymp T^{\frac{K}{2(2\beta+K)}},  \\
&\text{total number of weight parameters in the encoder is } O(n+T^{\frac{K}{2\beta+K}} \log T) . \nonumber
\end{align}
Let $d_1=\max_{1\le \ell\le L_1} d_{1,\ell} ,d_2=  \max_{1\le \ell\le L_2} d_{2,\ell}$.

\begin{proof}
Using the facts that
\[
\frac{1}{T_1} \sum_{t=1}^{T_1} \left( \phi_i^\star(\bF_t^\star) - \widehat{\phi}_i(\widehat{\rho}(\widetilde{\bY}_{t}(0))) \right)^2 \leq 4B^2, \exp(-cT) = o(T^{-1}), \exp(-cn) = o(n^{-1})
\]
and Proposition~\ref{prop1}, we have
\begin{align}\label{thm1:eq1}
&\frac{1}{T_1} \sum_{t=1}^{T_1} \mathbb{E} \left( \phi_i^\star(\bF_t^\star) - \widehat{\phi}_i(\widehat{\rho}(\widetilde{\bY}_{t}(0))) \right)^2 \nonumber\\
\lesssim& \left(n^{-1}K_1 + Nn^{-1}T^{-\frac{2\beta}{2\beta+K}} + Nn^{-2} \right)\log^{4}(T) + 4B^2 \cdot (\exp(-cT) + \exp(-cn)) \nonumber \\
\lesssim& \left(n^{-1}K_1 + Nn^{-1}T^{-\frac{2\beta}{2\beta+K}} + Nn^{-2} \right)\log^{4}(T) + 4B^2 \cdot o(T^{-1} + n^{-1}) \nonumber \\
\lesssim& \left(n^{-1}K_1 + Nn^{-1}T^{-\frac{2\beta}{2\beta+K}} + Nn^{-2} \right)\log^{4}(T)+ T^{-1}. 
\end{align}
Assuming that $\{(\widetilde\bF_t^\star, \widetilde{\varepsilon}_t)_{t=1}^T, \check{\bY}(0)\}$ is an independent copy of $\{(\bF_t^\star, \varepsilon_t)_{t=1}^T, \widetilde{\bY}(0)\}$, we have, for any $T_1+1 \le s \le T$,
\begin{align}\label{thm1:eq2}
&\left|  \mathbb{E} \left( \widehat{\phi}_i(\widehat{\rho}(\widetilde \bY_{s}(0))) - \phi_i^\star(\bF_{s}^\star) \right)^2 - \frac{1}{T_1} \sum_{t=1}^{T_1}  \mathbb{E} \left( \widehat{\phi}_i(\widehat{\rho}(\widetilde{\bY}_{ t}(0))) - \phi_i^\star(\bF_t^\star) \right)^2 \right| \nonumber \\
=& \left| \frac{1}{T_1} \sum_{t=1}^{T_1} \mathbb{E} \left( \widehat{\phi}_i(\widehat{\rho}(\check{\bY}_{ t}(0))) - \phi_i^\star(\widetilde\bF_t^\star) \right)^2 - \frac{1}{T_1} \sum_{t=1}^{T_1} \mathbb{E} \left( \widehat{\phi}_i(\widehat{\rho}(\widetilde{\bY}_{ t}(0))) - \phi_i^\star(\bF_t^\star) \right)^2 \right|. 
\end{align}
Observe that $\phi_i \circ \rho \in \mathcal{G}_{N}^1(L_1+L_2, \bd_1 +\bd_2, B, T^{5\beta + 5})$ and its total number of non-zero weights are asymptotically bounded by $n + T^{\frac{K}{2\beta + K}} \log T$. Therefore, by Lemma~\ref{lemm6}, there exists 
\[
\mathcal{G}_{N,\delta}^1(L_1+L_2, \bd_1 + \bd_2, B, T^{5\beta + 5})
\]
with $\delta = T^{-1}$ satisfying $\log |\mathcal{G}_{N,\delta}^1| \lesssim (n + T^{\frac{K}{2\beta + K}}) \log^4 (N T)$.
In addition, for any $\phi_i \circ \rho$ satisfying the conditions in \eqref{thm1:eqa} and Theorem \ref{thm1}, there exists a function $f(\cdot) \in \mathcal{G}_{N,\delta}^1$ such that $\| \phi_i \circ \rho(x) - f(x) \| \leq T^{-1}$ whenever $\| x \|_\infty \leq 2N T$. Without loss of generality, we can assume all the functions in $\mathcal{G}_{N,\delta}^1$ are bounded by $B$. Write $\mathcal{G}_{N,\delta}^1 = \{f_j, j \in [|\mathcal{G}_{N,\delta}^1|]\}$ and assume $f_j \in \mathcal{G}_{N,\delta}^1$ such that
\begin{equation}\label{thm1:eq3}
\| \widehat{\phi}_i \circ \widehat{\rho}(x) - f_{j^\star}(x) \|_\infty \leq \delta = T^{-1}, 
\end{equation}
for $\|x\|_\infty \leq 2N T$. Note that by Lemma~\ref{lemm2} and the fact that $\|\phi_i^\star(\bF_t^\star) \|_\infty\lesssim 1, \|\phi_i^\star(\widetilde\bF_t^\star) \|_\infty \lesssim 1$, it is easy to show that with probability at least $1 - C \exp(-cT)$, $ \| \widetilde \bY_{\cA}(0)\|_\infty\leq 2N T, \| \widetilde \bY_{\cB}(0)\|_\infty\leq 2N T, \| \widetilde \bY_{\cC}(0)\|_\infty\leq 2N T,\|\check{\bY}(0)\|_\infty \leq 2N T $.
Therefore, under the same event, \eqref{thm1:eq3} holds for $\widetilde \bY_{1}(0),...,\widetilde \bY_{T}(0), \check{\bY}_{1}(0),...,\check{\bY}_{T}(0)$. Let us define
\[
r_{ij} := \max \left( \sqrt{T^{-1} \log |\mathcal{G}_{N,\delta}^1|}, \sqrt{ \mathbb{E} \left[ \left( f_j(\widetilde\bY_{s}(0)) - \phi_i^\star(\bF_{s}^\star) \right)^2 \right] } \right),
\]
\[
V_i := \max_{j \in [|\mathcal{G}_{N,\delta}^1|]} \left| \frac{1}{r_{ij} B} \sum_{t=1}^{T_1} \left[ (f_j(\check{\bY}_{t}(0)) - \phi_i^\star(\widetilde{\bF}_t^\star))^2 - (f_j(\widetilde\bY_{t}(0)) - \phi_i^\star(\bF_t^\star))^2 \right] \right|,
\]
for some $s> T$.
Then by \eqref{thm1:eq3}, we have
\begin{align}\label{thm1:eq4}
r_{ij^\star} &= \max \left( \sqrt{T^{-1} \log |\mathcal{G}_{N,\delta}^1|}, \mathbb{E}^{1/2} \left[ \left( f_{j^\star}(\widetilde\bY_{s}(0)) - \phi_i^\star(\bF_{s}^\star) \right)^2 \big| \widetilde\bY(0)  \right] \right) \nonumber \\
&\leq \sqrt{T^{-1} \log |\mathcal{G}_{N,\delta}^1|} + \mathbb{E}^{1/2} \left[ \left( \widehat{\phi}_i(\widehat{\rho}(\widetilde\bY_{s}(0))) - \phi_i^\star(\bF_{s}^\star) \right)^2 \big| \widetilde\bY(0) \right] + T^{-1}.
\end{align}
Moreover, by \eqref{thm1:eq3}, with probability at least $1 - C \exp(-cT)$,
\begin{align*}
&\left| \frac{1}{T_1} \sum_{t=1}^{T_1}  \left( \widehat{\phi}_i(\widehat{\rho}(\check{\bY}_{t}(0))) - \phi_i^\star(\widetilde\bF_t^\star) \right)^2 - \frac{1}{T_1} \sum_{t=1}^{T_1}  \left( \widehat{\phi}_i(\widehat{\rho}(\widetilde{\bY}_{t}(0))) - \phi_i^\star(\bF_t^\star) \right)^2 \right| \\
\leq& \left|  \frac{1}{T_1} \sum_{t=1}^{T_1} \left( f_{j^\star}(\check{\bY}_{t}(0)) - \phi_i^\star(\widetilde{\bF}_t^\star) \right)^2 - \frac{1}{T_1} \sum_{t=1}^{T_1} \left( f_{j^\star}(\widetilde{\bY}_{t}(0)) - \phi_i^\star(\bF_t^\star) \right)^2  \right| + 9 T^{-1} B.
\end{align*}
Using the fact that $\exp(-cT) = o(T^{-1})$, it follows that
\begin{align}\label{thm1:eq5}
&\left| \frac{1}{T_1} \sum_{t=1}^{T_1} \mathbb{E} \left( \widehat{\phi}_i(\widehat{\rho}(\check{\bY}_{t}(0))) - \phi_i^\star(\widetilde\bF_t^\star) \right)^2 - \frac{1}{T_1} \sum_{t=1}^{T_1} \mathbb{E} \left( \widehat{\phi}_i(\widehat{\rho}(\widetilde{\bY}_{t}(0))) - \phi_i^\star(\bF_t^\star) \right)^2 \right| \nonumber \\
\leq& \frac{1}{T_1} \sum_{t=1}^{T_1} \mathbb{E} \left( f_{j^\star}(\check{\bY}_{t}(0)) - \phi_i^\star(\widetilde{\bF}_t^\star) \right)^2 - \frac{1}{T_1} \sum_{t=1}^{T_1} \mathbb{E} \left( f_{j^\star}(\widetilde{\bY}_{t}(0)) - \phi_i^\star(\bF_t^\star) \right)^2 + 10 T^{-1} B \nonumber \\
\leq& \frac{B}{T_1} \mathbb{E}(V_i r_{ij^\star}) + 10 T^{-1} B. 
\end{align}
By Cauchy-Schwarz inequality and \eqref{thm1:eq4}, we have
\begin{align}\label{thm1:eq6}
\mathbb{E}(V_i r_{ij^\star}) 
&\leq \left( \mathbb{E} \left( \mathbb{E} \left[ \left( \widehat{\phi}_i \circ \widehat{\rho}(\widetilde{\bY}_{s}(0)) - \phi_i^\star(\bF_{s}^\star) \right)^2 \big| \widetilde\bY(0) \right] \right)^2 \right)^{1/2} (\mathbb{E} V_i^2)^{1/2} \nonumber \\
&\quad + \sqrt{T^{-1} \log |\mathcal{G}_{N,\delta}^1|} \, \mathbb{E} V_i + T^{-1} \mathbb{E} V_i \nonumber \\
&= \left( \mathbb{E} \left( \widehat{\phi}_i \circ \widehat{\rho}(\widetilde{\bY}_{s}(0)) - \phi_i^\star(\bF_{s}^\star) \right)^2 \right)^{1/2} (\mathbb{E} V_i^2)^{1/2} + \sqrt{T^{-1} \log |\mathcal{G}_{N,\delta}^1|} \, \mathbb{E} V_i + T^{-1} \mathbb{E} V_i. 
\end{align}
Then we need to bound $V_i$. For any fixed $j \in [|\mathcal{G}_{N,\delta}^1|]$, define
\[
Z_{tj} := [f_j(\check{\bY}_{t}(0)) - \phi_i^\star(\widetilde{\bF}_t^\star)]^2 - [f_j(\widetilde\bY_{t}(0)) - \phi_i^\star(\bF_t^\star) ]^2 .
\]
Then it is straightforward to see that $\{Z_{tj}\}_{t=1}^T$ is a series of i.i.d. random variables satisfying $\mathbb{E} Z_{tj} = 0$, $|Z_{tj}| \leq 4B^2$, and
\[
\mathrm{Var}(Z_{tj}) = 2 \, \mathrm{Var} \left( \left( f_j(\check{\bY}_{t}(0)) - \phi_i^\star(\widetilde{\bF}_t^\star) \right)^2 \right) 
\leq 8B^2 r_{ij}^2.
\]
Applying Bernstein’s inequality, we will obtain
\[
\mathbb{P} \left( V_i \geq x \right) 
= \mathbb{P} \left( \max_{j \in [|\mathcal{G}_{N,\delta}^1|]} \left| \frac{1}{r_{ij} B} \sum_{t=1}^{T_1} Z_{tj} \right| \geq x \right)
\leq 2 |\mathcal{G}_{N,\delta}^1| \max_{j \in [|\mathcal{G}_{N,\delta}^1|]} \exp \left( -\frac{x^2}{8x/(3r_{ij}) + 16T} \right).
\]
Since $r_{ij} \geq \sqrt{T^{-1} \log |\mathcal{G}_{N,\delta}^1|}$, setting $x \geq 6 \sqrt{T \log |\mathcal{G}_{N,\delta}^1|}$,  it holds that
\[
\mathbb{P}(V_i \geq x) \leq 2 \exp \left( -3x \sqrt{\log |\mathcal{G}_{N,\delta}^1|} / (16 \sqrt{T}) \right) .
\]
Thus, for sufficiently large $T$,
\begin{align*}
\mathbb{E} V_i &= \int_0^\infty \mathbb{P}(V_i \geq x) \, dx 
\leq 6 \sqrt{T \log |\mathcal{G}_{N,\delta}^1|} + \int_{6 \sqrt{T \log |\mathcal{G}_{N,\delta}^1|} }^\infty 2 |\mathcal{G}_{N,\delta}^1| \exp \left( \frac{-3x \sqrt{\log |\mathcal{G}_{N,\delta}^1|}}{16 \sqrt{T}} \right) dx \nonumber \\
&\leq 6 \sqrt{T \log |\mathcal{G}_{N,\delta}^1|} + \frac{32}{3} \cdot \sqrt{ \frac{T}{\log |\mathcal{G}_{N,\delta}^1|}  } < 7 \sqrt{T \log |\mathcal{G}_{N,\delta}^1|}. 
\end{align*}
Similarly, we can show $\mathbb{E} V_i^2 \leq 40 T \log |\mathcal{G}_{N,\delta}^1|$. By \eqref{thm1:eq6}, since $\log |\mathcal{G}_{N,\delta}^1| \lesssim (T^{\frac{K }{2\beta + K}} + n) \log^4(N T)$, it follows that
\begin{align*}
&\frac{B}{T_1} \mathbb{E}(V_i r_{ij^\star}) \\
\lesssim& \frac{B}{T_1} \left( \sqrt{\mathbb{E} \left( \widehat{\phi}_i \circ \widehat{\rho}(\widetilde{\bY}_{s}(0)) - \phi_i^\star(\bF_{s}^\star) \right)^2  }   \sqrt{T \log |\mathcal{G}_{N,\delta}^1|} + \sqrt{T^{-1} \log |\mathcal{G}_{N,\delta}^1|} + \log |\mathcal{G}_{N,\delta}^1| \right)   \\
\lesssim& \left( T^{-\frac{\beta}{2\beta + K}} + T^{-1/2} n^{1/2} \right) \log^2(N T) \sqrt{\mathbb{E} \left( \widehat{\phi}_i \circ \widehat{\rho}(\widetilde{\bY}_{s}(0)) - \phi_i^\star(\bF_{s}^\star) \right)^2  } + \left( T^{-\frac{2\beta}{2\beta + K}} + T^{-1} n \right) \log^4(N T).
\end{align*}

Employing \eqref{thm1:eq2}, \eqref{thm1:eq5}, we have, for $s>T,i>N_1$,
\begin{align}
&\left|  \mathbb{E} \left( \widehat{\phi}_i(\widehat{\rho}(\widetilde \bY_{s}(0))) - \phi_i^\star(\bF_{s}^\star) \right)^2 - \frac{1}{T_1} \sum_{t=1}^{T_1}  \mathbb{E} \left( \widehat{\phi}_i(\widehat{\rho}(\widetilde{\bY}_{t}(0))) - \phi_i^\star(\bF_t^\star) \right)^2 \right| \nonumber \\
\lesssim&  \left( T^{-\frac{\beta}{2\beta + K}} + T^{-1/2} n^{1/2} \right) \log^2(N T) \sqrt{\mathbb{E} \left( \widehat{\phi}_i \circ \widehat{\rho}(\widetilde{\bY}_{\cB,s}(0)) - \phi_i^\star(\bF_{s}^\star) \right)^2  } + \left( T^{-\frac{2\beta}{2\beta + K}} + T^{-1} n \right) \log^4(N T).
\end{align}
By basic inequality and \eqref{thm1:eq1}, we have with probability at least $1-C\exp(-cT)$, for $s>T,i>N_1$,
\begin{align*}
&\left| \mathbb{E} \left( \widehat{\phi}_i(\widehat{\rho}(\widetilde \bY_{s}(0))) - \phi_i^\star(\bF_{s}^\star) \right)^2  \right| \nonumber \\
\lesssim&  \left( T^{-\frac{2\beta}{2\beta + K}} + T^{-1} n \right) \log^4(N T) + (n^{-1}K_1 + Nn^{-1}T^{-\frac{2\beta}{2\beta+K}} + Nn^{-2}) \log^4(T) +T^{-1}.
\end{align*}
As $\exp(-cT)=o(T^{-1})$ and $N_1\asymp N, T_1\asymp T$, we further conclude with probability at least $1-C\exp(-cT)$,
\begin{align}\label{thm1:eq7}
& \frac{1}{T_2} \sum_{s=T_1+1}^{T} \mathbb{E} \left( \widehat{\phi}_i(\widehat{\rho}(\widetilde \bY_{s}(0))) - \phi_i^\star(\bF_{s}^\star) \right)^2   \nonumber \\
\lesssim&  \left( (N+n)n^{-1}T^{-\frac{2\beta}{2\beta+K}} + T^{-1} n + n^{-1}K_1 +Nn^{-2}  \right) \log^4(N T) .
\end{align}

Therefore, by the triangle inequality, Proposition \ref{prop:cov} and \eqref{thm1:eq7}, we have
\begin{align*}
& \frac{1}{T_2} \sum_{t=T_1+1}^{T} \mathbb{E} \left( \widehat{\phi}_i(\widehat{\rho}(\widetilde \bY_{t}(0))) + \widehat{g}_t(\bX_i)  - \phi_i^\star(\bF_{t}^\star) - g_t^\star(\bX_i) \right)^2   \nonumber \\
=& O_{\P} \left(  \left( (N+n)n^{-1}T^{-\frac{2\beta}{2\beta+K}} + T^{-1} n + n^{-1}K_1 +Nn^{-2}  \right) \log^4(N T) +N^{-\frac{2\beta}{2\beta + k}}  \log^5 N \right).
\end{align*}

\end{proof}

\subsection{Technical Lemmas}

\begin{proposition}\label{prop1}
Under the condition of Theorem~\ref{thm1}, with probability at least $1-C\exp(-cT)-C\exp(-cn)$, for $N,T$ sufficiently large, we have
\begin{align*}
&\frac{1}{NT}\sum_{(i,t)\in\cA,\cB,\cC}\bigl(\phi_i^{\star}(\bF_t^{\star})-\widehat{\phi}_i(\widehat{\rho}(\widetilde \bY_t(0)))\bigr)^2  
\lesssim \left(N^{-1}K_1  +T^{-\frac{2\beta}{2\beta+K}}  +n^{-1} \right)\log^4(T),
\end{align*}
and
\begin{align*}
\frac{1}{T_1} \sum_{t=1}^{T_1} \bigl\| \phi_i^\star(\bF_t^\star) - \widehat{\phi}_i(\widehat{\rho}(\widetilde{\bY}_{t}(0))) \bigr\|^2  
\lesssim& \left(n^{-1}K_1 + Nn^{-1}T^{-\frac{2\beta}{2\beta+K}} + Nn^{-2} \right)\log^{4}(T).
\end{align*}
\end{proposition}

\begin{proof}
For simplicity, denote $\cG_1=\mathcal{G}^{K_1}_{N}(L_1, \bd_1, B, T^{5\beta + 5}), \, \cG_2= \mathcal{G}^{1}_{K_1}(L_2, \bd_2, B, T^{5\beta + 5})$.
By the definition of $(\widehat{\phi}_1,\ldots,\widehat{\phi}_N)\circ \widehat{\rho}$, for any
$(\phi_1,\ldots,\phi_N)\circ \rho \in \mathcal{G}^{K_1}_{AE}$, it holds that
\[
\sum_{(i,t)\in\cA,\cB,\cC}\bigl(\widetilde Y_{it}(0)-\widehat{\phi}_i(\widehat{\rho}(\widetilde \bY_t(0)))\bigr)^2
\le
\sum_{(i,t)\in\cA,\cB,\cC}\bigl(\widetilde Y_{it}(0)-\phi_i(\rho(\widetilde \bY_t(0)))\bigr)^2 .
\]

As a consequence, we obtain
\begin{align}\label{proposition1:eq1}
&\sum_{(i,t)\in\cA,\cB,\cC}\bigl(\phi_i^\star(\bF^\star_{t}) -\widehat{\phi}_i(\widehat{\rho}(\widetilde \bY_t(0)))\bigr)^2 \notag\\
=&
\sum_{(i,t)\in\cA,\cB,\cC}\Bigl[
\bigl(\widetilde Y_{it}(0)-\widehat{\phi}_i(\widehat{\rho}(\widetilde \bY_t(0)))\bigr)^2
- \varepsilon_{it}^2 + 2\widehat{\phi}_i(\widehat{\rho}(\widetilde \bY_t(0)))\varepsilon_{it} - 2\phi_i^\star(\bF^\star_{t}) \varepsilon_{it}
\Bigr] \notag\\
\le&
\sum_{(i,t)\in\cA,\cB,\cC}\Bigl[
\bigl(\widetilde Y_{it}(0)-\phi_i(\rho(\widetilde \bY_t(0)))\bigr)^2
- \varepsilon_{it}^2 + 2\widehat{\phi}_i(\widehat{\rho}(\widetilde \bY_t(0)))\varepsilon_{it} - 2\phi_i^\star(\bF^\star_{t}) \varepsilon_{it}
\Bigr] \notag\\
=&
\sum_{(i,t)\in\cA,\cB,\cC}\Bigl[
\bigl(\phi_i^\star(\bF^\star_{t}) -\phi_i(\rho(\widetilde \bY_t(0)))\bigr)^2
+ 2\varepsilon_{it}\Bigl(\bigl(\phi_i^\star(\bF^\star_{t}) -\phi_i(\rho(\widetilde \bY_t(0)))\bigr)
+ \bigl(\widehat{\phi}_i(\widehat{\rho}(\widetilde \bY_t(0)))-\phi_i^\star(\bF^\star_{t}) \bigr)\Bigr)
\Bigr].
\end{align}

Below, we establish three inequalities, denoted (I) to (III).

\noindent (I): Assume $(\phi_1,\ldots,\phi_N)\circ \rho \in \mathcal{G}^{K_1}_{AE}$.
As long as $\phi_1,\ldots,\phi_N$ are deterministic, with probability at least $1-C\exp(-cT)$,
\begin{align*}
    & \sum_{(i,t)\in\cA,\cB,\cC} \varepsilon_{it}\bigl(\phi_i^\star(\bF^\star_{t}) -\phi_i(\rho(\widetilde \bY_t(0)))\bigr) \\ 
\lesssim & 
T^{1/2}K_1^{1/2}\log(T)\left(\sum_{(i,t)\in\cA,\cB,\cC}\bigl(\phi_i^\star(\bF^\star_{t}) -\phi_i(\rho(\widetilde \bY_t(0)))\bigr)^2\right)^{1/2}
+K_1^{1/2}T^{1/2}N^{1/2}\log(T)+N .
\end{align*}

\noindent (II): With probability at least $1-C\exp(-cT)$, we have
\begin{align*}
& \sum_{(i,t)\in\cA,\cB,\cC}\bigl(\widehat{\phi}_i(\widehat{\rho}(\widetilde \bY_t(0)))-\phi_i^\star(\bF^\star_{t}) \bigr)\varepsilon_{it} \\
\lesssim & 
\left(NT^{\frac{K}{2\beta+K}}+TK_1\right)^{1/2}\log^2(T)
\left(
\left(\sum_{(i,t)\in\cA,\cB,\cC}\bigl(\widehat{\phi}_i(\widehat{\rho}(\widetilde \bY_t(0)))-\phi_i^\star(\bF^\star_{t}) \bigr)^2\right)^{1/2}
+N^{1/2}
\right)
+N .
\end{align*}

\noindent (III): There exists $(\phi_1^{\dagger},\ldots,\phi_N^{\dagger})\circ \rho^{\dagger}\in \mathcal{G}^{K_1}_{AE}$
such that $\phi_1^{\dagger},\ldots,\phi_N^{\dagger}$ are deterministic and
\[
\frac{1}{NT}\sum_{(i,t)\in\cA,\cB,\cC}\bigl(\phi_i^\star(\bF^\star_{t}) -\phi_i^{\dagger}(\rho^{\dagger}(\widetilde \bY_t(0)))\bigr)^2
\lesssim
T^{-\frac{2\beta}{2\beta+K}}
+T^{-1}\inf_{\rho\in \mathcal{F}^{K}_{N}}\sum_{t=1}^{T}\bigl\|\rho(\widetilde \bY_t(0))-\bF_t^{\star}\bigr\|^2 .
\]

\noindent {\bf Step 1: Proof of (I)}.

Write
\[
\epsilon_T = K_1^{-1}T^{-1-(5\beta+5)(L_2+1)}d_2^{-L_2}
\]
and define the set
\begin{align}\label{proposition1:eq2}
S=\left\{x\in\mathbb{R}^{K_1}:\ \forall\,1\le i\le K_1,\ \exists\,1\le j(\in\mathbb{N})\le 2B\epsilon_T^{-1},
\ \text{s.t.}\ x_i=-B+j\epsilon_T \right\}.    
\end{align}

By definition, we see that for any $F\in[-B,B]^{K_1}$, there exists a $\bar{F}\in S$ such that
$\|F-\bar{F}\|_{\infty}\le \epsilon_T$. Write the set as $\{\bar{F}_k,\,1\le k\le |S|\}$.
It is not hard to verify that
\begin{align}\label{proposition1:eq3}
\log|S|\le K_1\log\!\Bigl(2BK_1T^{1+(5\beta+5)(L_2+1)}d_2^{L_2}\Bigr)
\lesssim K_1\log^2(T).
\end{align}

For $1\le t\le T$, assume $\|\rho(\widetilde \bY_t(0))-\bar{F}_{k_t^{\star}}\|_{\infty}\le \epsilon_T$. As a consequence, by Lemma~\ref{lemm5},
\begin{equation}\label{proposition1:eq4}
\bigl|\phi_i(\rho(\widetilde \bY_t(0)))-\phi_i(\bar{F}_{k_t^{\star}})\bigr|\le T^{-1}.
\end{equation}
Together with Lemma~\ref{lemm2}, with probability at least $1-C\exp(-cT)$,
\begin{align}
&\sum_{(i,t)\in\cA,\cB,\cC} \varepsilon_{it}\bigl(\phi_i^\star(\bF^\star_{t}) -\phi_i(\rho(\widetilde \bY_t(0)))\bigr) \notag\\
&\lesssim
\sum_{(i,t)\in\cA,\cB,\cC} \varepsilon_{it}\bigl(\phi_i^\star(\bF^\star_{t}) -\phi_i(\bar{F}_{k_t^{\star}})\bigr)+N \notag\\
&\le
\max_{\substack{k_t\in[|S|]\\ 1\le t\le T}}
\frac{\left|\sum_{(i,t)\in\cA,\cB,\cC} \varepsilon_{it}\bigl(\phi_i^\star(\bF^\star_{t}) -\phi_i(\bar{F}_{k_t})\bigr)\right|}
{\left(\sum_{(i,t)\in\cA,\cB,\cC}\bigl(\phi_i^\star(\bF^\star_{t}) -\phi_i(\bar{F}_{k_t})\bigr)^2\right)^{1/2}}
\left(\sum_{(i,t)\in\cA,\cB,\cC}\bigl(\phi_i^\star(\bF^\star_{t}) -\phi_i(\bar{F}_{k_t^{\star}})\bigr)^2\right)^{1/2}
+N
\notag\\
&\lesssim
\max_{\substack{k_t\in[|S|]\\ 1\le t\le T}}
\frac{\left|\sum_{(i,t)\in\cA,\cB,\cC} \varepsilon_{it}\bigl(\phi_i^\star(\bF^\star_{t}) -\phi_i(\bar{F}_{k_t})\bigr)\right|}
{\left(\sum_{(i,t)\in\cA,\cB,\cC}\bigl(\phi_i^\star(\bF^\star_{t}) -\phi_i(\bar{F}_{k_t})\bigr)^2\right)^{1/2}}
\left(
\left(\sum_{(i,t)\in\cA,\cB,\cC}\bigl(\phi_i^\star(\bF^\star_{t}) -\phi_i(\rho(\widetilde \bY_t(0)))\bigr)^2\right)^{1/2}
+N^{1/2}
\right)
+N, \label{proposition1:eq5}
\end{align}
where we use \eqref{proposition1:eq4} and Lemma~\ref{lemm2} in the first inequality and \eqref{proposition1:eq4} in the last inequality.
For any $u\ge 0$, by Lemma~\ref{lemm1} and the property of subGaussian distribution, as well as the fact that
$\phi_1,\ldots,\phi_N$, $\bar{F}_{k_1},\ldots,\bar{F}_{k_T}$, and $\{\phi_i^\star(\bF^\star_{t}) \}_{i,t}$ are deterministic given
$\{\bF_t^{\star}\}_{t=1}^{T}$,
\[
\mathbb{P}\!\left(
\left.
\frac{\left|\sum_{(i,t)\in\cA,\cB,\cC} \varepsilon_{it}\bigl(\phi_i^\star(\bF^\star_{t}) -\phi_i(\bar{F}_{k_t})\bigr)\right|}
{\left(\sum_{(i,t)\in\cA,\cB,\cC}\bigl(\phi_i^\star(\bF^\star_{t}) -\phi_i(\bar{F}_{k_t})\bigr)^2\right)^{1/2}}
>u
\ \right| \ \{\bF_t^{\star}\}_{t=1}^{T}
\right)
\le 2\exp\!\left(-\frac{u^2}{2\sigma_{\epsilon}^2}\right).
\]

Therefore, by union bound inequality and the arbitrariness of $\{\bF_t^{\star}\}_{t=1}^{T}$ in the above inequality,
\[
\mathbb{P}\!\left(
\max_{\substack{k_t\in[|S|]\\ 1\le t\le T}}
\frac{\left|\sum_{(i,t)\in\cA,\cB,\cC} \varepsilon_{it}\bigl(\phi_i^\star(\bF^\star_{t}) -\phi_i(\bar{F}_{k_t})\bigr)\right|}
{\left(\sum_{(i,t)\in\cA,\cB,\cC}\bigl(\phi_i^\star(\bF^\star_{t}) -\phi_i(\bar{F}_{k_t})\bigr)^2\right)^{1/2}}
>u
\right)
\le 2|S|^{T}\exp\!\left(-\frac{u^2}{2\sigma_{\epsilon}^2}\right).
\]
By \eqref{proposition1:eq3}, \eqref{proposition1:eq5}, and setting $u=2\sigma_{\epsilon}\,(T\log|S|)^{1/2}$, we complete the proof for (I).

\noindent {\bf Step 2: Proof of (II)}.

To prove this assertion, we rely on Lemma~\ref{lemm6}. The lemma states that for $\delta=T^{-1}$, there
exists a function class $\mathcal{G}_{2,\delta}\subset \mathcal{G}_2$ with
\begin{equation}\label{proposition1:eq6}
\log|\mathcal{G}_{2,\delta}|
\lesssim
\bigl(d_2^2 L_2 + K_1 d_2\bigr)\,
\log\!\Bigl( 8\delta^{-1} B T^{(5\beta+5)(L_2+1)} (1+d_2)^{L_2} K_1 L_2 \Bigr)
\lesssim
T^{\frac{K}{2\beta+K}}\log^4(T)
\end{equation}
such that for any $\phi\in \mathcal{G}_2$, there exists $\bar{\phi}\in \mathcal{G}_{2,\delta}$
meeting the requirement $|\bar{\phi}(x)-\phi(x)|\le T^{-1}$, $\forall\|x\|_\infty\le B$.
Since $\|\phi\|_\infty\le B$, we have $|\bar{\phi}(x)|\le T^{-1}+B$. Hence we can assume all the
functions in $\mathcal{G}_{2,\delta}$ are bounded by $2B$ for simplicity, i.e.,
$\mathcal{G}_{2,\delta}\in \mathcal{G}^{1}_{K_1}(L_2,\bd_2,2B,T^{5\beta+5})$.
Write $\mathcal{G}_{2,\delta}=\{\bar{\phi}_j,\,1\le j\le |\mathcal{G}_{2,\delta}|\}$.
By definition, there exists $\ell_i^{\star}\in [|\mathcal{G}_{2,\delta}|]$ such that
$|\widehat{\phi}_i(x)-\bar{\phi}_{\ell_i^{\star}}(x)|\le T^{-1}$, $\forall\|x\|_\infty\le B$.
Additionally, recall the definition of $S$ in \eqref{proposition1:eq2}, there exists $k_t^{\star}\in [|S|]$ such that
$\|\bar{F}_{k_t^{\star}}-\widehat{\rho}(\widetilde \bY_t(0))\|_\infty\le \epsilon_T$.
Since $\widehat{\phi}_i\in \mathcal{G}^{1}_{K_1}(L_2,\bd_2,B,T^{5\beta+5})$, by Lemma~\ref{lemm5}, we have
\[
\bigl|\widehat{\phi}_i(\widehat{\rho}(\widetilde \bY_t(0)))-\bar{\phi}_{\ell_i^{\star}}(\bar{F}_{k_t^{\star}})\bigr|
\le
\bigl|\widehat{\phi}_i(\widehat{\rho}(\widetilde \bY_t(0)))-\widehat{\phi}_i(\bar{F}_{k_t^{\star}})\bigr|
+
\bigl|\widehat{\phi}_i(\bar{F}_{k_t^{\star}})-\bar{\phi}_{\ell_i^{\star}}(\bar{F}_{k_t^{\star}})\bigr|
\le 2T^{-1}.
\]
Together with Lemma~\ref{lemm1}, with probability at least $1-C\exp(-cT)$ we have
\begin{align}\label{proposition1:eq7}
&\left|
\sum_{(i,t)\in\cA,\cB,\cC}\bigl(\widehat{\phi}_i(\widehat{\rho}(\widetilde \bY_t(0))) - \phi_i^\star(\bF^\star_{t}) \bigr)\varepsilon_{it}
\right| \notag\\
&\lesssim
\left|
\sum_{(i,t)\in\cA,\cB,\cC}\bigl(\bar{\phi}_{\ell_i^{\star}}(\bar{F}_{k_t^{\star}}) - \phi_i^\star(\bF^\star_{t}) \bigr)\varepsilon_{it}
\right|
+ N
\notag\\
&\le
\max_{\substack{k_t\in[|S|],\,1\le t\le T\\ \ell_i\in[|\mathcal{G}_{2,\delta}|],\,1\le i\le N}}
\frac{
\left|\sum_{(i,t)\in\cA,\cB,\cC}\bigl(\bar{\phi}_{\ell_i}(\bar{F}_{k_t})-\phi_i^\star(\bF^\star_{t}) \bigr)\varepsilon_{it}\right|
}{
\left(\sum_{(i,t)\in\cA,\cB,\cC}\bigl(\bar{\phi}_{\ell_i}(\bar{F}_{k_t})-\phi_i^\star(\bF^\star_{t}) \bigr)^2\right)^{1/2}
}
\left(
\sum_{(i,t)\in\cA,\cB,\cC}\bigl(\bar{\phi}_{\ell_i^{\star}}(\bar{F}_{k_t^{\star}})-\phi_i^\star(\bF^\star_{t}) \bigr)^2
\right)^{1/2}
+N
\notag\\
&\lesssim
\max_{\substack{k_t\in[|S|],\,1\le t\le T\\ \ell_i\in[|\mathcal{G}_{2,\delta}|],\,1\le i\le N}}
\frac{
\left|\sum_{(i,t)\in\cA,\cB,\cC}\bigl(\bar{\phi}_{\ell_i}(\bar{F}_{k_t})-\phi_i^\star(\bF^\star_{t}) \bigr)\varepsilon_{it}\right|
}{
\left(\sum_{(i,t)\in\cA,\cB,\cC}\bigl(\bar{\phi}_{\ell_i}(\bar{F}_{k_t})-\phi_i^\star(\bF^\star_{t}) \bigr)^2\right)^{1/2}
}
\left(
\left(\sum_{(i,t)\in\cA,\cB,\cC}\bigl(\widehat{\phi}_i(\widehat{\rho}(\widetilde \bY_t(0))) - \phi_i^\star(\bF^\star_{t}) \bigr)^2\right)^{1/2}
+ N^{1/2}
\right)
+N .
\end{align}
By Lemma~\ref{lemm1}, the property of sub-Gaussian distribution, as well as the union bound inequality
and the arbitrariness of $\{\bF_t^{\star}\}_{t=1}^{T}$, we obtain
\[
\mathbb{P}\!\left(
\max_{\substack{k_t\in[|S|],\,1\le t\le T\\ \ell_i\in[|\mathcal{F}_{2,\delta}|],\,1\le i\le N}}
\frac{
\left|\sum_{(i,t)\in\cA,\cB,\cC}\bigl(\bar{\phi}_{\ell_i}(\bar{F}_{k_t})-\phi_i^\star(\bF^\star_{t}) \bigr)\varepsilon_{it}\right|
}{
\left(\sum_{(i,t)\in\cA,\cB,\cC}\bigl(\bar{\phi}_{\ell_i}(\bar{F}_{k_t})-\phi_i^\star(\bF^\star_{t}) \bigr)^2\right)^{1/2}
}
>u
\right)
\le
2|S|^{T}\cdot |\mathcal{G}_{2,\delta}|^{N}\exp\!\left(-\frac{u^2}{2\sigma_{\epsilon}^2}\right).
\]
The conclusion follows by \eqref{proposition1:eq3}, \eqref{proposition1:eq6} and \eqref{proposition1:eq7} with
$u = 2\sigma_{\epsilon}\bigl(T\log|S| + N\log|\mathcal{G}_{2,\delta}|\bigr)^{1/2}$.

\noindent {\bf Step 3: Proof of (III)}.

It is sufficient to construct $(\phi_1^{\dagger},\ldots,\phi_N^{\dagger})\circ \rho^{\dagger}\in \mathcal{G}^{K}_{AE}$
satisfying the inequality. If $K_1>K$, we can introduce useless neurons into the bottleneck with
zero weights and biases, ensuring that the output of the network remains unchanged.

Assume
\[
\rho^{\dagger}(\cdot) := \arg\min_{\rho\in \mathcal{G}^{K}_{N}}
\sum_{t=1}^{T}\bigl\|\rho(\widetilde \bY_t(0))-\bF_t^{\star}\bigr\|^2 .
\]
According to Proposition~\ref{prop2}, there exist deterministic $\phi_i^{\dagger}\in \mathcal{G}^{K}_{1}(L_2,\bd_2,B,T^{5\beta+5})$ such that:
\begin{equation}\label{proposition1:eq8}
\bigl|\phi_i^{\dagger}(x)-\phi_i^{\star}(x)\bigr|
\lesssim
T^{-\frac{\beta}{2\beta+K}},
\qquad
\forall x\in[-B,B]^K \ \text{and}\ i=1,\ldots,N.
\end{equation}

By the triangle inequality, the following inequality holds:
\begin{align*}
& \frac{1}{NT}\sum_{(i,t)\in\cA,\cB,\cC}
\bigl(\phi_i^{\star}(\bF_t^{\star})-\phi_i^{\dagger}(\rho^{\dagger}(\widetilde \bY_t(0)))\bigr)^2 \\ 
&\le 
\frac{2}{NT}\sum_{(i,t)\in\cA,\cB,\cC}
\bigl(\phi_i^{\star}(\bF_t^{\star})-\phi_i^{\star}(\rho^{\dagger}(\widetilde \bY_t(0)))\bigr)^2 \\
&\quad+
\frac{2}{NT}\sum_{(i,t)\in\cA,\cB,\cC}
\bigl(\phi_i^{\star}(\rho^{\dagger}(\widetilde \bY_t(0)))-\phi_i^{\dagger}(\rho^{\dagger}(\widetilde \bY_t(0)))\bigr)^2
=: A_1 + A_2 .
\end{align*}

By Assumptions \ref{asmp:smooth}, we confirm:
\begin{equation}\label{proposition1:eq9}
\bigl|\phi_i^{\star}(x)-\phi_i^{\star}(y)\bigr|
\lesssim
\|x-y\|_2,
\qquad
\forall i=1,\ldots,N.
\end{equation}

Therefore,
\[
A_1 \lesssim \frac{1}{T}\sum_{t=1}^{T}\bigl\|\bF_t^{\star}-\rho^{\dagger}(\widetilde \bY_t(0))\bigr\|^2
=
\inf_{\rho\in\mathcal{G}^{K}_{N}}\frac{1}{T}\sum_{t=1}^{T}\bigl\|\rho(\widetilde \bY_t(0))-\bF_t^{\star}\bigr\|^2 .
\]
For $A_2$, by \eqref{proposition1:eq8}, we have $A_2\lesssim T^{-\frac{2\beta}{2\beta+K}}$, which completes the proof.

\noindent {\bf Step 4: Total in Sample Error}.

Note that by (I) and (III), it can be proven that, with probability at least $1-C\exp(-cT)$,
\begin{align*}
&\left|
\sum_{(i,t)\in\cA,\cB,\cC}\bigl(\phi_i^{\dagger}(\rho^{\dagger}(\widetilde \bY_t(0))) - \phi_i^\star(\bF^\star_{t}) \bigr)\varepsilon_{it}
\right| \\
&\lesssim
T^{1/2}K_1^{1/2}\log T
\left(
\left(\sum_{(i,t)\in\cA,\cB,\cC}\bigl(\phi_i^{\dagger}(\rho^{\dagger}(\widetilde \bY_t(0))) - \phi_i^\star(\bF^\star_{t}) \bigr)^2\right)^{1/2}
+ N^{1/2}
\right)
+N \\
&\lesssim
T^{1/2}N^{1/2}K_1^{1/2}\log T
\left(
T^{1/2}\left(
T^{-\frac{2\beta}{2\beta+K}}
+T^{-1}\inf_{\rho\in\mathcal{G}^{K}_{N}}
\sum_{t=1}^{T}\bigl\|\rho(\widetilde \bY_t(0))-\bF_t^{\star}\bigr\|^2
\right)^{1/2}
+1
\right)
+N \\
&\lesssim
NT\left(
N^{-1}K_1
+T^{-\frac{2\beta}{2\beta+K}}
+T^{-1}\inf_{\rho\in\mathcal{G}^{K}_{N}}
\sum_{t=1}^{T}\bigl\|\rho(\widetilde \bY_t(0))-\bF_t^{\star}\bigr\|^2
\right)\log^2(T).
\end{align*}

Together with (II), (III) and \eqref{proposition1:eq1} with $(\phi_1,\ldots,\phi_N)\circ \rho$ on the right hand side
replaced by $(\phi_1^{\dagger},\ldots,\phi_N^{\dagger})\circ \rho^{\dagger}$, we conclude that with probability
at least $1-C\exp(-cT)$,
\begin{align*}
&\frac{1}{NT}\sum_{(i,t)\in\cA,\cB,\cC}\bigl(\phi_i^{\star}(\bF_t^{\star})-\widehat{\phi}_i(\widehat{\rho}(\widetilde \bY_t(0)))\bigr)^2 \\
\lesssim&
\left(T^{-\frac{2\beta}{2\beta+K}} + N^{-1}K_1\right)^{1/2}\log^2(T)
\left(
\frac{1}{NT}\sum_{(i,t)\in\cA,\cB,\cC}\bigl(\widehat{\phi}_i(\widehat{\rho}(\widetilde \bY_t(0)))-\phi_i^{\star}(\bF_t^{\star})\bigr)^2
\right)^{1/2} \\
&\;+\;
\left(
N^{-1}K_1
+T^{-\frac{2\beta}{2\beta+K}}
+T^{-1}\inf_{\rho\in\mathcal{G}^{K}_{N}}
\sum_{t=1}^{T}\bigl\|\rho(\widetilde \bY_t(0))-\bF_t^{\star}\bigr\|^2
\right)\log^2(T).
\end{align*}
By basic inequality $x \le a\sqrt{x}+b \Rightarrow x \le a^2+4b$, we have 
\begin{align}\label{proposition1:eq10}
&\frac{1}{NT}\sum_{(i,t)\in\cA,\cB,\cC}\bigl(\phi_i^{\star}(\bF_t^{\star})-\widehat{\phi}_i(\widehat{\rho}(\widetilde \bY_t(0)))\bigr)^2 
\lesssim \left(
N^{-1}K_1
+T^{-\frac{2\beta}{2\beta+K}}
+T^{-1}\inf_{\rho\in\mathcal{G}^{K}_{N}}
\sum_{t=1}^{T}\bigl\|\rho(\widetilde \bY_t(0))-\bF_t^{\star}\bigr\|^2
\right)\log^4(T).
\end{align}

Notably, since $\|\phi^{\star}\|_{\infty}\le B$ and $\|M^{\star}\|_{\infty}\lesssim n^{-1}$, $\|M^{\star}\|_{0}\asymp n$
by Assumption \ref{asmp:factor}, we observe that the domain, $M^{\star}\phi^{\star}([-B,B]^K)$, a compact set, is contained
within a bounded set $[-\tilde{C},\tilde{C}]^K$. Therefore, by the Whitney extension theorem, $\rho^{\star}$ can be
extended to \((\beta, C)\)-H\"older smooth on $[-\tilde{C}-1,\tilde{C}+1]^K$ for some fixed constant $C$.

According to Proposition~\ref{prop2}, there exist $\tilde{\rho}^{\dagger}\in \mathcal{G}^{K}_{K}(L_1-1,\bd_1,B,T^{5\beta+5})$ such that:
\begin{equation}\label{proposition1:eq11}
\bigl\|\tilde{\rho}^{\dagger}(x+1+\tilde{C})-\rho^{\star}(x)\bigr\|_{\infty}
\lesssim
T^{-\frac{\beta}{2\beta+K}},
\qquad
\forall x\in[-\tilde{C}-1,\tilde{C}+1]^K.
\end{equation}
Let $v^{\dagger}\in\mathbb{R}^{K}$ be a vector with all elements equal to $-1-\tilde{C}$ and define
$\rho^{\dagger}(x):=\tilde{\rho}^{\dagger}(\bar\sigma(M^{\star}x-v^{\dagger}))\in \mathcal{G}^{K}_{K}(L_1,\bd_1,B,T^{5\beta+5})$.
By Lemma~\ref{lemm3}, with probability at least $1-\exp(-cn)$, $\|M^{\star}\widetilde\bY_t(0)\|_{\infty}\le 1+\tilde{C}$.
As a consequence, by Lemma~\ref{lemm4}, \eqref{proposition1:eq11} and the fact that $\rho^{\star}$ is Lipschitz continuous, with probability at least $1-C\exp(-cT)-C\exp(-cn)$,
\begin{align*}
&\frac{1}{T}\sum_{t=1}^{T}\bigl\|\bF_t^{\star}-\rho^{\dagger}(\widetilde\bY_t(0))\bigr\|^2 \\
\lesssim&
\frac{1}{T}\sum_{t=1}^{T}\Bigl(
\bigl\|\bF_t^{\star}-\rho^{\star}(M^{\star}\bphi^{\star}(\bF_t^{\star}))\bigr\|^2
+
\bigl\|\rho^{\star}(M^{\star}\bphi^{\star}(\bF_t^{\star}))-\rho^{\star}(M^{\star}\widetilde\bY_t(0))\bigr\|^2 
+
\bigl\|\rho^{\star}(M^{\star}\widetilde\bY_t(0))-\tilde{\rho}^{\dagger}(\bar\sigma(M^{\star}\widetilde\bY_t(0)-v^{\dagger}))\bigr\|^2
\Bigr) \\
\lesssim&
\frac{1}{T}\sum_{t=1}^{T}\Bigl( \|M^{\star}\varepsilon_t\|^2 +
\bigl\|\rho^{\star}(M^{\star}\widetilde\bY_t(0))-\tilde{\rho}^{\dagger}(M^{\star}\widetilde\bY_t(0)+1+\tilde{C})\bigr\|^2
\Bigr) \\
\lesssim &
n^{-1}+T^{-\frac{2\beta}{2\beta+K}}.
\end{align*}
Substituting into \eqref{proposition1:eq10}, we have
\begin{align}\label{proposition1:eq12}
&\frac{1}{NT}\sum_{(i,t)\in\cA,\cB,\cC}\bigl(\phi_i^{\star}(\bF_t^{\star})-\widehat{\phi}_i(\widehat{\rho}(\widetilde \bY_t(0)))\bigr)^2  
\lesssim \left(N^{-1}K_1  +T^{-\frac{2\beta}{2\beta+K}}  +n^{-1} \right)\log^4(T).
\end{align}

\noindent {\bf Step 5: Unit Level in Sample Error}.

By Assumption \ref{asmp:factor}, \eqref{proposition1:eq9}, $\|M^{\star}\|^2 \le \mathrm{Tr}((M^{\star})^{\top}M^{\star}) \lesssim n^{-1}$,
and \eqref{proposition1:eq12}, we have with probability at least $1-C\exp(-cT)-C\exp(-cn)$,
\begin{align*}
&\frac{1}{T_1}\sum_{t=1}^{T_1}\bigl\|\rho^{\star}(M^{\star} \widehat{\bphi}(\widehat{\rho}(\widetilde \bY_t(0))) )-\bF_t^{\star}\bigr\|^2 \\
\lesssim&\frac{1}{T_1}\sum_{t=1}^{T_1}\bigl\|\rho^{\star}(M^{\star} \widehat{\bphi}(\widehat{\rho}(\widetilde \bY_t(0))) )-\rho^{\star}\!\bigl(M^{\star}\bphi^{\star}(\bF_t^{\star})\bigr)\bigr\|^2
+
\frac{1}{T_1}\sum_{t=1}^{T_1}\bigl\|\rho^{\star}\!\bigl(M^{\star}\bphi^{\star}(\bF_t^{\star})\bigr)-\bF_t^{\star}\bigr\|^2
\\
\lesssim&
\frac{1}{T_1}\sum_{t=1}^{T_1}\bigl\|M^{\star} \widehat{\bphi}(\widehat{\rho}(\widetilde \bY_t(0))) - M^{\star}\bphi^{\star}(\bF_t^{\star})\bigr\|^2 + n^{-1}
\lesssim
\frac{1}{T_1}\sum_{t=1}^{T_1}\|M^{\star}\|^2\,\bigl\| \widehat{\bphi}(\widehat{\rho}(\widetilde \bY_t(0))) -\bphi^{\star}(\bF_t^{\star})\bigr\|^2 + n^{-1}
\\
\lesssim&
\frac{1}{T_1 n}\sum_{t=1}^{T_1}\bigl\|\widehat{\bphi}(\widehat{\rho}(\widetilde \bY_t(0)))-\bphi^{\star}(\bF_t^{\star})\bigr\|^2 + n^{-1}
\lesssim
\left(n^{-1}K_1 + Nn^{-1}T^{-\frac{2\beta}{2\beta+K}} + Nn^{-2} \right)\log^{4}(T).
\end{align*}
Let $\widehat\bF_t=\widehat{\rho}(\widetilde{\bY}_{t}(0))$.
By Assumption \ref{asmp:factor}, \eqref{proposition1:eq9}, and the fact that $\rho^{\star}$ is Lipschitz continuous, we have with probability at least $1-C\exp(-cT)-C\exp(-cn)$,
\begin{align*}
&\frac{1}{T_1} \sum_{t=1}^{T_1} \bigl\| \phi_i^\star(\bF_t^\star) - \widehat{\phi}_i(\widehat{\rho}(\widetilde{\bY}_{t}(0))) \bigr\|^2  \\ 
\lesssim&  \frac{1}{T_1} \sum_{t=1}^{T_1} \bigl\| \phi_i^\star (\rho^{\star}( M^{\star}\widehat{\phi}(\widehat\bF_t))) - \phi_i^\star(\bF_t^\star) \bigr\|^2 + \frac{1}{T_1} \sum_{t=1}^{T_1} \bigl\| \phi_i^\star (\rho^{\star}( M^{\star}\widehat{\phi}(\widehat\bF_t)))  - \widehat{\phi}_i(\widehat\bF_t) \bigr\|^2 \\
\lesssim&  \frac{1}{T_1} \sum_{t=1}^{T_1} \bigl\| \rho^{\star}( M^{\star}\widehat{\phi}(\widehat\bF_t)) - \bF_t^\star \bigr\|^2 + n^{-1} \\
\lesssim& \left(n^{-1}K_1 + Nn^{-1}T^{-\frac{2\beta}{2\beta+K}} + Nn^{-2} \right)\log^{4}(T).
\end{align*}

\end{proof}

\begin{proposition}\label{prop2}
Assume $f$ is \((p, C)\)-H\"older smooth on $[-a,a]^r$ for some $p=q+s$, $q\in \mathbb{N}_0$ and $s\in(0,1)$, and $C>0$.
Then for $a\ge 1$ and $\zeta$ sufficiently large (only depends on fixed constants including $p,a,C$ and $\|f\|_{C^{q}}$),
when
\[
L \asymp \left\lceil \log_{4}\bigl(\zeta^{2p}\bigr)\right\rceil\cdot
\left(\left\lceil \log_{2}\bigl(\max\{q,r\}+1\bigr)\right\rceil + 1\right) + 1
\]
and
\[
d=\max_{1\le\ell\le L}d_{\ell} \asymp 2^{r}\cdot \binom{r+q}{r}\cdot r^{2}\cdot (q+1)\cdot \zeta^{r},
\]
there exists a neural network $\hat{f}_{\mathrm{wide}}\in \mathcal{G}^{1}_{r}\bigl(L,\bd,B,\zeta^{5p+5}\bigr)$ such that
\[
\bigl\|f-\hat{f}_{\mathrm{wide}}\bigr\|_{\infty}\le C\,\zeta^{-2p},
\]
for some constant $C$ only depends on fixed parameters.
\end{proposition}
Proposition~\ref{prop2} is proved in Proposition C1 of \citet{xiu2024deep}, and follows the same argument as in \citet{kohler2021rate}.

\begin{lemma}\label{lemm1}
Under Assumption~\ref{asmp:density}, given $\{\bF_t^{\star}\}_{t=1}^{T}$, for any constants
$\{C_{it}\}_{i=1,\ldots,N,\;t=1,\ldots,T}$,
\[
\frac{\sum_{(i,t)\in\cA,\cB,\cC} C_{it}\varepsilon_{it}}
{\left(\sum_{(i,t)\in\cA,\cB,\cC} C_{it}^{2}\right)^{1/2}}
\]
is subGaussian with its subGaussian norm bounded by some fixed constant $\sigma_{\epsilon}^{2}$.
\end{lemma}

\begin{proof}
Recall that the subGaussian norm of $\varepsilon_{it}$ is bounded by $\sigma_{\epsilon}^{2}$.
Let $C=\{C_{it}\}_{(i,t)\in\cA,\cB,\cC},\varepsilon=\{\varepsilon_{it}\}_{(i,t)\in\cA,\cB,\cC}$, we have
\[
\frac{\sum_{(i,t)\in\cA,\cB,\cC} C_{it}\varepsilon_{it}}
{\left(\sum_{(i,t)\in\cA,\cB,\cC} C_{it}^{2}\right)^{1/2}}
=
\frac{\mathrm{vec}(C)^{\top}\mathrm{vec}(\varepsilon)}
{\left(\sum_{(i,t)\in\cA,\cB,\cC} C_{it}^{2}\right)^{1/2}}.
\]
Since $\mathrm{vec}(\varepsilon)$ is composed of independent subGaussian random variables with bounded subGaussian norm,
the subGaussian norm of the right hand side is bounded by
\[
\frac{\sigma_{\epsilon}^{2}\,\mathrm{vec}(C)^{\top} \mathrm{vec}(C)}
{\sum_{(i,t)\in\cA,\cB,\cC} C_{it}^{2}}
\le
\frac{\sigma_{\epsilon}^{2} \|\mathrm{vec}(C)\|^{2}}
{\sum_{(i,t)\in\cA,\cB,\cC} C_{it}^{2}}
=
\sigma_{\epsilon}^{2} .
\]
\end{proof}

\begin{lemma}\label{lemm2}
Under Assumption~\ref{asmp:density}, with probability at least $1-C\exp(-cT)$,
\[
\sum_{t=1}^{T}\|\varepsilon_{t}\|_{1}\lesssim NT.
\]
\end{lemma}
\begin{proof}
Recall that the subGaussian norm of $\varepsilon_{it}$ is bounded by $\sigma_{\epsilon}^2$. Observe that
\begin{align*}
\sum_{t=1}^{T}\|\varepsilon_{t}\|_{1}
&\le \sum_{t=1}^{T} N^{1/2}\|\varepsilon_{t}\|
\le N^{1/2}T^{1/2}\left(\sum_{i=1}^{N}\sum_{t=1}^{T}\varepsilon_{it}^2\right)^{1/2} \\
&= N^{1/2}T^{1/2}\bigl(\mathrm{vec}(\varepsilon)^{\top} \mathrm{vec}(\varepsilon)\bigr)^{1/2} .
\end{align*}
Therefore, we only need to prove $N^{-1}T^{-1}\mathrm{vec}(\varepsilon)^{\top}\mathrm{vec}(\varepsilon)$ is bounded by some constant.
By Hanson--Wright inequality,
\[
\mathbb{P}\!\left(
\frac{1}{NT}\sum_{i=1}^{N}\sum_{t=1}^{T}\bigl(\varepsilon_{it}^{2}-\mathbb{E}\varepsilon_{it}^{2}\bigr)\ge \sigma_{\epsilon}^{2}
\right)
\le 2\exp(-cNT).
\]
By the property of sub-Gaussian variable, we have $\mathbb{E}\varepsilon_{it}^{2}\le 2\sigma_{\epsilon}^{2}$.
As a consequence, it holds that
\[
\frac{1}{NT}\sum_{i=1}^{N}\sum_{t=1}^{T} \varepsilon_{it}^{2}\le 3\sigma_{\epsilon}^{2}
\]
with probability at least $1-C\exp(-cT)$.
\end{proof}

\begin{lemma}\label{lemm3}
Under Assumptions~\ref{asmp:density} and \ref{asmp:factor}, if $\log T=o(n)$, with probability at least $1-C\exp(-cn)$,
\[
\|M^{\star}\varepsilon_{t}\|_{\infty}\le 1 \quad \text{for all } 1\le t\le T.
\]
\end{lemma}

\begin{proof}
For $i=1,\ldots,K$, write $M_i^{\star}$ as the $i$-th row of $M^{\star}$.
Then we have
\[
\|M^{\star}\varepsilon_{t}\|_{\infty}=\max_{i=1,\ldots,K}\bigl\|M_i^{\star}\varepsilon_{t}\bigr\|.
\]
Observe that $M_i^{\star}\varepsilon_{t}$ is a subGaussian random variable with its subGaussian norm bounded by
\[
\sigma_{\epsilon}^{2}\, M_i^{\star}M_i^{\star\top}
\lesssim
\|M_i^{\star}\|^{2}
\lesssim
n^{-1} .
\]
By the property of subGaussian variable,
\[
\mathbb{P}\!\left(\bigl|M_i^{\star}\varepsilon_{t}\bigr|\ge 1\right)\le C\exp(-cn).
\]
Together with union bound inequality and the fact that $\log(T)=o(L)$,
\[
\mathbb{P}\!\left(
\max_{1\le t\le T}\max_{1\le i\le K}\bigl|M_i^{\star}\varepsilon_{t}\bigr|\ge 1
\right)
\le CTK\exp(-cn)\lesssim C\exp(-cn/2).
\]
\end{proof}

\begin{lemma}\label{lemm4}
With probability at least $1-C\exp(-cT)$,
\[
T^{-1}\sum_{t=1}^{T}\|M^{\star}\varepsilon_{t}\|^{2}\lesssim n^{-1}.
\]
\end{lemma}

\begin{proof}
Let $\varepsilon=(\varepsilon_1,...,\varepsilon_T)\in\mathbb R^{N\times T}$ and $Q=I_T\otimes M^{\star \top}M^{\star}$. Observe that
\[
T^{-1}\sum_{t=1}^{T}\|M^{\star}\varepsilon_{t}\|^{2}
= T^{-1}\mathrm{vec}(\varepsilon)^{\top}Q \mathrm{vec}(\varepsilon) .
\]
Note that $\|Q\|\le \|M^{\star \top}M^{\star}\|\le \mathrm{Tr}((M^{\star})^{\top}M^{\star})\lesssim n^{-1}$.
Additionally, observe that $\mathrm{rank}(Q)\le T\,\mathrm{rank}(M^{\star}) = TK$.
As a consequence, we have
\[
T^{-1}\mathrm{Tr}( Q )
\le \|T^{-1} Q \|\,\mathrm{rank}( Q )
\lesssim n^{-1},
\]
\[
\bigl\|T^{-1} Q \bigr\|_{F}^{2}
\lesssim \bigl\|T^{-1} Q \bigr\|^{2}\,
\mathrm{rank}( Q )
\lesssim T^{-1} n^{-2}.
\]
By applying Hanson--Wright inequality, we conclude that
\[
\mathbb{P}\!\left(
\left|(T^{-1}\mathrm{vec}(\varepsilon)^{\top} Q \mathrm{vec}(\varepsilon)-\mathbb{E} T^{-1}\mathrm{vec}(\varepsilon)^{\top} Q \mathrm{vec}(\varepsilon)\right|
\ge n^{-1}
\right)
\le C\exp(-cT).
\]
Together with the fact that
\[
\mathbb{E}\,T^{-1}\mathrm{vec}(\varepsilon)^{\top} Q \mathrm{vec}(\varepsilon)
\asymp T^{-1}\mathrm{Tr}( Q )
\lesssim n^{-1},
\]
we complete the proof.
\end{proof}

\begin{lemma}\label{lemm5}
For any $\phi\in \mathcal{G}^{d_{L+1}}_{d_0}(L,\bd,T^{5\beta+5})$, it holds that
\[
|\phi(x)-\phi(y)|
\le d_0T^{(5\beta+5)(L+1)}\,d^{L}\,\|x-y\|_{\infty}.
\]
\end{lemma}

\begin{proof}
Write $\phi(x)=M_{L+1}\bar\sigma_{v_L}M_{L}\bar\sigma_{v_{L-1}}\cdots M_2\bar\sigma_{v_1}M_1x$
where $M_i\in\mathbb{R}^{d_{i}\times d_{i-1}}$ and $\bar\sigma_{v_i}M_{i}x=M_{i}x-v_i$.
It is not hard to verify that
$\|M_i x - M_i y\|_{\infty}\le d_{i-1} T^{5\beta+5}\|x-y\|_{\infty}$
and $\|\bar\sigma_{v_i}x-\bar\sigma_{v_i}y\|_{\infty}\le \|x-y\|_{\infty}$.
Note that for two Lipschitz functions $f_1$ and $f_2$ satisfying
$\|f_1(x)-f_1(y)\|_{\infty}\le c_1\|x-y\|_{\infty}$ and
$\|f_2(x)-f_2(y)\|_{\infty}\le c_2\|x-y\|_{\infty}$, it holds that
\[
\|f_2\circ f_1(x) - f_2\circ f_1(y)\|_{\infty}\le c_1 c_2\|x-y\|_{\infty}.
\]
By repeatedly applying this fact to $M_{L+1}\bar\sigma_{v_L}M_{L}\bar\sigma_{v_{L-1}}\cdots M_2\bar\sigma_{v_1}M_1x$, we have
\[
\|\phi(x)-\phi(y)\|
\le
T^{(5\beta+5)(L+1)}\left(\prod_{i=0}^{L} d_i\right)\|x-y\|_{\infty}
\le
d_0 T^{(5\beta+5)(L+1)}\,d^{L}\,\|x-y\|_{\infty}.
\qquad \Box
\]
\end{proof}

\begin{lemma}\label{lemm6}
For the function class $\mathcal{G}^{d_{L+1}}_{d_0}(L,\bd,B,T^{5\beta+5})$ with width vector
$\bd=(d_1,\ldots,d_{L+1})$, assume the total number of non-zero weights are bounded by $S$.
There exists a function class
$\mathcal{G}^{d_{L+1}}_{d_0,\delta}\subset \mathcal{G}^{d_{L+1}}_{d_0}(L,\bd,B,T^{5\beta+5})$ such that
\begin{enumerate}
\item[(a)]
\[
\bigl|\mathcal{G}^{d_{L+1}}_{d_0,\delta}\bigr|
\le
\left(
8\delta^{-1}C T^{(5\beta+5)(L+1)}(1+d)^{L} d_0 d_{L+1}L
\right)^{2S};
\]
\item[(b)]
For any $\phi\in \mathcal{G}^{d_{L+1}}_{d_0}(L,\bd,B,T^{5\beta+5})$, there exists
$\bar{\phi}\in \mathcal{G}^{d_{L+1}}_{d_0,\delta}$ satisfying
\[
\|\phi(x)-\bar{\phi}(x)\|_{\infty}\le \delta
\qquad \text{whenever } \|x\|_{\infty}\le C.
\]
\end{enumerate}
\end{lemma}
Lemma~\ref{lemm6} is proved in Lemma 7 of \citet{xiu2024deep}, and follows the same idea as Lemma 5 of \citet{schmidt2020nonparametric}.

\clearpage
\newpage
\section*{NeurIPS Paper Checklist}

\begin{enumerate}
\item {\bf Claims}
    \item[] Question: Do the main claims made in the abstract and introduction accurately reflect the paper's contributions and scope?
    \item[] Answer: \answerYes{} %
    \item[] Justification: All claims made in the abstract and introduction accurately reflect the paper’s contributions and scope.
    \item[] Guidelines:
    \begin{itemize}
        \item The answer \answerNA{} means that the abstract and introduction do not include the claims made in the paper.
        \item The abstract and/or introduction should clearly state the claims made, including the contributions made in the paper and important assumptions and limitations. A \answerNo{} or \answerNA{} answer to this question will not be perceived well by the reviewers. 
        \item The claims made should match theoretical and experimental results, and reflect how much the results can be expected to generalize to other settings. 
        \item It is fine to include aspirational goals as motivation as long as it is clear that these goals are not attained by the paper. 
    \end{itemize}

\item {\bf Limitations}
    \item[] Question: Does the paper discuss the limitations of the work performed by the authors?
    \item[] Answer: \answerYes{} %
    \item[] Justification: The limitations of this work are discussed in Appendix \ref{append:limitation}.
    \item[] Guidelines:
    \begin{itemize}
        \item The answer \answerNA{} means that the paper has no limitation while the answer \answerNo{} means that the paper has limitations, but those are not discussed in the paper. 
        \item The authors are encouraged to create a separate ``Limitations'' section in their paper.
        \item The paper should point out any strong assumptions and how robust the results are to violations of these assumptions (e.g., independence assumptions, noiseless settings, model well-specification, asymptotic approximations only holding locally). The authors should reflect on how these assumptions might be violated in practice and what the implications would be.
        \item The authors should reflect on the scope of the claims made, e.g., if the approach was only tested on a few datasets or with a few runs. In general, empirical results often depend on implicit assumptions, which should be articulated.
        \item The authors should reflect on the factors that influence the performance of the approach. For example, a facial recognition algorithm may perform poorly when image resolution is low or images are taken in low lighting. Or a speech-to-text system might not be used reliably to provide closed captions for online lectures because it fails to handle technical jargon.
        \item The authors should discuss the computational efficiency of the proposed algorithms and how they scale with dataset size.
        \item If applicable, the authors should discuss possible limitations of their approach to address problems of privacy and fairness.
        \item While the authors might fear that complete honesty about limitations might be used by reviewers as grounds for rejection, a worse outcome might be that reviewers discover limitations that aren't acknowledged in the paper. The authors should use their best judgment and recognize that individual actions in favor of transparency play an important role in developing norms that preserve the integrity of the community. Reviewers will be specifically instructed to not penalize honesty concerning limitations.
    \end{itemize}

\item {\bf Theory assumptions and proofs}
    \item[] Question: For each theoretical result, does the paper provide the full set of assumptions and a complete (and correct) proof?
    \item[] Answer: \answerYes{} %
    \item[] Justification: The theoretical results, assumptions, and proofs are presented in Section \ref{subsec:theoretical-results} and Appendix \ref{section:proofs}.
    \item[] Guidelines:
    \begin{itemize}
        \item The answer \answerNA{} means that the paper does not include theoretical results. 
        \item All the theorems, formulas, and proofs in the paper should be numbered and cross-referenced.
        \item All assumptions should be clearly stated or referenced in the statement of any theorems.
        \item The proofs can either appear in the main paper or the supplemental material, but if they appear in the supplemental material, the authors are encouraged to provide a short proof sketch to provide intuition. 
        \item Inversely, any informal proof provided in the core of the paper should be complemented by formal proofs provided in appendix or supplemental material.
        \item Theorems and Lemmas that the proof relies upon should be properly referenced. 
    \end{itemize}

    \item {\bf Experimental result reproducibility}
    \item[] Question: Does the paper fully disclose all the information needed to reproduce the main experimental results of the paper to the extent that it affects the main claims and/or conclusions of the paper (regardless of whether the code and data are provided or not)?
    \item[] Answer: \answerYes{} %
    \item[] Justification: Methodological details of the proposed methods are provided in Section \ref{subsec:method-CoDEAL}, and experimental details necessary to reproduce the results are provided in Sections \ref{sec:simulation}-\ref{sec:realdata} and Appendices \ref{append:simulation}-\ref{append: real}.
    \item[] Guidelines:
    \begin{itemize}
        \item The answer \answerNA{} means that the paper does not include experiments.
        \item If the paper includes experiments, a \answerNo{} answer to this question will not be perceived well by the reviewers: Making the paper reproducible is important, regardless of whether the code and data are provided or not.
        \item If the contribution is a dataset and\slash or model, the authors should describe the steps taken to make their results reproducible or verifiable. 
        \item Depending on the contribution, reproducibility can be accomplished in various ways. For example, if the contribution is a novel architecture, describing the architecture fully might suffice, or if the contribution is a specific model and empirical evaluation, it may be necessary to either make it possible for others to replicate the model with the same dataset, or provide access to the model. In general. releasing code and data is often one good way to accomplish this, but reproducibility can also be provided via detailed instructions for how to replicate the results, access to a hosted model (e.g., in the case of a large language model), releasing of a model checkpoint, or other means that are appropriate to the research performed.
        \item While NeurIPS does not require releasing code, the conference does require all submissions to provide some reasonable avenue for reproducibility, which may depend on the nature of the contribution. For example
        \begin{enumerate}
            \item If the contribution is primarily a new algorithm, the paper should make it clear how to reproduce that algorithm.
            \item If the contribution is primarily a new model architecture, the paper should describe the architecture clearly and fully.
            \item If the contribution is a new model (e.g., a large language model), then there should either be a way to access this model for reproducing the results or a way to reproduce the model (e.g., with an open-source dataset or instructions for how to construct the dataset).
            \item We recognize that reproducibility may be tricky in some cases, in which case authors are welcome to describe the particular way they provide for reproducibility. In the case of closed-source models, it may be that access to the model is limited in some way (e.g., to registered users), but it should be possible for other researchers to have some path to reproducing or verifying the results.
        \end{enumerate}
    \end{itemize}

\item {\bf Open access to data and code}
    \item[] Question: Does the paper provide open access to the data and code, with sufficient instructions to faithfully reproduce the main experimental results, as described in supplemental material?
    \item[] Answer: \answerYes{} %
    \item[] Justification: Our code is provided in the supplement. We use open-access datasets, and we cite original sources of this dataset in Section \ref{sec:realdata}.
    \item[] Guidelines:
    \begin{itemize}
        \item The answer \answerNA{} means that paper does not include experiments requiring code.
        \item Please see the NeurIPS code and data submission guidelines (\url{https://neurips.cc/public/guides/CodeSubmissionPolicy}) for more details.
        \item While we encourage the release of code and data, we understand that this might not be possible, so \answerNo{} is an acceptable answer. Papers cannot be rejected simply for not including code, unless this is central to the contribution (e.g., for a new open-source benchmark).
        \item The instructions should contain the exact command and environment needed to run to reproduce the results. See the NeurIPS code and data submission guidelines (\url{https://neurips.cc/public/guides/CodeSubmissionPolicy}) for more details.
        \item The authors should provide instructions on data access and preparation, including how to access the raw data, preprocessed data, intermediate data, and generated data, etc.
        \item The authors should provide scripts to reproduce all experimental results for the new proposed method and baselines. If only a subset of experiments are reproducible, they should state which ones are omitted from the script and why.
        \item At submission time, to preserve anonymity, the authors should release anonymized versions (if applicable).
        \item Providing as much information as possible in supplemental material (appended to the paper) is recommended, but including URLs to data and code is permitted.
    \end{itemize}

\item {\bf Experimental setting/details}
    \item[] Question: Does the paper specify all the training and test details (e.g., data splits, hyperparameters, how they were chosen, type of optimizer) necessary to understand the results?
    \item[] Answer: \answerYes{} %
    \item[] Justification: Experimental settings and details are provided in Sections \ref{sec:simulation} and Appendix \ref{append:simulation}.
    \item[] Guidelines:
    \begin{itemize}
        \item The answer \answerNA{} means that the paper does not include experiments.
        \item The experimental setting should be presented in the core of the paper to a level of detail that is necessary to appreciate the results and make sense of them.
        \item The full details can be provided either with the code, in appendix, or as supplemental material.
    \end{itemize}

\item {\bf Experiment statistical significance}
    \item[] Question: Does the paper report error bars suitably and correctly defined or other appropriate information about the statistical significance of the experiments?
    \item[] Answer: \answerYes{} %
    \item[] Justification: Standard errors of MAE and MSE from multiple replications are reported along with the experiment results. 
    \item[] Guidelines:
    \begin{itemize}
        \item The answer \answerNA{} means that the paper does not include experiments.
        \item The authors should answer \answerYes{} if the results are accompanied by error bars, confidence intervals, or statistical significance tests, at least for the experiments that support the main claims of the paper.
        \item The factors of variability that the error bars are capturing should be clearly stated (for example, train/test split, initialization, random drawing of some parameter, or overall run with given experimental conditions).
        \item The method for calculating the error bars should be explained (closed form formula, call to a library function, bootstrap, etc.)
        \item The assumptions made should be given (e.g., Normally distributed errors).
        \item It should be clear whether the error bar is the standard deviation or the standard error of the mean.
        \item It is OK to report 1-sigma error bars, but one should state it. The authors should preferably report a 2-sigma error bar than state that they have a 96\% CI, if the hypothesis of Normality of errors is not verified.
        \item For asymmetric distributions, the authors should be careful not to show in tables or figures symmetric error bars that would yield results that are out of range (e.g., negative error rates).
        \item If error bars are reported in tables or plots, the authors should explain in the text how they were calculated and reference the corresponding figures or tables in the text.
    \end{itemize}

\item {\bf Experiments compute resources}
    \item[] Question: For each experiment, does the paper provide sufficient information on the computer resources (type of compute workers, memory, time of execution) needed to reproduce the experiments?
    \item[] Answer: \answerYes{} %
    \item[] Justification: Compute resources needed to reproduce the experiments are reported in Appendix \ref{appen:Experimental Setting}.
    \item[] Guidelines:
    \begin{itemize}
        \item The answer \answerNA{} means that the paper does not include experiments.
        \item The paper should indicate the type of compute workers CPU or GPU, internal cluster, or cloud provider, including relevant memory and storage.
        \item The paper should provide the amount of compute required for each of the individual experimental runs as well as estimate the total compute. 
        \item The paper should disclose whether the full research project required more compute than the experiments reported in the paper (e.g., preliminary or failed experiments that didn't make it into the paper). 
    \end{itemize}
    
\item {\bf Code of ethics}
    \item[] Question: Does the research conducted in the paper conform, in every respect, with the NeurIPS Code of Ethics \url{https://neurips.cc/public/EthicsGuidelines}?
    \item[] Answer: \answerYes{} %
    \item[] Justification: The research conducted in this paper conforms to the NeurIPS Code of Ethics.
    \item[] Guidelines:
    \begin{itemize}
        \item The answer \answerNA{} means that the authors have not reviewed the NeurIPS Code of Ethics.
        \item If the authors answer \answerNo, they should explain the special circumstances that require a deviation from the Code of Ethics.
        \item The authors should make sure to preserve anonymity (e.g., if there is a special consideration due to laws or regulations in their jurisdiction).
    \end{itemize}

\item {\bf Broader impacts}
    \item[] Question: Does the paper discuss both potential positive societal impacts and negative societal impacts of the work performed?
    \item[] Answer: \answerYes{} %
    \item[] Justification: Discussions of broader impacts are included in Appendix \ref{append:broader-impacts}.
    \item[] Guidelines: 
    \begin{itemize}
        \item The answer \answerNA{} means that there is no societal impact of the work performed.
        \item If the authors answer \answerNA{} or \answerNo, they should explain why their work has no societal impact or why the paper does not address societal impact.
        \item Examples of negative societal impacts include potential malicious or unintended uses (e.g., disinformation, generating fake profiles, surveillance), fairness considerations (e.g., deployment of technologies that could make decisions that unfairly impact specific groups), privacy considerations, and security considerations.
        \item The conference expects that many papers will be foundational research and not tied to particular applications, let alone deployments. However, if there is a direct path to any negative applications, the authors should point it out. For example, it is legitimate to point out that an improvement in the quality of generative models could be used to generate Deepfakes for disinformation. On the other hand, it is not needed to point out that a generic algorithm for optimizing neural networks could enable people to train models that generate Deepfakes faster.
        \item The authors should consider possible harms that could arise when the technology is being used as intended and functioning correctly, harms that could arise when the technology is being used as intended but gives incorrect results, and harms following from (intentional or unintentional) misuse of the technology.
        \item If there are negative societal impacts, the authors could also discuss possible mitigation strategies (e.g., gated release of models, providing defenses in addition to attacks, mechanisms for monitoring misuse, mechanisms to monitor how a system learns from feedback over time, improving the efficiency and accessibility of ML).
    \end{itemize}
    
\item {\bf Safeguards}
    \item[] Question: Does the paper describe safeguards that have been put in place for responsible release of data or models that have a high risk for misuse (e.g., pre-trained language models, image generators, or scraped datasets)?
    \item[] Answer: \answerYes{} %
    \item[] Justification: This paper poses no such risks. 
    \item[] Guidelines:
    \begin{itemize}
        \item The answer \answerNA{} means that the paper poses no such risks.
        \item Released models that have a high risk for misuse or dual-use should be released with necessary safeguards to allow for controlled use of the model, for example by requiring that users adhere to usage guidelines or restrictions to access the model or implementing safety filters. 
        \item Datasets that have been scraped from the Internet could pose safety risks. The authors should describe how they avoided releasing unsafe images.
        \item We recognize that providing effective safeguards is challenging, and many papers do not require this, but we encourage authors to take this into account and make a best faith effort.
    \end{itemize}

\item {\bf Licenses for existing assets}
    \item[] Question: Are the creators or original owners of assets (e.g., code, data, models), used in the paper, properly credited and are the license and terms of use explicitly mentioned and properly respected?
    \item[] Answer: \answerYes{} %
    \item[] Justification: This paper uses open-access datasets, and we cite original sources of this dataset in Section \ref{sec:realdata}.
    \item[] Guidelines:
    \begin{itemize}
        \item The answer \answerNA{} means that the paper does not use existing assets.
        \item The authors should cite the original paper that produced the code package or dataset.
        \item The authors should state which version of the asset is used and, if possible, include a URL.
        \item The name of the license (e.g., CC-BY 4.0) should be included for each asset.
        \item For scraped data from a particular source (e.g., website), the copyright and terms of service of that source should be provided.
        \item If assets are released, the license, copyright information, and terms of use in the package should be provided. For popular datasets, \url{paperswithcode.com/datasets} has curated licenses for some datasets. Their licensing guide can help determine the license of a dataset.
        \item For existing datasets that are re-packaged, both the original license and the license of the derived asset (if it has changed) should be provided.
        \item If this information is not available online, the authors are encouraged to reach out to the asset's creators.
    \end{itemize}

\item {\bf New assets}
    \item[] Question: Are new assets introduced in the paper well documented and is the documentation provided alongside the assets?
    \item[] Answer: \answerYes{} %
    \item[] Justification: Our code for implementing the proposed CoDEAL method is available in the supplementary material, along with the proper documentation. 
    \item[] Guidelines:
    \begin{itemize}
        \item The answer \answerNA{} means that the paper does not release new assets.
        \item Researchers should communicate the details of the dataset\slash code\slash model as part of their submissions via structured templates. This includes details about training, license, limitations, etc. 
        \item The paper should discuss whether and how consent was obtained from people whose asset is used.
        \item At submission time, remember to anonymize your assets (if applicable). You can either create an anonymized URL or include an anonymized zip file.
    \end{itemize}

\item {\bf Crowdsourcing and research with human subjects}
    \item[] Question: For crowdsourcing experiments and research with human subjects, does the paper include the full text of instructions given to participants and screenshots, if applicable, as well as details about compensation (if any)? 
    \item[] Answer: \answerNA{} %
    \item[] Justification: This paper does not involve crowdsourcing or research with human subjects. 
    \item[] Guidelines:
    \begin{itemize}
        \item The answer \answerNA{} means that the paper does not involve crowdsourcing nor research with human subjects.
        \item Including this information in the supplemental material is fine, but if the main contribution of the paper involves human subjects, then as much detail as possible should be included in the main paper. 
        \item According to the NeurIPS Code of Ethics, workers involved in data collection, curation, or other labor should be paid at least the minimum wage in the country of the data collector. 
    \end{itemize}

\item {\bf Institutional review board (IRB) approvals or equivalent for research with human subjects}
    \item[] Question: Does the paper describe potential risks incurred by study participants, whether such risks were disclosed to the subjects, and whether Institutional Review Board (IRB) approvals (or an equivalent approval/review based on the requirements of your country or institution) were obtained?
    \item[] Answer: \answerNA{} %
    \item[] Justification: This paper does not involve research with human subjects.
    \item[] Guidelines:
    \begin{itemize}
        \item The answer \answerNA{} means that the paper does not involve crowdsourcing nor research with human subjects.
        \item Depending on the country in which research is conducted, IRB approval (or equivalent) may be required for any human subjects research. If you obtained IRB approval, you should clearly state this in the paper. 
        \item We recognize that the procedures for this may vary significantly between institutions and locations, and we expect authors to adhere to the NeurIPS Code of Ethics and the guidelines for their institution. 
        \item For initial submissions, do not include any information that would break anonymity (if applicable), such as the institution conducting the review.
    \end{itemize}

\item {\bf Declaration of LLM usage}
    \item[] Question: Does the paper describe the usage of LLMs if it is an important, original, or non-standard component of the core methods in this research? Note that if the LLM is used only for writing, editing, or formatting purposes and does \emph{not} impact the core methodology, scientific rigor, or originality of the research, declaration is not required.
    \item[] Answer: \answerNA{} %
    \item[] Justification: The core method development in this research does not involve LLMs as any important, original, or non-standard components.
    \item[] Guidelines:
    \begin{itemize}
        \item The answer \answerNA{} means that the core method development in this research does not involve LLMs as any important, original, or non-standard components.
        \item Please refer to our LLM policy in the NeurIPS handbook for what should or should not be described.
    \end{itemize}

\end{enumerate}

\end{document}